\pdfoutput=1
\documentclass[11pt]{article}

\usepackage[margin=1in]{geometry}
\usepackage{amsmath,amssymb,amsthm,booktabs,enumitem,mathtools,mathrsfs,array}
\usepackage{algorithm,algpseudocode,placeins}
\usepackage[title]{appendix}
\usepackage{graphicx}

\usepackage[authoryear,round]{natbib}
\bibpunct[, ]{(}{)}{,}{a}{}{,}
\usepackage{xcolor}
\usepackage[colorlinks=true,citecolor=blue,linkcolor=black,urlcolor=black]{hyperref}
\hypersetup{pdftitle={Pricing Today, Shaping Tomorrow: Contextual Dynamic Pricing in Purchase-Driven Evolving Systems},
            pdfauthor={Jinjiang Zou, Yang Li, Zemin Zheng}}

\newtheorem{assumption}{Assumption}
\newtheorem{theorem}{Theorem}
\newtheorem{proposition}{Proposition}
\newtheorem{lemma}{Lemma}
\newtheorem{corollary}{Corollary}
\theoremstyle{definition}
\newtheorem{definition}{Definition}
\theoremstyle{remark}
\newtheorem{remark}{Remark}

\newcommand{\E}{\mathbb E}
\newcommand{\Pp}{\mathbb P}
\newcommand{\Reg}{\operatorname{Reg}}
\newcommand{\cP}{\mathcal P}

\newcommand{\spn}{\operatorname{span}}
\newcommand{\one}{\mathbf 1}
\newcommand{\abs}[1]{\left\lvert#1\right\rvert}

\newcommand{\cX}{\mathcal X}
\newcommand{\cQ}{\mathcal Q}
\newcommand{\cW}{\mathcal W}
\newcommand{\cB}{\mathcal B}
\newcommand{\cC}{\mathcal C}

\newcommand{\cE}{\mathcal E}

\newcommand{\R}{\mathbb R}
\newcommand{\SHAPE}{\textsc{SHAPE}}
\numberwithin{equation}{section}

\title{Pricing Today, Shaping Tomorrow: Contextual Dynamic Pricing in Purchase-Driven Evolving Systems}
\author{%
  \begin{tabular*}{.96\textwidth}{@{\extracolsep{\fill}}ccc@{}}
    Jinjiang Zou$^{\dagger}$ & Yang Li$^{\dagger}$ & Zemin Zheng$^{\dagger}$ \\[3pt]
    {\footnotesize\href{mailto:zjj22chuang2@mail.ustc.edu.cn}{\texttt{zjj22chuang2@mail.ustc.edu.cn}}} &
    {\footnotesize\href{mailto:tjly@ustc.edu.cn}{\texttt{tjly@ustc.edu.cn}}} &
    {\footnotesize\href{mailto:zhengzm@ustc.edu.cn}{\texttt{zhengzm@ustc.edu.cn}}}
  \end{tabular*}\\[6pt]
  {\footnotesize $^{\dagger}$School of Management, University of Science and Technology of China,}\\[-1pt]
  {\footnotesize Hefei 230026, China}
}
\date{}

\begin{document}
\maketitle
\begin{abstract}
Across diverse markets, it is important for firms to balance attracting
customers with managing operational conditions, such as workloads and
delivery times. Dynamic pricing plays a central role in this tradeoff, as
the orders generated today can change the conditions faced by future
customers. We develop a novel framework for contextual dynamic pricing
with an observed shared state and unknown purchase-driven transitions.
The valuation model allows unknown mean functions of customer covariates
and state, unknown state-specific scales and a common unknown standardized
noise distribution. We propose Shadow-Adjusted Pricing in Evolving
Systems (\SHAPE), which incorporates a purchase's future revenue
consequences into current pricing. Purchase-shadow adjustment and scale
standardization reveal a common one-dimensional oracle price map across
states, allowing \SHAPE\ to retain its pricing history across value
updates without repeating coarse exploration. The same active sales
records support price-map learning and joint updates of continuation
values and purchase shadows, without fitting the noise distribution or
full transition matrices. To support this reuse under purchase-driven
feedback, we establish pricing guarantees for history-dependent inputs
and stability bounds for the joint updates. Under regularity, coverage
and design conditions, we obtain sublinear expected regret against a
finite-horizon dynamic oracle for linear, H\"older and sparse linear mean
models. For affine valuations at fixed state and feature dimensions,
\SHAPE\ matches the static semiparametric minimax time exponent up to
logarithmic factors for every fixed noise-tail smoothness order
$\beta\ge2$. Numerical experiments further demonstrate the appealing
performance of \SHAPE.
\end{abstract}
\section{Introduction}
\label{sec:introduction}

Contextual pricing uses observed features to learn customers' willingness
to pay and tailor prices \citep{cohen2020features,bankeskin2021}.
In a shared sales system, current prices can also affect the
conditions under which later customers buy
\citep{celikmaglaras2008,zhang2022relationships}. For a service provider,
outstanding orders may lengthen announced delivery times, reducing the
attractiveness of an offer to customers who value speed. A new purchase
can add to this workload and change the conditions faced by the next
customer. Evidence from retail queues shows that purchase incidence
responds to waiting and that waiting sensitivity differs across customers
\citep{lu2013queues}. Customer covariates describe the current buyer and
request; the shared state carries the effects of earlier sales into later
pricing decisions.

This link between current sales and future conditions creates an
intertemporal pricing tradeoff. A lower price may encourage a purchase
and raise current revenue, yet move the system toward states in which
later sales are less valuable. When customer response and purchase-driven
transitions are unknown, the seller must learn to evaluate both effects
from prices, binary purchase responses and observed next states. The
prices used to collect these observations also influence future sales
conditions. Pricing must therefore support learning while accounting
for the revenue consequences of the state path it creates.

Existing contextual-pricing methods learn customer response from binary
sales without specifying the valuation-noise distribution
\citep{fan2024semiparametric,chen2024doubly}, and \citet{orbit} learn the
oracle price map directly. Adding the system state to a myopic pricing
rule can capture its effect on current purchases, but not the future
value of purchase-induced state changes.
\citet{zhang2022relationships} analyze purchase-driven customer
relationships with a known model, while \citet{huliwager} study
shared-state contextual decisions in an offline policy-learning framework.
We ask how to learn contextual prices online when both valuations and
purchase-driven shared-state transitions are unknown, with performance
evaluated over the state path generated by the pricing policy.

In this paper, we develop a novel framework for contextual dynamic pricing
in purchase-driven evolving systems. The seller observes each customer's
covariates and a shared finite system state, while mean valuation,
state-specific scales and the common standardized noise distribution are
unknown. Mean valuations may depend nonlinearly on covariates and differ
across states, and purchases induce unknown stochastic state transitions.
We propose Shadow-Adjusted Pricing in Evolving Systems (\SHAPE), which
values a purchase through its current revenue and its effect on future
revenue. Its key device is the purchase shadow, the difference in expected
continuation value after a purchase and a nonpurchase. Shadow adjustment
and scale standardization reveal a common one-dimensional oracle price
map across states. Future-value estimates can change its inputs without
changing the map, allowing the pricing policy to reuse observations as
continuation-value estimates are updated.

We establish regret guarantees against a finite-horizon dynamic oracle
that knows the valuation and transition models. Under the stated
structural and design conditions, \SHAPE\ matches the static
semiparametric minimax time exponent for affine valuations at fixed state
and feature dimensions, despite learning unknown purchase-driven
dynamics. The same policy gives explicit regret upper bounds for nonlinear
H\"older and sparse linear valuations. In a workload system, we compare
\SHAPE\ with state-aware myopic pricing, dynamic plug-in pricing and a
state-blind price-map method. Across 200 paired replications in the reported
setting, \SHAPE\ has the lowest mean regret among these four implementations,
41.25\% below myopic pricing and 14.80\% below dynamic plug-in pricing
(Table~\ref{tab:workload-regret}).

\subsection{Main contributions}
\label{sec:contributions}

Our contributions are threefold. First, to our knowledge, this paper
provides the first online-learning treatment of semiparametric contextual
pricing with shared purchase-driven state transitions and regret
guarantees against a finite-horizon dynamic oracle. Our valuation model
allows nonlinear mean effects and unknown state-specific dispersion while
retaining a common unknown standardized noise distribution. Performance
is measured by comparing the learner's total revenue with the oracle's,
allowing each policy to generate its own state trajectory. This criterion
accounts for how current prices influence future sales opportunities,
in addition to their immediate revenue.

Second, we develop a novel learning architecture that couples one
persistent, cross-state price-map learner with joint value and
purchase-shadow updates. The shared pricing structure allows the learner
to retain its observations as future-value estimates change, avoiding
repeated coarse exploration after each value update. Building on direct
oracle-map learning \citep{orbit}, \SHAPE\ learns the common map from
binary purchases without first fitting the noise distribution. It
estimates purchase shadows by comparing estimated continuation values
at the next states observed after purchases and nonpurchases, without
fitting full transition matrices. A further methodological innovation is
to use the same active sales records to update continuation values and
purchase shadows, so value learning progresses alongside price-map
learning. Different mean-estimation routines specialize this same
architecture to linear, nonlinear H\"older and sparse linear valuations.

Third, we develop a joint analysis of price learning and value estimation
under purchase-driven state transitions. Prices affect the purchase and
next-state observations used for value updates, while these updates shape
subsequent pricing inputs. We control the resulting feedback through
regret bounds for the persistent pricing learner and a stability analysis
of the joint value and shadow updates. Combining these results with mean
and scale estimation bounds yields regret guarantees against the
finite-horizon dynamic oracle.
For affine mean valuations with fixed state and feature dimensions, the
upper and lower bounds match up to logarithmic factors under the stated
structural and design conditions.
Thus, for every fixed noise-tail smoothness order $\beta\ge2$, \SHAPE\
matches the static semiparametric minimax time exponent \citep{orbit}
while also learning unknown purchase-driven dynamics. The same analysis
gives explicit regret upper bounds for nonlinear H\"older and sparse linear
valuations through the mean-estimation interface.

\subsection{Related literature}
\label{sec:related-literature}

\paragraph{Pricing with evolving customer and market states.}
Closely related pricing models link current choices to future customer
relationships and customer pools. \citet{zhang2022relationships}
characterize optimal prices when purchases change a firm's relationship
with a customer, taking the response and transition models as given.
\citet{kostamikoenigsberg} jointly study pricing, customer acquisition,
retention and capacity with unknown demand and customer-transition
functions. Their learning guarantee uses stage-constrained one-period
optima during learning and a best fixed action during deployment,
evaluated at the realized state and context. Our focus is online pricing
for a state shared across successive customers, with regret comparing
the learner's and dynamic oracle's revenues along their respective
state paths.

Other pricing models capture persistence through customer preferences,
reference prices, adoption or arrival history.
\citet{keskinli2024markov} study quality-differentiated products when
customer preferences follow a Markovian process with unknown transition
probabilities. Under reference effects, \citet{denboerkeskin2022}
study learning with unobserved reference prices, while
\citet{agrawaltang2024} analyze long-term reference effects with a
dynamic-oracle regret guarantee. Purchases drive adoption in the Bass
model of \citet{agrawal2021bass} and stimulate future arrivals in the
self-exciting model of \citet{yuan2026selfexciting}. In our model, the
horizon counts quotation opportunities; purchase outcomes drive unknown
transitions of a shared state faced by later customers.

\paragraph{Contextual pricing with unknown valuations.}
Learning prices from binary sales is a classical exploration--exploitation
problem, including under parametric choice models
\citep{broderrusmevichientong2012}. Observed features introduce
heterogeneity into this problem, with results for linear valuations
\citep{cohen2020features} and sparse high-dimensional response models
\citep{bankeskin2021}. \citet{xuwang2024heteroscedastic} further study
contextual elasticity and heteroscedastic valuations under a known noise
distribution. Nonparametric approaches allow the purchase-response
function to vary flexibly with covariates and price
\citep{chengallego2021}. Within random-utility models,
\citet{fan2024semiparametric} combine linear-mean estimation with an
unknown noise distribution, while \citet{chen2024doubly} allow both mean
utility and noise distribution to be nonparametric.

Recent work also addresses what must be learned and which observations
can be used. \citet{chai2026featurediversity} use feature diversity to
refine noise estimates with observations collected during exploitation.
\citet{orbit} instead learn the oracle price map directly and provide a
modular utility-estimation interface, including nonlinear and sparse
extensions. We adopt ORBIT's scalar grid-search and local-polynomial
construction. Its scalar-pilot analysis excludes feedback from its own
active pricing calls when forming future inputs and stopping decisions.
Here active purchases change subsequent shared states and shadow
estimates, so the learner's later inputs can depend on its own earlier
pricing feedback. Retaining pooled observations therefore requires a
guarantee for this feedback-dependent pricing process.

\paragraph{Pricing and learning in service systems.}
Service-system pricing also links customer response to operating
conditions. \citet{celikmaglaras2008} jointly optimize
dynamic prices, quoted lead times and operational controls for
price- and delay-sensitive customers. With unknown customer response,
\citet{jia2024reusable} learn prices in service systems with reusable
resources and unknown price-dependent arrival and service rates.
\citet{chen2026queues} learn prices and capacity when the
arrival-rate function and service distribution are unknown, accounting
for transient queueing effects while optimizing steady-state profit.
In a competitive setting, \citet{lian2026competition} studies
queue-based service systems,
learning customers' sensitivity to total price-and-waiting cost from
routine arrivals.
Our focus is learning from individual quotations, with binary purchase
outcomes and observed next states linked by purchase-conditioned
transitions.

\paragraph{Policy learning in dynamic systems.}
Regret against an optimal policy in an unknown controlled Markov system
is an established objective in reinforcement learning; for example,
\citet{jaksch2010rl} give guarantees for communicating finite-state,
finite-action models. \citet{huliwager} study optimal targeting when
covariates arrive within a shared evolving system and actions influence
subsequent states. Their offline policy-learning framework separates
fresh covariates from persistent system state and uses relative-value
methods to evaluate long-run consequences. Our continuous prices must
instead be learned online while the seller competes with a finite-horizon
dynamic oracle. This requires joint control of price-map learning and
value and shadow updates computed from the same sales records.

\subsection{Organization and notation}
\label{sec:organization}

Section~\ref{sec:model} introduces the model, benchmark and structural
conditions. Section~\ref{sec:method} presents \SHAPE, and
Section~\ref{sec:regret} establishes its regret guarantees and the
fixed-dimensional affine minimax result.
Section~\ref{sec:valuation-models} develops the nonlinear and sparse
specializations.
Section~\ref{sec:experiments} reports the numerical comparisons, and
Section~\ref{sec:conclusion} concludes the paper.
The appendices provide implementation details and proofs.

For a positive integer $n$, write $[n]=\{1,\ldots,n\}$.
We use $\one$ for an all-ones vector and $\one\{\cdot\}$ for an
indicator. For a vector $v$, $\|v\|_q$ denotes its $\ell_q$ norm,
$\|v\|_0$ counts its nonzero coordinates, and $\operatorname{supp}(v)$
is its support. Matrix spectral and entrywise maximum norms
are written $\|A\|_{\rm op}$ and $\|A\|_{\max}$.
Unsubscripted norms are Euclidean for vectors and spectral for matrices.
For symmetric matrices, $A\preceq B$ means that $B-A$
is positive semidefinite, and $\lambda_{\min}(A)$ is the smallest
eigenvalue. For a function, $\|f\|_\infty$ is the supremum norm
over its stated domain.

For $h\in\R^K$, its span seminorm is
$\spn(h)=\max_s h(s)-\min_s h(s)$.
For a closed interval $C$, $\Pi_C$ denotes clipping to $C$.
A transition kernel acts on a state-value vector through
$(Ph)(s)=\sum_{s'=1}^K P(s'\mid s)h(s')$.
For nonnegative sequences, $a_n\lesssim b_n$ means that
$a_n\le Cb_n$ for a constant independent of $n$, and
$a_n\asymp b_n$ means $a_n\lesssim b_n$ and $b_n\lesssim a_n$.
The symbols $\widetilde O$ and $\widetilde\Theta$ suppress
logarithmic factors. Dependence of constants on model parameters
is specified with each result.

\section{Model, information, and benchmark}
\label{sec:model}

\subsection{Customer arrivals and purchase-driven dynamics}

We consider an evolving multi-state sales system in which a seller sells a single product
over a horizon of $T$ periods. At each time $t=1,\ldots,T$, the seller observes the
current system state $S_t\in[K]=\{1,\ldots,K\}$ and the arriving customer's
covariate vector $X_t\in\cX\subseteq[0,1]^d$, drawn from an unknown
distribution $P_X$ with support $\cX$.
We model the customer's latent valuation as
\begin{equation}
 V_t=m_*(X_t,S_t)+\sigma_{S_t}\xi_t.
 \label{eq:valuation-model}
\end{equation}
Here $m_*(x,s)$ is the unknown mean valuation for covariates $x$ and
system state $s\in[K]$, and $\sigma_s$ is the unknown noise scale
in that state, with known bounds
$0<\underline\sigma\le\sigma_s\le\overline\sigma<\infty$.
The valuation noise $\xi_t$ has a common unknown distribution
$P_\xi$, normalized to mean zero and variance one. The covariate vector
and valuation noise are drawn independently from $P_X$ and $P_\xi$,
respectively, and are jointly independent of the current state and past
observations.
Throughout the paper, we study a normalized model class in which
$V_t\in[p_L,p_U]$ almost surely in every state. Accordingly, at time $t$,
the seller posts a price $p_t\in\cP=[p_L,p_U]$, with
$0\le p_L<p_U$, and then observes $Y_t$ but not the customer's
valuation $V_t$ or valuation noise $\xi_t$, where
$Y_t=\one\{V_t\ge p_t\}$ indicates whether the customer purchases.

\begin{figure}[!htbp]
\centering
\includegraphics[width=.98\linewidth]{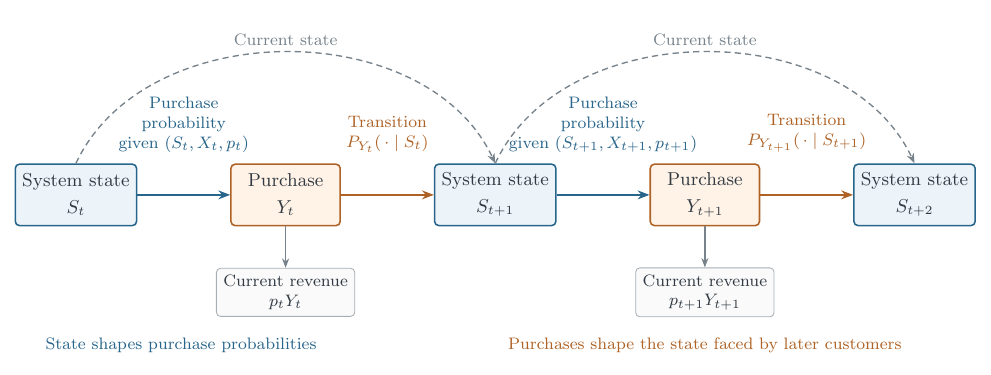}
\caption{Purchase-driven feedback across successive customers.}
\label{fig:purchase-state-feedback}
\end{figure}

After the purchase decision, the system evolves from $S_t$ to $S_{t+1}$,
following the fixed and unknown transition distributions
$P_0(\cdot\mid S_t)$ and $P_1(\cdot\mid S_t)$ after a nonpurchase and
a purchase, respectively (Figure \ref{fig:purchase-state-feedback}):
\begin{equation}
 S_{t+1}\mid(S_t,Y_t)\sim P_{Y_t}(\cdot\mid S_t).
 \label{eq:purchase-transition}
\end{equation}
Given $S_t$ and $Y_t$, the next state is drawn independently of the current
covariates, price, valuation noise, and past observations, and the seller
then observes $S_{t+1}$. The seller's objective
is to learn a pricing policy that maximizes expected cumulative revenue
over the $T$ periods.

\begin{remark}[Special cases and economic interpretation]
\label{rem:state-scale-interpretation}
The mean function $m_*(x,s)$ accommodates nonlinear effects of customer
characteristics and their interactions with system state. It includes
state-specific linear utilities $m_*(x,s)=a_s+x^\top\theta_s$ and
additive state effects $m_*(x,s)=f(x)+a_s$, while setting
$\sigma_s=\sigma$ gives a common noise scale. With $K=1$, the model
reduces to contextual pricing with additive valuation noise \citep{orbit}.
Price- and delay-sensitive service models motivate the following
illustration \citep{celikmaglaras2008}. Consider heterogeneous delay
costs, and let $w(S_t)>0$ be the announced
delivery time, $r(X_t)$ the service value without delay, and
$A_t=\bar a(X_t)-\tau_A\xi_t>0$ the customer's delay sensitivity,
where $\tau_A>0$ is constant. Then
\[
 V_t=r(X_t)-A_tw(S_t)=
 \underbrace{r(X_t)-\bar a(X_t)w(S_t)}_{m_*(X_t,S_t)}
 +\underbrace{\tau_Aw(S_t)}_{\sigma_{S_t}}\xi_t.
\]
\end{remark}
\FloatBarrier

\subsection{Dynamic oracle and regret}
\label{sec:oracle-regret}

We compare the seller with an oracle that knows the covariate
distribution, purchase probabilities, and transition distributions.
From the valuation model, we have
\begin{equation}
 \begin{aligned}
 D_s(p\mid x)
 &:=\Pp(V_t\ge p\mid X_t=x,S_t=s)\\
 &=\Pp\!\left(\xi_t\ge\frac{p-m_*(x,s)}{\sigma_s}\right)\\
 &=g\!\left(\frac{p-m_*(x,s)}{\sigma_s}\right),
 \end{aligned}
 \label{eq:demand}
\end{equation}
where $g(w)=\Pp(\xi_t\ge w)$ is the survival function of the valuation
noise. At price $p$, the seller's expected revenue is $pD_s(p\mid x)$,
while the next-state distribution is
\[
 \Pp(S_{t+1}=s'\mid S_t=s,X_t=x,p_t=p)
 =\{1-D_s(p\mid x)\}P_0(s'\mid s)+D_s(p\mid x)P_1(s'\mid s).
\]

Let $\Pi_T$ denote all possibly randomized pricing policies that choose
prices in $\cP$ using only the information available to the seller at
each pricing decision.
The oracle's optimal expected cumulative revenue from initial state $s$ is
\begin{equation}
 \mathcal V_T^*(s)=
 \sup_{\pi\in\Pi_T}
 \E_s^\pi\sum_{t=1}^T p_tY_t.
 \label{eq:finite-horizon-oracle}
\end{equation}
We define the regret of a learning policy $\pi\in\Pi_T$ relative to
this oracle as
\begin{equation}
 \Reg_T(\pi;s)=\mathcal V_T^*(s)-
 \E_s^\pi\sum_{t=1}^T p_tY_t.
 \label{eq:finite-horizon-regret}
\end{equation}
The two expected revenues are evaluated over the same $T$ periods
from the same initial state $s$, with each policy generating its own
state path. We write $\Reg_T$ when the policy and initial state are clear
from context.

To characterize the oracle's optimal revenue, let $J_n^*(s)$ denote its
optimal expected revenue from state $s$ with $n$ periods remaining and
set $J_0^*\equiv0$. Since future covariates are independent of past
observations, the continuation value after the current period is
$J_{n-1}^*(S_{t+1})$. Conditional on purchase outcome $y$, its expected
value is $(P_yJ_{n-1}^*)(s)$, where
$(P_yh)(s)=\sum_{s'=1}^K P_y(s'\mid s)h(s')$ for a continuation
vector $h$. Combining current revenue with this continuation value
gives the Bellman recursion
\begin{equation}
 J_n^*(s)=\E_X\sup_{p\in\cP}
 \left[(P_0J_{n-1}^*)(s)
 +\{p+((P_1-P_0)J_{n-1}^*)(s)\}D_s(p\mid X)\right],
 \qquad \mathcal V_T^*(s)=J_T^*(s).
 \label{eq:finite-horizon-recursion}
\end{equation}

In \eqref{eq:finite-horizon-recursion}, $(P_0J_{n-1}^*)(s)$ is the
expected continuation value after a nonpurchase and does not depend
on the current price. The difference
$b_s(h)=\{(P_1-P_0)h\}(s)$, called the \emph{purchase shadow}, measures
how a purchase changes expected future revenue relative to a
nonpurchase. The oracle therefore selects its current price by
maximizing $(p+b_s(J_{n-1}^*))D_s(p\mid X)$, which accounts for both
current revenue and the purchase's effect on future revenue.

\subsection{Pricing regularity}
\label{sec:structural-conditions}

This pricing objective can be written in a common scalar form across
states. For a continuation vector $h$, set
\[
 q=\frac{p+b_s(h)}{\sigma_s},\qquad
 z=\frac{m_*(x,s)+b_s(h)}{\sigma_s},
\]
to obtain
\[
 (p+b_s(h))D_s(p\mid x)=\sigma_s qg(q-z).
\]
Here $q-z=\{p-m_*(x,s)\}/\sigma_s$ is the standardized valuation
residual.
Let $\mathcal Z$ and $\cQ$ be the supplied compact input and action
intervals. Write $f_z(q)=qg(q-z)$ and let $\psi(z)$ denote a maximizer
of $f_z$ over $\cQ$. The positive factor $\sigma_s$ does not affect
maximization on this common standardized-action interval, giving one
scalar map for all states. Assumption~\ref{ass:utility-domain} below
relates this map to the original physical price interval. We first
state conditions on the tail $g$ and the scalar pricing objective.

\begin{assumption}[Tail smoothness]
\label{ass:tail-regularity}
On a known compact residual interval $\mathcal I$, $g$ belongs to the
$\beta$-H\"older ball of radius $L_g$, where $\beta\ge2$. We use the
H\"older convention in Appendix \ref{app:domain-details}.
\end{assumption}

Assumption \ref{ass:tail-regularity} bounds how purchase probabilities
change with the valuation residual. As in \citet{orbit}, we impose
smoothness on the noise tail without specifying a parametric noise
distribution. Smoothness alone does not determine the shape of the
pricing objective, so we also impose the following revenue conditions.

\begin{assumption}[Revenue geometry and oracle-map regularity]
\label{ass:demand-geometry}
The following conditions hold on $\mathcal Z\times\cQ$ and on the
fixed enlarged domains in Appendix \ref{app:domain-details}.
\begin{enumerate}[label=(\roman*),leftmargin=*]
\item \emph{Quadratic growth.}
The maximizer $\psi(z)$ is unique, lies in $\operatorname{int}(\cQ)$
and has a fixed positive distance from its boundary. For fixed constants
$0<\lambda_r\le L_r$,
\begin{equation}
 \frac{\lambda_r}{2}\abs{q-\psi(z)}^2
 \le f_z(\psi(z))-f_z(q)
 \le\frac{L_r}{2}\abs{q-\psi(z)}^2
 \label{eq:quadratic-growth}
\end{equation}
uniformly over the stated domains.
\item \emph{Map regularity and local concavity.}
The map $\psi$ belongs to the $(\beta-1)$-H\"older ball of radius
$L_\psi$ on the stated input domain. In addition, there is a fixed
$\rho_0>0$ such that
\[
 -\frac{\partial^2}{\partial q^2}f_z(q)\ge\lambda_r/2
 \quad\text{whenever }\abs{q-\psi(z)}\le\rho_0.
\]
\end{enumerate}
\end{assumption}

Part (i) bounds the loss in $f_z$ above and below by squared action
error; it does not require concavity over the whole action interval.
Part (ii) controls how the best action changes with $z$ and gives a
concave neighborhood in which to refine a local polynomial rule.
ORBIT derives corresponding map regularity and local concavity from
its basic pricing assumptions. Here we require these properties
explicitly on the enlarged domains used for input and scale errors.
Appendix \ref{app:domain-details} specifies those domains and margins.

\subsection{State coverage and the stationary oracle}
\label{sec:stationary-target}

Learning the purchase shadow requires observations of both purchase
outcomes and the states that follow them. Prices affect which of
these observations the seller receives: at purchase probability
$\lambda\in[0,1]$, the next-state distribution is
\[
 Q_s^\lambda=(1-\lambda)P_0(\cdot\mid s)
                 +\lambda P_1(\cdot\mid s).
\]
We therefore impose coverage on purchase outcomes and on this
effective transition kernel.

\begin{assumption}[Purchase-outcome and state coverage]
\label{ass:coverage}
The algorithm is supplied with a full-support probability vector $\nu$ on
$[K]$ and constants $\kappa_D>0$, $a_0>0$ and
$\kappa_0\in(0,1/2)$, with $\kappa_0\le\kappa_D$, such that
\begin{enumerate}[label=(\roman*),leftmargin=*]
\item $g(w)\in[\kappa_D,1-\kappa_D]$ for every $w\in\mathcal I$;
\item the effective kernel satisfies
\begin{equation}
 Q_s^\lambda\ge a_0\nu
 \quad\text{coordinatewise for every }s\in[K]
 \text{ and }\lambda\in[\kappa_0,1-\kappa_0].
 \label{eq:effective-minorization}
\end{equation}
\end{enumerate}
\end{assumption}

Part (i) keeps both purchase outcomes observable on the residual
interval used for learning. Part (ii) gives every state a positive
visitation probability in the stated purchase-probability range.
This is a one-step coverage condition,
stronger than irreducibility in the stated purchase-probability
range; neither $P_0$ nor $P_1$ must satisfy it separately. A
purchase-independent reset into a common full-support distribution
with uniformly positive probability is one sufficient mechanism.
\citet{huliwager} instead impose uniform mixing on true and
estimated kernels across stationary policies; the
kernel and policy ranges differ from those considered here.

The finite-horizon oracle uses a different continuation vector for
each number of periods remaining. For learning, we instead use a
stationary relative value and relate its pricing loss to finite-horizon
regret. Following the relative-value analysis of \citet{huliwager},
consider a reference Bellman problem with prices restricted to
\[
 \cP^{\rm b}(x,s)=
 \{p\in\cP:(p-m_*(x,s))/\sigma_s\in\cW^{\rm b}\},
\]
where $\cW^{\rm b}$ is a fixed compact residual subinterval of
$\mathcal I$; Appendix~\ref{app:domain-details} specifies its margins.
When these price sets are nonempty, as required below, coverage makes
the reference Bellman operator a contraction in span. It therefore
has a relative value $h^*$ and an average revenue $\rho^*$.
The vector $h^*$ compares future revenue across starting states and
is defined up to an additive constant.

Normalize $h^*(s_{\rm ref})=0$ at a fixed reference state. Coverage
gives $\operatorname{span}(h^*):=\max_s h^*(s)-\min_s h^*(s)
\le H_V:=p_U/a_0$, and its purchase shadow is
$b_s^*=\{(P_1-P_0)h^*\}(s)$. Substituting this shadow
into the common pricing map gives the candidate stationary price
\begin{equation}
 p^*(x,s)=\sigma_s\psi\!\left(
   \frac{m_*(x,s)+b_s^*}{\sigma_s}\right)-b_s^*.
 \label{eq:physical-oracle}
\end{equation}
To compare with the oracle in Section \ref{sec:oracle-regret}, this
price must maximize purchase-adjusted revenue over the original
interval $\cP$, not just over the reference prices.

\begin{assumption}[Full-interval stationary oracle]
\label{ass:utility-domain}
There is a known compact interval $\cB=[b_L,b_U]$ such that, for every
model in the class, all true shadows $b_s^*$ belong to $\cB$ and all
oracle indices $z_s^*(x)=\{m_*(x,s)+b_s^*\}/\sigma_s$ belong to
$\mathcal Z$.

The reference price sets defined in Appendix \ref{app:domain-details}
are nonempty. For every $(x,s)$, the price in
\eqref{eq:physical-oracle} lies a fixed distance inside that reference
set and maximizes $(p+b_s^*)D_s(p\mid x)$ over the full interval $\cP$.
\end{assumption}

Assumption \ref{ass:utility-domain} retains a full-price optimum
inside the reference set at the true shadow $b^*$, with a positive
margin. As in ORBIT, it requires an interior optimum; here the
reference prices must also retain the optimum over the original
interval $\cP$. The reference solution then satisfies
\begin{equation}
 \rho^*+h^*(s)=(P_0h^*)(s)+
 \E_X\max_{p\in\cP}(p+b_s^*)D_s(p\mid X).
 \label{eq:full-aroe}
\end{equation}
Implementing this stationary pricing rule requires estimates of mean
valuations, state scales, the common price map and purchase shadows.
Section~\ref{sec:method} develops \SHAPE\ to learn these quantities
from purchase responses and next-state observations.
Section~\ref{sec:finite-horizon-bridge} relates its Bellman pricing
loss to the finite-horizon regret in
\eqref{eq:finite-horizon-regret}, with a boundary term at most $H_V$.

\section{The SHAPE pricing policy}
\label{sec:method}

\SHAPE{} (Shadow-Adjusted Pricing in Evolving Systems) uses the
purchase shadow to account for future revenue when setting the
current price. As \eqref{eq:physical-oracle} shows, adding this
shadow to mean valuation and dividing by the state scale gives a
scalar input to the common oracle-price map. The policy learns one
map across states and converts each standardized action back to a
physical price using the same scale and shadow. Updating the shadow
changes this input and price conversion, but not the target map
$\psi$. Observations from different states and earlier blocks can
therefore contribute to the same pricing learner, whose history is
retained as the estimated future values change.

\begin{figure}[!htbp]
\centering
\includegraphics[width=.98\linewidth]{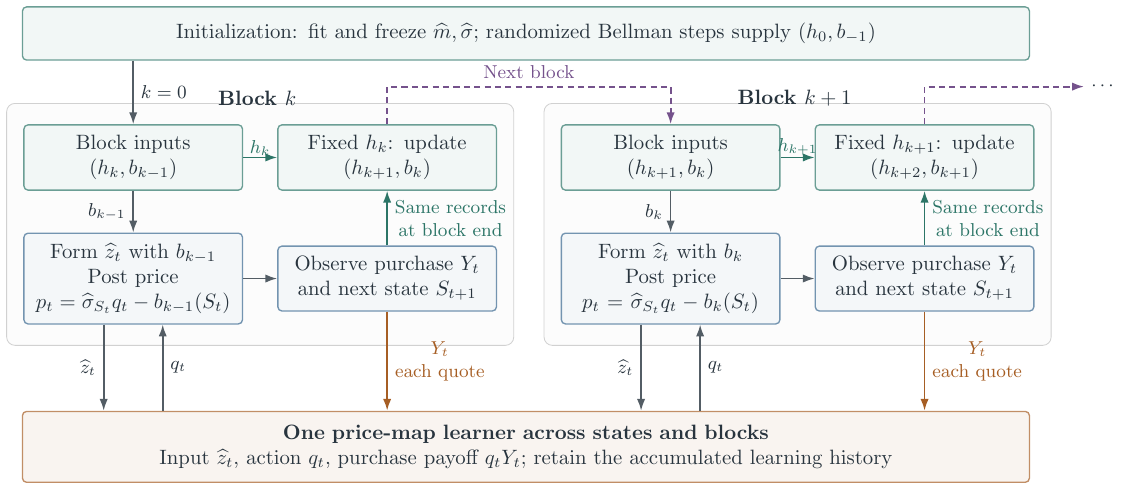}
\caption{Shadow-adjusted pricing and joint value learning across blocks.}
\label{fig:shape-overview}
\end{figure}

Figure \ref{fig:shape-overview} shows how pricing and value learning
work together across blocks. Initialization estimates and freezes
mean valuations and state scales, then uses randomized residual
prices to obtain initial continuation and shadow estimates. Active
block $k$ starts with a fixed continuation vector $h_k$ and an
inherited shadow $b_{k-1}$. Within the block, every purchase outcome
updates the shared pricing learner through the payoff $q_tY_t$.
At the end of a complete block, the same pricing records, combined
with the fixed $h_k$, produce both the next
continuation vector $h_{k+1}$ and shadow $b_k$. These estimates enter
the next block while the pricing learner continues from its existing
history. The shadow is learned by comparing next-state continuation
values after purchases and nonpurchases, without fitting transition
matrices. We first describe
the shared pricing step, then the initialization and joint updates
that supply its state-dependent inputs.

\FloatBarrier

\subsection{Shadow-adjusted pricing with a shared map}
\label{sec:pricing-idea}
\label{sec:policy}

On a pricing call with $(X_t,S_t)=(x,s)$, the outer policy supplies
estimates $\widehat m,\widehat\sigma$ and a shadow $\widehat b$,
which determine the raw scalar input
\begin{equation}
 \widehat z_t=\frac{\widehat m(x,s)+\widehat b_s}{\widehat\sigma_s}.
 \label{eq:actor-input}
\end{equation}
The module projects this input onto its prescribed interval,
$\bar z_t=\Pi_{\mathcal Z}(\widehat z_t)$, before locating its bin
and evaluating its action rule. It selects $q_t\in\cQ$. The seller posts
$p_t=\widehat\sigma_s q_t-\widehat b_s$, observes $Y_t$, and passes
it back to the module. The module computes $q_tY_t$, the normalized
purchase-adjusted payoff, since
$(p_t+\widehat b_s)Y_t/\widehat\sigma_s=q_tY_t$.

The construction below uses locally accurate inputs and feasible
prices, with the margins stated in Assumption
\ref{ass:implementation-domains} in Section \ref{sec:unified-policy}.
For these inputs, we adapt the scalar learner of \citet{orbit}:
a grid search finds a reference action in each input bin, and a
polynomial action rule learns how the action varies within the bin.

\paragraph{Grid search within each input bin.}
Partition $\mathcal Z$ into deterministic bins of width at most $h_T$,
and place a fixed coarse grid $\mathcal G_{\cQ}$ in $\cQ$. Each bin
maintains its own counts and payoff means, pooling calls from all
system states. On successive visits it samples each grid action
$m_{\rm coarse}$ times and records the empirical mean of $q_tY_t$.
After completing this schedule, it stores the empirical maximizer
$\widetilde q_j$ over the full grid using fixed tie-breaking. This anchor identifies the
action range in which local refinement can exploit concavity.
A bin with an incomplete grid schedule continues that schedule on
its next visit.

\paragraph{Learning the within-bin action rule.}
The grid search identifies an action level. A polynomial rule then
learns how the best action varies with the scalar input within the
bin. For
bin midpoint $z_j$, width $\ell_j$ and
$d_{\rm act}=\lceil\beta-1\rceil-1$, define
\begin{equation}
 \varphi(v)=(1,v,\ldots,v^{d_{\rm act}})^\top,\qquad
 q_j(z;\theta)=\theta^\top
       \varphi\!\left(\frac{2(z-z_j)}{\ell_j}\right).
 \label{eq:local-action}
\end{equation}
The allowed coefficient vectors satisfy
$\sup_{|v|\le1}|\theta^\top\varphi(v)-\widetilde q_j|
 \le\rho_{\rm loc}/4$, keeping every candidate rule near the anchor
throughout the bin. As in \citet{orbit}, the module projects the
generated action onto the allowed interval before returning it:
$q_t=\Pi_{\cQ}\{q_j(\bar z_t;\theta_t)\}$.
Under successful localization and the supplied precision settings,
the entire unprojected action range lies inside $\cQ$ and within a
concave part of the expected payoff, so projection is inactive.
Since $q_j(z;\theta)$ is linear in $\theta$, the negative expected
payoff is then convex in the coefficients. A
bin-specific bandit optimizer selects $\theta$ and receives the normalized loss
$\ell_t=1/2-q_tY_t/(2Q_{\max})$, where
$Q_{\max}=1\vee\sup_{q\in\cQ}|q|$.

As in \citet{orbit}, we implement refinement with the normalized
online-Newton bandit routine of \citet{fokkema}, using local doubling
restarts indexed by each bin's refinement visits. The module is
initialized once with budget $T$, which sets
the bin width and coarse sampling counts. Each bin retains its
anchor and ordered action-feedback history between calls; its
optimizer is not given the eventual number of visits to that bin.
A change in the supplied shadow changes the input and physical
price, but not the target map $\psi$, so the bin histories are kept
when the shadow is updated. These outer updates neither restart the
bin clocks nor repeat the coarse search. Algorithm \ref{alg:pooled-module}
gives the pricing and feedback steps.

\begin{algorithm}[H]
\caption{The pooled scalar pricing module}
\label{alg:pooled-module}
\small
\begin{algorithmic}[1]
\Require $T$, domains $\mathcal Z,\cQ$, bin width $h_T$, grid
$\mathcal G_{\cQ}$, $m_{\rm coarse},\rho_{\rm loc}$, and the
refinement generator specified in Appendix \ref{app:operational-details}.
\State Initialize the bins, their grid counters and empirical
payoff means once; retain them across all later calls.
\Statex \textbf{On a pricing call with scalar input $\widehat z_t$:}
\State Check that $\widehat z_t$ is finite; on failure return a flag.
Set $\bar z_t=\Pi_{\mathcal Z}(\widehat z_t)$ and locate its bin $j$.
\If{bin $j$ has not completed coarse localization}
 \State Return its next scheduled grid action $q_t$.
\Else
 \If{this is the first refinement visit to bin $j$}
  \State Store the grid maximizer $\widetilde q_j$ and construct
  its symmetric coefficient trust body.
  \State Initialize a bin-specific refinement generator with an
  empty action-feedback history.
 \EndIf
 \State Obtain $\theta_t$ from that generator and return
 $q_t=\Pi_{\cQ}\{q_j(\bar z_t;\theta_t)\}$.
\EndIf
\Statex \textbf{After the outer policy posts the physical price and
returns $Y_t$:}
\If{the call used a coarse-grid action}
 \State Update that action's count and empirical mean of $q_tY_t$
 in bin $j$.
\Else
 \State Append $(\theta_t,\ell_t)$ to bin $j$'s refinement history.
\EndIf
\end{algorithmic}
\end{algorithm}

The feedback differs from the ideal scalar objective because the
mean and scale are estimated. Conditional on past observations and
the current state and covariates, before fresh optimizer
randomization, a candidate action $q$ has expected payoff
\begin{equation}
 \E[qY_t\mid\text{pre-purchase information and }q]
 =qg\!\left(
    \frac{\widehat\sigma_s}{\sigma_s}q
    -\frac{m_*(x,s)+\widehat b_s}{\sigma_s}\right).
 \label{eq:predictable-feedback}
\end{equation}
With exact mean and scale inputs, this expression reduces to
$f_{\widehat z_t}(q)$. Under the local precision conditions, the raw
input belongs to $\mathcal Z$ and projection leaves it unchanged.
Estimation errors perturb that objective, and
state evolution makes later inputs depend on earlier purchases.
We therefore analyze feedback conditional on the information
available at each call, rather than condition on the full future
input stream as in ORBIT's pilot interface. Appendices
\ref{app:predictable-bandit} and \ref{app:pricing-certificate-proof}
give this analysis. We next construct the estimates supplied to
the module.

\subsection{Constructing the initial inputs}
\label{sec:calibration}

We construct the initial inputs in two stages. Uniform-price
calibration estimates mean valuation and state scales from binary
purchases. Empirical Bellman iterations then initialize the
continuation vector and its purchase shadow using the observed
next states as well as purchase outcomes.

\paragraph{Joint mean and scale calibration.}
The seller does not observe valuations, so the first task is to recover
their conditional means and state scales from binary purchases. During
$n_u$ initial periods, draw $p_t$ uniformly from $[p_L,p_U]$ independently
of the current valuation given the observed state and covariates. With
$B=p_U-p_L$, form the observed responses
\begin{equation}
 Z_t^{(1)}=p_L+BY_t,\qquad
 Z_t^{(2)}=2B(p_t-p_L)Y_t.
 \label{eq:pseudo-responses}
\end{equation}
Since $V_t\in[p_L,p_U]$ and the normalized noise has mean zero and
variance one, integration over the random quote gives
\begin{equation}
 \begin{aligned}
 \E[Z_t^{(1)}\mid X_t=x,S_t=s]&=m_*(x,s),\\
 \E[Z_t^{(2)}\mid X_t=x,S_t=s]
 &=\{m_*(x,s)-p_L\}^2+\sigma_s^2.
 \end{aligned}
 \label{eq:calibration-moments}
\end{equation}
The first response is an unbiased signal of mean valuation; the
second gives a second moment centered at $p_L$. Let $\mathcal M$
be a specified function class containing the statewise mean functions
$m_*(\cdot,s)$. We fit these functions using a regression routine
with the uniform guarantee in Definition
\ref{def:regression-interface} below, and use that fit to remove
the mean contribution from the second moment.

Let $q_S=a_0\min_s\nu(s)$ and
$N_U=\lfloor q_S(n_u-1)/2\rfloor$. Let $\mathcal I_s^U$ contain
all calibration visits to state $s$, and write $N_s^U=|\mathcal I_s^U|$.
Use all $N_s^U$ records for both mean fitting and scale estimation,
with $N_U$ as the required minimum, and project the mean fit into
$[p_L,p_U]$.
Set
\begin{equation}
 \widehat\sigma_s^2=
 \Pi_{[\underline\sigma^2,\overline\sigma^2]}
 \left(\frac1{N_s^U}\sum_{t\in\mathcal I_s^U}
 \left[Z_t^{(2)}-\{\widehat m(X_t,s)-p_L\}^2\right]\right).
 \label{eq:scale-estimator}
\end{equation}
Replacing $m_*$ by $\widehat m$ in the squared correction changes
each summand by at most $2B\|\widehat m-m_*\|_\infty$. This
deterministic bound permits the mean and scale estimates to use the
same records. After calibration, freeze $\widehat m$ and
$\widehat\sigma_s=\sqrt{\widehat\sigma_s^2}$ for all subsequent
blocks.

The choice of regression routine determines the mean-estimation
accuracy, while the pricing module uses its estimates and error
radii. Following the general-utility construction of \citet{orbit},
we express the required regression guarantee separately from the
choice of estimator.

\begin{definition}[Bounded-response regression interface]
\label{def:regression-interface}
For each state $s$, the supplied routine has a deterministic
sample-size gate and a known uniform error radius
$r_{\mathcal M}(n,\varepsilon)$. Whenever the gate passes, for $n$
independent pairs $(X_i,Z_i)$ with $X_i\sim P_X$,
$Z_i\in[p_L,p_U]$ and
$\E[Z_i\mid X_i]=m_*(X_i,s)$, its projected output
$\widehat m_s:\cX\to[p_L,p_U]$ satisfies
\begin{equation}
 \Pp\!\left(\text{the routine succeeds and }
 \sup_{x\in\cX}|\widehat m_s(x)-m_*(x,s)|
 \le r_{\mathcal M}(n,\varepsilon)\right)\ge1-\varepsilon.
 \label{eq:regression-interface}
\end{equation}
Any reported empirical design failure counts toward the failure
probability $\varepsilon$.
\end{definition}

Appendix~\ref{app:regret-proofs} applies this fixed-sample contract
simultaneously over statewise calibration-visit prefixes, so that it
holds at the random retained count $N_s^U$.
Local-polynomial regression, least squares and sparse regression are
possible choices when their respective design conditions give this
guarantee. The subsequent pricing construction is the same for each
such choice.

\paragraph{Empirical Bellman initialization.}
Calibration supplies mean and scale estimates, but the module also
needs an initial shadow. We obtain it through Bellman iterations
based on randomized residual prices, whose construction does not
use a shadow estimate. Choose the supplied compact residual interval
$\cW$ from Appendix
\ref{app:domain-details}, let $w_1,\ldots,w_A$ be a grid in it, and
partition the fitted-mean
interval $[p_L,p_U]$ into $M$ bins, denoting the bin of $u$ by $I(u)$.
A policy selects one residual arm for each state and fitted-mean bin,
posting $\widehat m(x,s)+\widehat\sigma_s w_a$. Assumption
\ref{ass:implementation-domains}(i) gives feasibility and retention
of the oracle price for this family. For a fixed continuation vector
$h$, we use the finite policy class to approximate the maximum
expected one-step return $p_tY_t+h(S_{t+1})$.

Each initialization iteration uses one randomized residual-price
block. Begin with $v_0=0$ and freeze $v_j$ throughout iteration
$j=0,\ldots,J-1$. Choose the residual arm $A_t$ uniformly from $[A]$
and record
\[
 \mathcal R_t^j=p_tY_t+v_j(S_{t+1})-\min_s v_j(s).
\]
Subtracting the minimum continuation value makes this return
nonnegative and bounds its continuation component by the span of $v_j$.
Let $\mathcal D_{j,s}^{\rm w}$ index all visits to state $s$
in this block and put $N_{j,s}=|\mathcal D_{j,s}^{\rm w}|$.
After checking $N_{j,s}\ge n_{\rm w}$, score each
fitted-mean-bin/arm pair by
\begin{equation}
 \widehat Q_{j,s,i}(a)=
 \frac{A}{N_{j,s}}\sum_{t\in\mathcal D_{j,s}^{\rm w}}
 \one\{I(\widehat m(X_t,s))=i,\ A_t=a\}\mathcal R_t^j.
 \label{eq:warm-ips}
\end{equation}
The factor $A$ corrects for uniform arm sampling. Maximization
separates across fitted-mean bins. Adding back the subtracted baseline
gives the empirical Bellman value
\[
 \bar Z_j(s)=\min_u v_j(u)+\sum_{i=1}^M
                   \max_{a\in[A]}\widehat Q_{j,s,i}(a).
\]
Thus the update is computed from the randomized block by choosing
the best empirical arm in each bin. Set
\begin{equation}
 \begin{aligned}
 v_{j+1}&=\mathcal P_{H_V}\bar Z_j,\qquad
 \mathcal P_H v=\mathcal N C_Hv,\\
 (C_Hv)(s)&=\min\{v(s),\min_u v(u)+H\},\qquad
 \mathcal Nv=v-v(s_{\rm ref})\one.
 \end{aligned}
 \label{eq:span-projection}
\end{equation}
The projection keeps the continuation span at most $H_V$ and anchors
one coordinate. Both the scores and their maxima use the same
frozen $v_j$, so the randomized block supplies the update directly.

After $J$ iterations, retain
$h_{-1}=v_{J-1}$ and $h_0=v_J$. The same final randomized block also
provides the initial shadow $b_{-1}$: for each state and each purchase
outcome, average $v_{J-1}(S_{t+1})$ over all of that pair's observations,
and project the purchase-minus-nonpurchase difference into $\cB$.
The final block must supply at least $n_b$ observations per pair.
Both calculations use $h_{-1}(S_{t+1})$; the resulting
$h_0$ is first used in the active stage.
Appendix \ref{app:operational-details} specifies all grids, quotas
and the number of iterations from the supplied bounds.

\subsection{Updating continuation values and purchase shadows}
\label{sec:critic-update}

After initialization, the policy updates continuation values and
shadows in blocks of pricing calls. Block $k$ starts with $h_k$ and
the inherited shadow $b_{k-1}$, both fixed before the block begins.
The shadow enters the input and physical price conversion in
Section \ref{sec:policy}; the continuation vector determines the
returns used for the two updates. For each observed record, store
\begin{equation}
 Z_t^k=p_tY_t+h_k(S_{t+1}).
 \label{eq:observed-return}
\end{equation}
Conditioning on the current state, covariates and quote gives
\[
 \E[Z_t^k\mid S_t=s,X_t=x,p_t=p,\text{past}]
 =(P_0h_k)(s)+(p+b_s(h_k))D_s(p\mid x).
\]
The expected return is the Bellman objective evaluated at the posted
price, so averaging these returns evaluates the pricing decisions
actually made in the block.

For a complete block $\mathcal I_k$, let $N_k(s)$ count visits to
state $s$ and $N_{y,k}(s)$ count its outcome-$y$ observations.
Form the statewise return mean and outcome-conditional continuation
means
\begin{align}
 \bar Z_k(s)&=\frac1{N_k(s)}
       \sum_{\substack{t\in\mathcal I_k\\S_t=s}} Z_t^k,\notag\\
 \widehat\mu_{y,k}(s)&=\frac1{N_{y,k}(s)}
       \sum_{\substack{t\in\mathcal I_k\\S_t=s,\ Y_t=y}}
                 h_k(S_{t+1}).
 \label{eq:shared-block-means}
\end{align}
The block produces two outputs:
\begin{equation}
 h_{k+1}=\mathcal P_{H_V}\bar Z_k,\qquad
 b_k(s)=\Pi_{\cB}
       \{\widehat\mu_{1,k}(s)-\widehat\mu_{0,k}(s)\}.
 \label{eq:joint-update}
\end{equation}
The contrast estimates $b_s(h_k)$, so both outputs use the same
pre-block continuation vector. The next block starts with
$(h_{k+1},b_k)$ and continues the existing pricing module. This
timing leaves a one-block lag: the incoming shadow $b_{k-1}$ was
estimated from $h_{k-1}$, while current returns use $h_k$.
The block schedule below accounts for this lag.

\subsection{The complete policy}
\label{sec:unified-policy}

Algorithm \ref{alg:shape} combines initialization, pricing and the
block updates. Mean and scale fits remain frozen after calibration;
each complete active block produces the next continuation vector
and shadow. For these calls, the outer policy projects fitted means
into $[p_L,p_U]$, scales into
$[\underline\sigma,\overline\sigma]$ and shadows into $\cB$.
The following condition states the price-domain requirements for
initialization and active learning.

\begin{assumption}[Feasibility under estimated inputs]
\label{ass:implementation-domains}
There is a supplied radius $\eta_0>0$ such that the following hold
uniformly over states and covariates.
\begin{enumerate}[label=(\roman*),leftmargin=*]
\item When the mean and scale errors are at most $\eta_0$, every
price in the residual-price family used to construct the initialization
grid is feasible and has residual
in $\mathcal I$. This price set contains $p^*(x,s)$ with a fixed
interior margin.
\item If the shadow error is also at most $\eta_0$, every action
$q\in\cQ$ produces a feasible price
$\widehat\sigma_s q-\widehat b_s$ with residual in $\mathcal I$.
Its fitted input belongs to $\mathcal Z$, and its transformed oracle
action remains inside $\cQ$ with a known margin $\rho_Q>0$, as
specified in Appendix \ref{app:domain-details}.
\item The coverage threshold satisfies
$\kappa_0\le\underline\sigma^2/(p_U-p_L)^2$.
\end{enumerate}
\end{assumption}

Parts (i) and (ii) require positive price-domain margins: small input
errors must neither remove the oracle price from the exploration
family nor move a quoted price outside $\cP$. Projection keeps each
estimate in its own range, but does not ensure these joint margins.
Compared with ORBIT's scalar pilot condition, we also account for
scale and shadow errors when converting actions back to prices.
Part (iii) places the coverage threshold within the average purchase
overlap provided by uniform-price calibration; it does not require
overlap at each individual randomized quote. Appendix
\ref{app:domain-details} gives the precise domains and margins.

The policy is supplied with the horizon, the mean-estimation routine
and conservative bounds for scales, domains, smoothness, coverage
and revenue geometry. Active blocks have lengths
$L_k=\lceil L_0\varrho^k\rceil$ for a supplied growth factor
$\varrho>1$. Longer blocks improve the continuation estimates;
their growth is chosen to control the one-block lag in the shadow
update. Counts are checked at each completed block, and every quote
is checked for feasibility before it is posted.
Appendix \ref{app:operational-details} specifies $L_0$, $\varrho$,
initialization precision and all refinement settings from the
declared bounds.

\begin{algorithm}[H]
\caption{Shadow-Adjusted Pricing in Evolving Systems (\SHAPE{})}
\label{alg:shape}
\small
\begin{algorithmic}[1]
\Require $T,n_u,\delta$, a mean-estimation routine and the structural
settings in Appendix \ref{app:operational-details}.
\State Check deterministic budgets and precision gates. On failure,
use $p_{\rm fb}=p_L$ for the entire horizon and stop.
\State Collect $n_u$ uniform-price records; fit and freeze
$(\widehat m,\widehat\sigma)$ using all records from each state,
checking empirical design and minimum counts.
\State Set $v_0=0$. Run $J$ empirical Bellman initialization blocks
from Section \ref{sec:calibration}.
\State Retain $h_{-1}=v_{J-1}$, $h_0=v_J$, and $b_{-1}$ estimated
using $v_{J-1}$ in the final initialization block.
\State Initialize the pooled pricing module once with global budget $T$.
\For{$k=0,1,\ldots$ while periods remain}
 \State Freeze $(h_k,b_{k-1})$ for the continuation updates and
 provide $b_{k-1}$ to the existing pricing module without resetting it.
 \State Let $N=\min\{L_k,\text{remaining periods}\}$.
 \For{the next $N$ periods}
  \State Observe $(S_t,X_t)$ and compute \eqref{eq:actor-input}
  using $b_{k-1}$. Call Algorithm \ref{alg:pooled-module} for $q_t$;
  on failure, use fallback and stop learning. Otherwise set
  $p_t=\widehat\sigma_{S_t}q_t-b_{k-1}(S_t)$.
  \State If $p_t\notin\cP$, use fallback and stop learning.
  Otherwise post $p_t$ once and observe $(Y_t,S_{t+1})$.
  \State Complete the feedback update in Algorithm
  \ref{alg:pooled-module} using $Y_t$, and store
  \eqref{eq:observed-return}.
 \EndFor
 \If{$N=L_k$}
  \State If a state/outcome count check fails, use fallback and stop
  learning. Otherwise compute \eqref{eq:joint-update} using the same
  frozen $h_k$.
 \Else
  \State Stop without a continuation or shadow update.
 \EndIf
\EndFor
\end{algorithmic}
\end{algorithm}

Initialization count shortfalls may first use the bounded recovery rule
in Appendix \ref{app:operational-details}. All other failures follow
the same convention: before posting an invalid quote, or after a failed
design, count or numerical check, switch to $p_{\rm fb}$ for all
remaining periods and perform no further updates. Every initialization period counts toward the
horizon; the minimum count gates do not discard later records. A
final partial active block continues to quote prices and return
purchase feedback to the module, but does not update the
continuation vector or shadow.

\FloatBarrier

\section{Regret guarantees and a linear benchmark}
\label{sec:regret}

The regret analysis must account for the effect of prices on both
current revenue and future system states. We first give an overall
guarantee that separates learning the common pricing map from
calibrating mean valuations and state scales. We then relate regret
to pricing loss on the learner's state path and show how the joint
value and shadow updates control this loss. A linear benchmark pairs
an explicit calibration routine with a matching time-exponent lower
bound; Section~\ref{sec:valuation-models} applies the same guarantee
to nonlinear and sparse mean valuations.

\subsection{Overall regret guarantee}
\label{sec:overall-regret}

For fixed $K$, $\beta$ and structural bounds, the main statistical
costs are learning the shared price map, with time power
$T^\alpha$ for $\alpha=(2\beta-1)/(4\beta-3)$, and accumulating
squared mean and scale errors over the horizon. After accounting
for these calibration errors, value learning contributes the
smaller time power $T^{2\alpha-1}$. The bound also counts all
calibration and initialization periods and the cost of failures.

Write $\eta^2=\eta_m^2+\eta_\sigma^2$ for the squared calibration
radii in Appendix~\ref{app:operational-details}.
Let $\tau_0=n_u+\tau_{\rm w}+L_0$ be the nominal startup
budget, including calibration, Bellman initialization and one
initial active block. We use
$\ell_T=\log\{64K(T+1)^2/\delta\}$.
The constants below may depend on fixed $K,\beta$, the structural
envelopes and certified module constants, including the coverage
margins and fixed admissible initialization, block-growth and
action-neighborhood settings specified in
Appendix~\ref{app:operational-details}, but not on $T,n_u$ or $\delta$.
These structural choices are fixed independently of the horizon
and confidence level.

\begin{theorem}[Regret of \SHAPE{}]
\label{thm:conditional-master}
Assume the model and Assumptions~\ref{ass:tail-regularity}--
\ref{ass:implementation-domains}.
Run Algorithm \ref{alg:shape} with a mean-estimation routine satisfying
Definition \ref{def:regression-interface}, the copy settings and certified
optimization model of Proposition~\ref{prop:pooled-pricing}, and suppose the
deterministic gates in Appendix~\ref{app:operational-details} pass.
Then there is a constant $C$ such
that, uniformly over the initial state,
\begin{align}
 \Reg_T\le C\big\{&
 1+\tau_0+\ell_T^{17}T^\alpha+T\eta^2
 +\ell_T^{34}T^{2\alpha-1}\log(eT)\notag\\
 &+\ell_T^2\log(eT)+T\eta^4+T\delta\big\}.
 \label{eq:conditional-master}
\end{align}
If a deterministic startup gate fails, the stated fallback policy
instead has the unconditional bound $\Reg_T\le p_UT$.
\end{theorem}

In \eqref{eq:conditional-master}, $T^\alpha$ is the map-learning
power and $T\eta^2$ is the cost of mean and scale errors.
The value-update term $T^{2\alpha-1}$ has a smaller exponent,
since $1/2<\alpha\le3/5$. Keeping one pricing history avoids
repeating the grid search at each value update, although the global
input partition requires an initial active block of length
$L_0=\widetilde O(T^{2\alpha-1})$ at polynomial confidence.
This cost is included in $\tau_0$, while $T\delta$ charges failure
events in the expected-revenue bound. The following corollary collects
these terms and makes the calibration tradeoff explicit. Write $M_T$
for the number of input bins in the shared pricing module.

\begin{corollary}[Regret through the calibration interface]
\label{cor:regression-transfer}
Under the conditions of Theorem~\ref{thm:conditional-master}, take
$\delta=T^{-3}$, write $\varepsilon_T=T^{-4}/(8K)$ for the
per-state, per-prefix fitting failure level, and keep $K$ and the structural
quantities fixed.
For budgets satisfying the deterministic gates, with
$N_U=\lfloor q_S(n_u-1)/2\rfloor$, use the prescribed initialization
minimum counts and the smallest $L_0\ge\max\{2,M_T\}$ satisfying
\eqref{eq:block-gates}. Put
$\bar r_{\mathcal M}=\sup_{N_U\le n\le n_u}r_{\mathcal M}(n,\varepsilon_T)$.
The policy then satisfies
\begin{equation}
 \Reg_T=\widetilde O\!\left(
 n_u+T^\alpha+
 T\bar r_{\mathcal M}^2
 +\frac{T}{N_U}\right).
 \label{eq:regression-transfer}
\end{equation}
The bound is uniform over the initial system state.
\end{corollary}

The $n_u$ calibration periods incur an initial cost; a more accurate
fit reduces the subsequent cost
$T\bar r_{\mathcal M}^2$. For a nonincreasing radius this is
$T r_{\mathcal M}(N_U,T^{-4}/(8K))^2$. The additional $T/N_U$
comes from estimating the state scales after correcting for the
fitted means. In particular, the deterministic correction bound on
the shared calibration records gives
$\eta^2\le C\{\bar r_{\mathcal M}^2+\ell_T/N_U\}$.
The precision gates also give $\eta^4\le C\eta^2$.
The nominal Bellman-initialization budget $\tau_{\rm w}$ is polylogarithmic,
and $L_0=\widetilde O(T^{2\alpha-1})$ is dominated by the map-learning
term. Substitution into Theorem~\ref{thm:conditional-master} proves
\eqref{eq:regression-transfer}. Thus a change in the valuation class
changes the calibration cost, while the pricing and value-update
architecture remains the same.

\subsection{Pricing loss along the evolving state path}
\label{sec:finite-horizon-bridge}

To connect regret with the pricing module, we compare current
revenue together with the continuation value after a purchase or
nonpurchase. The true stationary shadow incorporates this future
effect into the conditional decision loss
\begin{equation}
 \Delta^*(x,s,p)=
 \max_{a\in\cP}(a+b_s^*)D_s(a\mid x)
 -(p+b_s^*)D_s(p\mid x).
 \label{eq:true-bellman-gap}
\end{equation}
Assumption~\ref{ass:utility-domain} gives the maximizer over the full
physical interval, so $0\le\Delta^*(x,s,p)\le R:=p_U+H_V$ for every
feasible price. The bound follows by writing the objective as current
revenue plus expected next-state value: the first lies in $[0,p_U]$
and the second lies between $\min h^*$ and $\max h^*$.

\begin{proposition}[Finite-horizon comparison]
\label{prop:benchmark-bridge}
Under Assumptions~\ref{ass:tail-regularity}, \ref{ass:demand-geometry},
\ref{ass:coverage} and \ref{ass:utility-domain}, every
admissible policy satisfies, uniformly over its initial state,
\begin{equation}
 \left|\Reg_T(\pi;s)-
 \E_s^\pi\sum_{t=1}^T\Delta^*(X_t,S_t,p_t)\right|\le H_V.
 \label{eq:regret-gap-bridge}
\end{equation}
\end{proposition}

The comparison evaluates conditional decision loss on the learner's
own state path, while the two full-horizon revenues follow their
respective policy-induced paths.
If $P_0=P_1$, those state distributions are independent of prices.
The two policies then have the same terminal-state distribution, and
the comparison is exact without the $H_V$ boundary term.

The comparison uses the true shadow $b^*$, whereas the policy
prices with an estimate inherited from the preceding block.
To measure map learning at the supplied shadow, write $\widehat b_t$
for that input on active call $t$ and define the comparison price
\[
 p_t^\dagger=\sigma_{S_t}\psi\!\left(
 \frac{m_*(X_t,S_t)+\widehat b_{t,S_t}}{\sigma_{S_t}}\right)
 -\widehat b_{t,S_t}.
\]
The corresponding adjusted pricing loss is
\begin{equation}
 \Gamma_t=2\left\{
 (p_t^\dagger+\widehat b_{t,S_t})D_{S_t}(p_t^\dagger\mid X_t)
 -(p_t+\widehat b_{t,S_t})D_{S_t}(p_t\mid X_t)\right\}.
 \label{eq:adjusted-pricing-loss}
\end{equation}
The local domain contract makes $p_t^\dagger$ feasible, and scalar
quadratic growth gives $\Gamma_t\ge0$.
Here $\Delta_t^*=\Delta^*(X_t,S_t,p_t)$ uses the true shadow $b^*$,
whereas $\Gamma_t$ uses the inherited shadow. We allocate failure
probability $\delta_A=\delta/16$ to the pricing module.

\begin{proposition}[Persistent pooled pricing]
\label{prop:pooled-pricing}
Under Assumptions~\ref{ass:tail-regularity}, \ref{ass:demand-geometry},
\ref{ass:utility-domain} and \ref{ass:implementation-domains}, the
single pricing module of Section~\ref{sec:policy}, with the precision gates
in Appendix~\ref{app:operational-details} and the certified
convex-bandit implementation in Appendix~\ref{app:predictable-bandit},
has the following guarantee. Condition on its initialization with
valid frozen fits. For a predictable shadow process in $\cB^K$,
constant within each outer block and satisfying
$\|\widehat b_t-b^*\|_\infty\le\eta_0$, with probability at
least $1-\delta_A$, it produces valid
quotes without a failure flag and, simultaneously for all active
prefixes $1\le N\le T$,
\begin{equation}
 \sum_{t=1}^N\Gamma_t\le\Phi_T(N)+C_pN\eta^2,\qquad
 \Delta_t^*\le\Gamma_t+C_p|\widehat b_{t,S_t}-b_{S_t}^*|^2.
 \label{eq:pricing-certificate}
\end{equation}
Here $\Phi_T$ is specified in \eqref{eq:pricing-map-envelope}, with
the structural constant in \eqref{eq:pricing-map-constant}.
The stopped guarantee holds before the first local-shadow exit,
without conditioning on eventual local success.
\end{proposition}

The envelope $\Phi_T(N)$, with $T$ fixed at initialization and $N$
counting active calls, collects the one-time grid search, bandit
refinement and polynomial approximation costs; under the prescribed
tuning, $\Phi_T(T)=\widetilde O(T^\alpha)$.
For the pricing module, the predictable-loss analysis in
Appendices~\ref{app:predictable-bandit}
and \ref{app:pricing-certificate-proof} permits shadows to change
with earlier observations and requires no positive input-bin
probability or covariate Gram bound. Nonnegativity of $\Gamma_t$
bounds its sum within a block by the global prefix loss through
that block, supplying the bound needed by the continuation update.
The second inequality in \eqref{eq:pricing-certificate} leaves the
squared shadow error to control. We next show that the joint
updates keep this error small while reusing the pricing records.

\subsection{Joint value learning and the proof argument}
\label{sec:regret-proof-argument}

Proposition~\ref{prop:pooled-pricing} controls pricing loss while
the supplied shadow is close to $b^*$. Maintaining this accuracy
requires controlling value and pricing errors together: current
prices affect the purchase and next-state observations used to
update both the continuation vector and shadow. The joint-block
bound below expresses this dependence through the pricing loss;
combining it with the pricing certificate gives a stable recursion
with a one-block lag.

Calibration and randomized Bellman initialization provide the
starting precision for this recursion. The moment identities in
\eqref{eq:calibration-moments} control the mean and scale estimates
on the same statewise records. Initialization then uses a fixed
residual-price family containing the stationary oracle price.
Its Bellman operator contracts span, so the iterations reach the
required local precision without accurate intermediate shadows.
Proposition~\ref{prop:warm-start} in Appendix~\ref{app:regret-proofs}
gives the corresponding finite-sample guarantee.

For a complete active block, write $e_k=\spn(h_k-h^*)$,
$A_k=\sum_{t\in\mathcal I_k}\Delta_t^*$ and
$N_{k,\min}=\min_sN_k(s)$.
Lemma~\ref{lem:joint-block} gives, on its count and concentration
event,
\[
 e_{k+1}\le\vartheta e_k+\frac{A_k}{N_{k,\min}}+r_v(L_k),
 \qquad
 \|b_k-b^*\|_\infty\le e_k+r_b(L_k),
 \qquad \vartheta=1-a_0.
\]
Here $r_v$ and $r_b$ are the sampling radii in
Appendix~\ref{app:operational-details}. The value error contracts
up to the displayed pricing-loss and sampling terms, while
the shadow estimate inherits the error in $h_k$. Both updates use
this fixed pre-block vector. Their joint control follows from
chronological martingales on the same records, without splitting
those records into separate update samples.

The pricing guarantee bounds $A_k$ by map-learning loss,
calibration error and the squared error of the inherited shadow.
Since that shadow was obtained from $h_{k-1}$, substitution gives
the two-step recursion
\[
 e_{k+1}\le\vartheta e_k+ce_{k-1}^2+u_k,
 \qquad c=8C_p/q_S,
\]
where \eqref{eq:delayed-recursion} specifies the pricing and sampling
contributions in $u_k$. The linear contraction acts on the current
value error, whereas the squared term comes from the preceding
value error through the inherited shadow. Starting from the
initialization neighborhood, the block schedule yields contraction
of a weighted sum of successive errors, up to the pricing and
sampling contributions in $u_k$.
Proposition~\ref{prop:delayed-stability} proves that the shadows
remain in the local domain, so the stopped pricing guarantee
continues to apply throughout active learning.

The same weighted estimate controls the sum of squared shadow
errors over the geometrically growing blocks. Applying the pricing
prefix bound once to all active calls then controls total decision
loss. A terminal partial block contributes pricing loss without
requiring another value update. Adding
calibration, initialization, failure-event costs and the
finite-horizon boundary term gives Theorem~\ref{thm:conditional-master}. Appendix~\ref{app:regret-proofs}
contains this aggregation and the auxiliary finite-sample
statements.

\subsection{Linear mean valuations}
\label{sec:linear-valuations}

A linear mean valuation provides a concrete benchmark for the
general guarantee: least squares supplies an explicit calibration
routine, and a matching lower bound determines the time exponent
for fixed dimension and state count. We retain the unknown state scales and purchase-driven
dynamics of the model and specialize only the mean function.

Let $\phi:\cX\to\R^D$ be a known feature vector, with
$\sup_x\|\phi(x)\|_\infty\le L_\infty$ and
$\sup_x\|\phi(x)\|_2\le L_2$.
Consider state-specific coefficients satisfying
\begin{equation}
 m_*(x,s)=p_L+\phi(x)^\top\theta_s,\qquad
 0\le\phi(x)^\top\theta_s\le B.
 \label{eq:linear-valuation}
\end{equation}
For affine mean valuations, take
$\phi(x)=(1,x^\top)^\top$ and $D=d+1$.
The constant coordinate represents the unknown intercept.
The following condition supports least squares on the fixed
calibration sample.

\begin{assumption}[Linear calibration design]
\label{ass:linear-calibration-design}
The common feature Gram matrix satisfies
$G_\phi=\E\{\phi(X)\phi(X)^\top\}\succeq\lambda_\phi I_D$
for a supplied $\lambda_\phi>0$.
\end{assumption}

For each state, fit the shifted response $Z^{(1)}-p_L$ by least
squares and clip the resulting mean function.
The routine checks that the empirical Gram matrix has smallest
eigenvalue at least $\lambda_\phi/2$.
Lemma~\ref{lem:bounded-ols} includes this design failure in the
confidence budget and proves
\begin{equation}
 r_{\rm L}(n,\varepsilon)
 =B\wedge
 \frac{2BL_2L_\infty}{\lambda_\phi}
       \sqrt{\frac{2D\log(4D/\varepsilon)}n}.
 \label{eq:ols-radius-main}
\end{equation}
The population Gram condition is a sufficient condition for this
uniform frozen fit. In contrast, ORBIT's adaptive linear pilot
certifies the currently arriving feature direction without a
population Gram bound \citep{orbit}. Here the fixed calibration
sample supplies both the mean and state-scale estimates used
throughout subsequent value updates.

\begin{corollary}[Linear mean valuations]
\label{cor:linear-valuation}
Under the common conditions of Corollary~\ref{cor:regression-transfer}, suppose
\eqref{eq:linear-valuation} and Assumption~\ref{ass:linear-calibration-design} hold.
Use the least-squares routine of Lemma~\ref{lem:bounded-ols}
and its calibration budget in \eqref{eq:parametric-calibration-choice}.
For fixed $K,D,\beta$, feature envelopes and design bounds,
the deterministic gates pass for sufficiently large $T$, and
\[
 \Reg_T=\widetilde O(T^\alpha+\sqrt T),
 \qquad \alpha=\frac{2\beta-1}{4\beta-3}.
\]
The bound is uniform over the initial state and common class bounds.
\end{corollary}

To assess the time dependence of this guarantee, we use the
linear-model lower bound of \citet{orbit}.
The embedding below keeps the full physical price interval and
preserves valuations when standardizing the noise. It also places
the hard instances inside a single class with the preceding
Gram lower bound.

\begin{theorem}[A fixed-dimension lower bound]
\label{thm:linear-map-lower}
Fix $K,d\ge1$ and $\beta\ge2$, and take the physical price interval
to be $[0,1]$. There are horizon-independent structural envelopes for
Assumptions~\ref{ass:tail-regularity}--\ref{ass:implementation-domains}
and a constant $\lambda_\phi>0$ with the following
property. Let $\mathfrak M_{\rm lin}$ be the class of models within these
envelopes having statewise affine mean valuations and
$\lambda_{\min}\E[\phi(X)\phi(X)^\top]\ge\lambda_\phi$, where
$\phi(x)=(1,x_1,\ldots,x_d)^\top$. For every fixed initial state and all
sufficiently large $T$,
\[
 \inf_\pi\sup_{M\in\mathfrak M_{\rm lin}}\Reg_T(\pi;M)
 \ge c_\beta T^{(2\beta-1)/(4\beta-3)}.
\]
The infimum includes all randomized policies observing the information
in Section~\ref{sec:model}, without restricting their prices to the
algorithm's exploration or standardized-action grids.
\end{theorem}

Corollary~\ref{cor:linear-valuation} and Theorem~\ref{thm:linear-map-lower} therefore give, for the same fixed-dimensional
class with these common envelopes,
\begin{equation}
 \inf_\pi\sup_{M\in\mathfrak M_{\rm lin}}\Reg_T(\pi;M)
   =\widetilde\Theta(T^\alpha).
 \label{eq:linear-minimax}
\end{equation}
The hard family has purchase-independent transitions, so the lower
bound reflects the intrinsic difficulty of learning the common price
map. The upper bound also accommodates purchase-driven state changes
within the maintained class. This comparison fixes $d,K$ and the
structural bounds; the design dependence of the calibration routine
remains explicit in \eqref{eq:ols-radius-main}.
Appendix~\ref{app:valuation-models} proves the same-class embedding.

\section{Nonlinear and sparse valuation models}
\label{sec:valuation-models}

The calibration interface also accommodates nonlinear covariate
effects and high-dimensional sparse features. In each case, we
specify the mean class, the design condition needed to estimate it
and the resulting regret bound. Only the calibration routine and
its budget change; the shared pricing module, scale estimation and
joint continuation updates remain the same. Throughout this
section, we use $\delta=T^{-3}$, the per-state failure level
$\varepsilon_T=T^{-4}/(8K)$ for each state and possible prefix,
and the initialization and starting-block
choices of Corollary~\ref{cor:regression-transfer}.

\subsection{Nonlinear mean valuations}
\label{sec:holder-mean-specialization}

To allow nonlinear responses to covariates within each state,
consider $\cX=[0,1]^d$ and assume that
$u_s(x)=m_*(x,s)-p_L$ belongs to a multivariate $\gamma$-H\"older
class with common radius $L_m$, where $\gamma>0$ and
$0\le u_s\le B=p_U-p_L$. The mean smoothness $\gamma$ and the
noise-tail smoothness $\beta$ describe different functions and
need not coincide. We use the precise Taylor convention in
Appendix~\ref{app:holder-cell-regression}, including its integer-order
definition. Uniform calibration also requires observations
throughout the covariate domain.

\begin{assumption}[Covariate density for nonlinear calibration]
\label{ass:calibration-density}
The common covariate distribution has a density $f_X$ on $[0,1]^d$
with supplied bounds
$0<f_{\min}\le f_X(x)\le f_{\max}<\infty$ almost everywhere.
\end{assumption}

Under this condition, cellwise polynomial least squares supplies
the required mean estimate. For each state's retained calibration
sample, fit degree-$j$ polynomials, where $j=\lceil\gamma\rceil-1$,
and clip the fitted mean to $[p_L,p_U]$.
With the empirical design check specified in
Lemma~\ref{lem:holder-cell-regression}, the estimator has uniform
error radius
\begin{equation}
 r_{\rm H}(n,\varepsilon)
 =B\wedge C_{\rm H}
 \left\{\frac{\log(4D_{\rm H}^2n/\varepsilon)}n\right\}^{
              \gamma/(2\gamma+d)},
 \qquad D_{\rm H}=\binom{d+j}{d}.
 \label{eq:holder-regression-radius}
\end{equation}
The constant and deterministic sample gate depend only on the
displayed density, smoothness and value bounds. This density
condition is needed for the mean fit, while the pricing learner
continues to operate on the scalar input without a bin-coverage
condition. Substitution into Corollary~\ref{cor:regression-transfer}
gives the following guarantee.

\begin{corollary}[Nonlinear mean valuations]
\label{cor:holder-valuation}
Retain the model, structural conditions and certified pricing
implementation of Theorem~\ref{thm:conditional-master}.
Suppose that the statewise means satisfy
\eqref{eq:holder-mean-taylor} and Assumption~\ref{ass:calibration-density} holds.
Use the cellwise polynomial estimator in calibration and take
$n_u\asymp T^{(2\gamma+d)/(4\gamma+d)}$.
For fixed $K,d,\beta,\gamma$ and common structural bounds,
all deterministic gates pass for sufficiently large $T$, and
\begin{equation}
 \Reg_T=\widetilde O\!\left(
 T^{(2\beta-1)/(4\beta-3)}
 +T^{(2\gamma+d)/(4\gamma+d)}\right).
 \label{eq:holder-pilot-consequence}
\end{equation}
The bound is uniform over the initial state and the stated class.
\end{corollary}

The first term is the cost of learning the common price map; the
second balances calibration periods against squared mean-estimation
error during pricing. State-dependent scale estimation uses the same
records and has squared error of order
$r_{\rm H}(N_U,\varepsilon_T)^2+\ell_T/N_U$.
Consequently it introduces no larger time power in this bound.
Appendix~\ref{app:valuation-models} verifies the estimator interface,
statewise sample selection and the calibration balance.

\subsection{Sparse linear mean valuations}
\label{sec:sparse-valuations}

When the feature vector is large but only a few coordinates affect
mean valuation, retain \eqref{eq:linear-valuation} and assume
$J_s=\operatorname{supp}(\theta_s)$ has size at most a supplied $s_0$.
Sparsity allows least squares to be replaced by an $\ell_1$-penalized
fit. Its design requirement concerns the directions in which a
sparse coefficient vector can be estimated.

\begin{assumption}[Sparse calibration design]
\label{ass:sparse-calibration-design}
For a supplied $\kappa_{\rm c}>0$, each nonempty true support $J_s$
satisfies
\begin{equation}
 v^\top G_\phi v\ge
 \frac{\kappa_{\rm c}^2}{|J_s|}\|v_{J_s}\|_1^2
 \quad\text{whenever }\|v_{J_s^c}\|_1\le3\|v_{J_s}\|_1.
 \label{eq:sparse-compatibility}
\end{equation}
The learner is supplied with $s_0$ and the compatibility lower
bound, not with the supports.
\end{assumption}

The full Gram bound implies this condition with
$\kappa_{\rm c}^2=\lambda_\phi$; the sparse routine only needs
the stated cone bound. It is the compatibility condition used by
ORBIT's sparse extension, restricted here to the true supports
needed in the proof. For each state, use all $N_s^U$ calibration
observations and minimize
\[
 \frac1{2N_s^U}\sum_{i=1}^{N_s^U}
       \{Z_i^{(1)}-p_L-\phi(X_i)^\top\theta\}^2
 +\lambda_{{\rm pen},s}\|\theta\|_1,
 \qquad
 \lambda_{{\rm pen},s}=2BL_\infty
       \sqrt{\frac{2\log(4D/\varepsilon_T)}{N_s^U}},
\]
and clip the fitted mean to $[p_L,p_U]$.
All coefficients, including an intercept coordinate, are penalized
and counted in $s_0$.
Lemma~\ref{lem:bounded-lasso} proves a uniform error radius
\begin{equation}
 r_{\rm S}(n,\varepsilon)
 =B\wedge
 \frac{48BL_\infty^2s_0}{\kappa_{\rm c}^2}
       \sqrt{\frac{2\log(4D/\varepsilon)}n}.
 \label{eq:lasso-radius-main}
\end{equation}
A deterministic sample-size gate makes empirical compatibility a
high-probability proof event; the policy need not test compatibility
over unknown supports.

\begin{corollary}[Sparse mean valuations]
\label{cor:sparse-valuation}
Retain the common conditions of Corollary~\ref{cor:regression-transfer}, the representation
\eqref{eq:linear-valuation}, the feature bound $L_\infty$ and
Assumption~\ref{ass:sparse-calibration-design}.
Use the Lasso routine and its calibration budget in
\eqref{eq:parametric-calibration-choice}.
Whenever the remaining startup and active-block gates pass, with
fixed $K,\beta$, value, feature and dynamic-system bounds,
\begin{equation}
 \Reg_T=\widetilde O\!\left(
 T^\alpha+\frac{s_0}{\kappa_{\rm c}^2}\sqrt T
                +\frac{s_0^2}{\kappa_{\rm c}^4}\right).
 \label{eq:sparse-valuation-regret}
\end{equation}
The hidden constant is independent of $D,s_0$ and $\kappa_{\rm c}$;
the logarithmic factors may depend on $D$.
For fixed $D,s_0,\kappa_{\rm c}$ the gates pass for all sufficiently
large $T$.
\end{corollary}

The linear dependence on $s_0$ in the leading calibration cost comes
from controlling the fitted mean uniformly over covariates:
bounded features convert an $\ell_1$ coefficient error into
\eqref{eq:lasso-radius-main}.
The additional squared-sparsity term retains the finite-sample
design and precision floors.
When $s_0/\kappa_{\rm c}^2\le\sqrt T$, it is absorbed by the
calibration term proportional to $\sqrt T$.
The dimension enters through the regression problem, while the
pricing learner continues to use a scalar input.

\section{Experiments}
\label{sec:experiments}

We evaluate online pricing when purchases add to the system's workload
and the resulting state affects subsequent customers' valuations.
We compare \SHAPE\ with state-aware myopic pricing, dynamic plug-in
pricing and a state-blind ORBIT adaptation in this workload system.

\subsection{Experimental setup}
\label{sec:experiment-setup}

We consider $K=4$ workload bands, ordered from least to most loaded,
and independent covariates $X_t\sim\mathrm{Unif}[0,1]$. The mean valuation
and state scales are
\[
 \begin{aligned}
 m_*(x,s)&=\mu_s+0.2\sin(2\pi x),
 &(\mu_1,\mu_2,\mu_3,\mu_4)&=(4.7,3.9,3.1,2.3),\\
 &&(\sigma_1,\sigma_2,\sigma_3,\sigma_4)&=(1.6,1.3,1.05,0.8).
 \end{aligned}
\]
Higher workload lowers mean valuation, while the largest and smallest
noise scales differ by a factor of two. The common valuation noise has
density
\[
 f_\xi(w)=\frac{3}{5L_\xi}
 \left(1-\frac{w^2}{2L_\xi^2}\right)\one\{|w|\le L_\xi\},
 \qquad L_\xi=\frac5{\sqrt7},
\]
which has mean zero and variance one. The physical price interval is
$[0,8]$, containing the valuation support in every state. The true mean
function, state scales, noise distribution and transition kernels are
unknown to all methods.

Without a purchase, the system can recover to the next lower workload
band. A purchase can instead add one band of workload, replacing that
opportunity for recovery. To specify these transitions, let
\[
 \begin{aligned}
 Q_r(s'\mid s)
   &=(1-r)\one\{s'=s\}+r\one\{s'=\max(1,s-1)\},\\
 P_0(s'\mid s)&=0.05/4+0.95Q_r(s'\mid s),\\
 P_1(s'\mid s)&=0.05/4+
   0.95\bigl[(1-\theta)Q_r(s'\mid s)
          +\theta\one\{s'=\min(4,s+1)\}\bigr].
 \end{aligned}
\]
The common refresh component allows changes unrelated to the purchase
outcome. The displayed experiment sets purchase pressure at $\theta=0.6$
and recovery probability at $r=0.6$.\footnote{This section presents one
regime from the completed two-regime evaluation.}
Every run starts from $S_1=1$ and lasts $T=48{,}000$ periods.

We compare \SHAPE\ with three feasible methods.
\emph{State-aware myopic} uses state-specific mean and scale estimates
but sets the purchase shadow to zero.
\emph{Dynamic plug-in} estimates the common noise tail from purchase
observations and maximizes estimated shadow-adjusted revenue over a
physical-price grid. Both methods observe the system state.
The \emph{state-blind ORBIT adaptation}, based on the general-utility
explore-then-price-map construction of \citet{orbit}, fits a pooled
utility pilot and prices without state labels or transition observations.
The comparisons therefore assess state-blind pricing, state-aware
myopic pricing and two approaches to learning dynamic quotes.
Each method generates its own purchases and state path.

All methods begin with 3,328 uniform-price calibration periods.
The state-aware methods use the pseudo-response $8Y_t$ to fit separate
linear mean models within two covariate cells in each state, estimating
the state scales from the same records as in
Section~\ref{sec:calibration}. \SHAPE\ then uses 2,560 periods for
dynamic initialization, whereas plug-in uses a nominal 768 periods;
myopic and state-blind pricing do not require this stage.
Initialization blocks can extend when their observation-count gates
are not met. These extensions, as well as all exploration and fallback
periods, count toward the horizon.

As in the numerical implementation of \citet{orbit}, local price-map
refinement uses projected zeroth-order gradient updates rather than the
theoretical online-Newton subroutine. \SHAPE\ and myopic pricing use
local linear price rules on four input bins, a trust radius of $0.3$,
and respectively 7 and 10 coarse-grid actions, with 100 observations
per action in each bin. The state-blind implementation uses one input
bin and 25 physical-price grid actions. Plug-in uses 81 noise-tail bins
and a grid of 481 physical prices, refitting the tail every 256 active
observations. Each method has one numerical configuration, fixed
before the 200 evaluation replications.

We use 200 new independent Monte Carlo replications per method, with
common exogenous random draws across methods within each
replication. Regret is evaluated against the finite-horizon dynamic
oracle of Section~\ref{sec:oracle-regret}, using continuous-price
maximization and 32-point Gauss--Legendre integration over the covariate.
We evaluate each quote through its conditional expected revenue and
average
$\mathcal V_t^*(1)-\sum_{\tau=1}^t
p_\tau D_{S_\tau}(p_\tau\mid X_\tau)$ over replications.
Each policy is configured once for $T=48{,}000$; its curve records
performance over elapsed periods $t$, not separate runs tuned for each
shorter horizon.

In addition to absolute regret, we plot the regret of each state-aware
comparator minus that of \SHAPE. Positive differences favor \SHAPE.
We first subtract the two regret paths within each replication, then
compute the mean and its Monte Carlo standard error. Shaded bands give
pointwise 95\% Student-$t$ intervals; the terminal comparisons use the
same paired replications. All 200 replications are retained for every
method in the displayed experiment.

\FloatBarrier

\subsection{Pricing performance under purchase-driven transitions}
\label{sec:experiment-results}

\begin{figure}[!htbp]
 \centering
 \includegraphics[width=.8\linewidth]{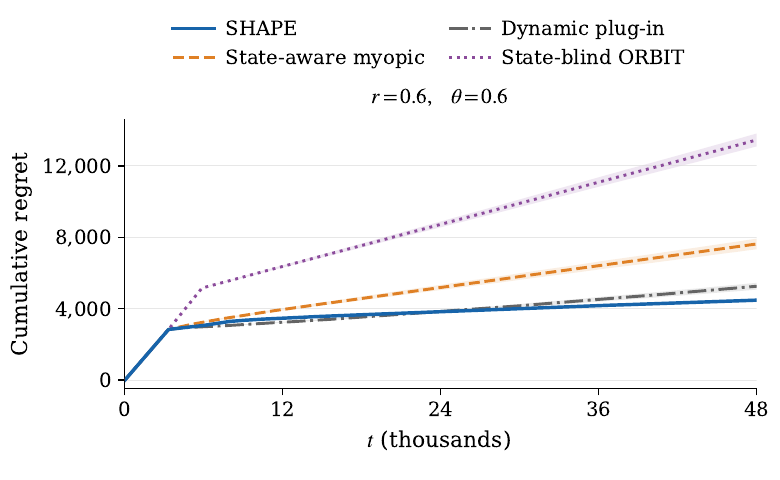}
 \caption{Cumulative regret: means and pointwise 95\% intervals
 over 200 replications.}
 \label{fig:workload-regret}
\end{figure}

Figure~\ref{fig:workload-regret} shows the cumulative regret trajectories,
and Table~\ref{tab:workload-regret} reports terminal means and paired
comparisons. \SHAPE\ has the lowest mean regret in this system.
Its terminal mean regret is 4,480.74, compared with
7,627.40 for state-aware myopic pricing, a reduction of 41.25\%.
The paired advantage is 3,146.67, with 95\% interval
$[2{,}848.44,3{,}444.89]$. Both methods use the current state and the same
calibration data, but only \SHAPE\ incorporates continuation values into
its price-map inputs.

\begin{figure}[!htbp]
 \centering
 \includegraphics[width=.8\linewidth]{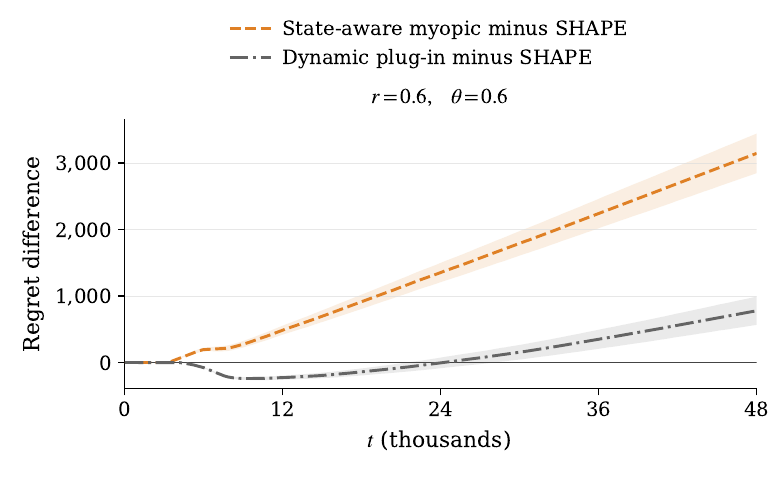}
 \caption{Comparator regret minus \SHAPE\ regret, with paired
 pointwise 95\% intervals; positive values favor \SHAPE.}
 \label{fig:workload-regret-gap}
\end{figure}

The difference curves in Figure~\ref{fig:workload-regret-gap} make the
comparison with the two state-aware methods easier to see.
\SHAPE's advantage over myopic pricing grows
over the horizon. The policies also induce different state visits:
\SHAPE\ spends 16.22\% of periods in the highest workload band, compared
with 27.44\% for myopic pricing. These visitation frequencies describe
each method's own path, including its learning periods, and show how
pricing changes the operating conditions faced by subsequent customers.

The dynamic plug-in method initially has lower mean regret than
\SHAPE, but the ordering reverses later in the horizon.
At $T=48{,}000$, \SHAPE\ reduces regret from 5,259.21 to
4,480.74, or by 14.80\%. Its paired advantage is 778.47, with interval
$[566.03,990.91]$. Both policies account for future states, but plug-in
first estimates the noise tail, whereas \SHAPE\ learns the common
price map directly.

All three state-aware methods also have lower mean regret than the
state-blind ORBIT adaptation, whose mean regret is 13,450.05.
Omitting the state combines customers facing different mean
valuations and dispersion even when their covariates coincide.
The absolute-regret plot compares all four implementations, while
the paired plots distinguish \SHAPE\ from methods that already use
the state.

\begin{table}[H]
 \centering
 \caption{Terminal cumulative regret and paired comparisons at
 $T=48{,}000$.}
 \label{tab:workload-regret}
 \small
 \setlength{\tabcolsep}{6pt}
 \begin{tabular}{lr}
 \toprule
 Method & Mean regret (Monte Carlo SE)\\
 \midrule
\SHAPE & $4{,}480.74\ (52.75)$\\
State-aware myopic & $7{,}627.40\ (150.40)$\\
Dynamic plug-in & $5{,}259.21\ (96.64)$\\
State-blind ORBIT & $13{,}450.05\ (182.86)$\\
 \midrule
 \multicolumn{2}{l}{Comparator minus \SHAPE: mean [paired 95\% interval]}\\
State-aware myopic & $3{,}146.67\ [2{,}848.44,3{,}444.89]$\\
Dynamic plug-in & $778.47\ [566.03,990.91]$\\
State-blind ORBIT & $8{,}969.32\ [8{,}603.91,9{,}334.72]$\\
 \bottomrule
 \end{tabular}
 \par\smallskip
 \begin{minipage}{\linewidth}
 \footnotesize
 Parentheses give Monte Carlo standard errors across 200 replications.
 Positive paired differences favor \SHAPE. All startup, exploration,
 extension and fallback costs are included; all 200 replications in
 this regime are retained.
 \end{minipage}
\end{table}

\FloatBarrier

\section{Conclusion}
\label{sec:conclusion}

This paper develops a novel framework for contextual dynamic pricing in
systems where purchases change the conditions faced by subsequent customers. The
purchase shadow measures that sale's contribution to future revenue;
combining it with mean valuation and the state scale gives a common
scalar pricing problem across states. \SHAPE\ uses this structure to
learn one price map while updating continuation values and purchase
shadows from the same sales records. The pricing history is retained
as those values change, and the policy learns the quantities needed
for pricing without fitting the full noise distribution or transition
matrices. This connects customer-level price responses with the value
of the evolving sales environment.

Under the stated regularity, coverage and design conditions, the regret
analysis separates price-map learning from valuation calibration and
controls their interaction with value updates. For affine mean
valuations, fixed state and feature dimensions, and every fixed
$\beta\ge2$, matching bounds identify the minimax time exponent
$(2\beta-1)/(4\beta-3)$. The same policy architecture gives explicit
upper bounds for nonlinear and sparse mean valuations. The reported
workload experiment complements these guarantees: \SHAPE\ has lower
mean regret than state-aware myopic and dynamic plug-in pricing,
and its prices lead to fewer visits to the highest workload band than
myopic pricing. Together, the structural reduction, learning guarantee
and numerical comparison show how pricing can account for both the
customer facing an offer today and the system inherited by later customers.

\FloatBarrier

\clearpage
\begin{appendices}
\section*{Appendices}
The appendices give the algorithm settings and proofs.
Appendix~\ref{app:operational-details} specifies the domains, sample
budgets and initialization rules. Appendix~\ref{app:regret-proofs}
proves the regret guarantee through the joint value and shadow updates.
Appendix~\ref{app:predictable-bandit} establishes the conditional
convex-bandit guarantee used in Appendix~\ref{app:pricing-certificate-proof}
to control the persistent pricing learner.
Appendix~\ref{app:valuation-models} verifies the valuation estimators
and the fixed-dimensional lower bound.

\section{Operational settings for the method}
\label{app:operational-details}

This appendix gives the deterministic settings used by the algorithm
in Section \ref{sec:method}. Precision checks use error radii computed
from minimum sample counts and supplied bounds.
For $0<\delta<1$, put $\nu_{\min}=\min_{s\in[K]}\nu(s)$ and write
$H_V=p_U/a_0$, $q_S=a_0\nu_{\min}$,
$R=p_U+H_V$, and $\ell_T=\log\{64K(T+1)^2/\delta\}$.
Every grid and tie-breaking rule below is fixed before the data it uses.

\subsection{Domains and feasibility margins}
\label{app:domain-details}

We use the Taylor convention for H\"older balls. For an order
$\alpha>0$ and radius $L$, put $j=\lceil\alpha\rceil-1$.
A function $f$ on a compact interval belongs to this ball if it has
continuous derivatives through order $j$, each bounded in absolute
value by $L$, and
\[
 \left|f(v)-\sum_{k=0}^{j}\frac{f^{(k)}(u)}{k!}(v-u)^k\right|
 \le L|v-u|^\alpha
 \quad\text{for every }u,v\text{ in that interval}.
\]
At integer orders this is the $C^{\alpha-1,1}$ convention. In
particular, order one means Lipschitz continuity, so Assumptions
\ref{ass:tail-regularity} and \ref{ass:demand-geometry} give the
$L_g$- and $L_\psi$-Lipschitz bounds used below. The pointwise
second-derivative condition near the scalar maximizer is stated
separately in Assumption \ref{ass:demand-geometry}.

The policy uses three compact intervals: $\mathcal Z$ for its scalar
input bins, $\cQ$ for standardized actions and $\cW$ for the residual
grid used in Bellman initialization. The common tail is regular on
$\mathcal I$ from Assumption \ref{ass:tail-regularity}. For analysis,
the supplied domains satisfy
\[
 \cW\subset\operatorname{int}(\cW^{\rm b})
 \subset\operatorname{int}(\mathcal I),\qquad
 \mathcal Z\subset\operatorname{int}(\mathcal Z^+),\qquad
 \cQ\subset\operatorname{int}(\cQ^+).
\]
All intervals here are compact. The intermediate residual interval
$\cW^{\rm b}$ defines a reference operator; $\mathcal Z^+$ and
$\cQ^+$ accommodate index and scale perturbations.

The enlarged-domain requirement in Assumption \ref{ass:demand-geometry}
means that its quadratic bounds hold for every
$(z,q)\in\mathcal Z^+\times\cQ^+$, with the same $\lambda_r,L_r$.
For each $z\in\mathcal Z^+$, $\psi(z)$ is the unique maximizer over
$\cQ^+$ and has a fixed positive distance from $\partial\cQ$.
Its H\"older and Lipschitz bounds hold on $\mathcal Z^+$, and the
local-concavity bound holds on this domain whenever
$|q-\psi(z)|\le\rho_0$.

Recall the reference price sets from Section~\ref{sec:stationary-target},
\[
 \cP^{\rm b}(x,s)=
 \{p\in\cP:(p-m_*(x,s))/\sigma_s\in\cW^{\rm b}\},
\]
and define
\begin{equation}
 (\mathcal T h)(s)=(P_0h)(s)+
 \E_X\max_{p\in\cP^{\rm b}(X,s)}
 (p+b_s(h))D_s(p\mid X).
 \label{eq:bellman-operator}
\end{equation}
Assumption \ref{ass:utility-domain} requires these price sets to be
nonempty and the displayed stationary selector to lie inside them
with a fixed margin. Coverage then gives the anchored solution
\begin{equation}
 \rho^*\one+h^*=\mathcal T h^*,\qquad h^*(s_{\rm ref})=0,
 \qquad \spn(h^*)\le H_V.
 \label{eq:aroe}
\end{equation}
$|b_s^*|\le\spn(h^*)\le H_V$, so a universal shadow bound follows
from coverage. The policy may use a tighter supplied interval $\cB$.

Here are the precise domains in Assumption
\ref{ass:implementation-domains}. With uniform mean and scale errors
at most $\eta_0$, every price
$\widehat m(x,s)+\widehat\sigma_s w$, $w\in\cW$, is in $\cP$ and
has true residual in $\mathcal I$; this price set contains
$p^*(x,s)$ with a fixed interior margin. If the shadow error is
also at most $\eta_0$, define
\[
 a_s=\widehat\sigma_s/\sigma_s,\qquad
 c(x,s)=\{m_*(x,s)+\widehat b_s\}/\sigma_s.
\]
The fitted input lies in $\mathcal Z$, every $q\in\cQ$ gives a valid
physical price with true residual in $\mathcal I$, and
\[
 q^\dagger(x,s)=a_s^{-1}\psi\{c(x,s)\}
 \quad\text{has distance at least }\rho_Q\text{ from }\partial\cQ.
\]
The shadow always belongs to $\cB$.

The containments $c(x,s)\in\mathcal Z^+$ and
$a_s\cQ\subset\cQ^+$ follow from the fixed interval margins.
Let $d_Z>0$ be the distance from $\mathcal Z$ to
$\partial\mathcal Z^+$, and $d_Q>0$ the distance from $\cQ$ to
$\partial\cQ^+$. Put $Q_{\max}=1\vee\sup_{q\in\cQ}|q|$.
Choose the supplied local radius
small enough that
\begin{equation}
 \eta_0\le\frac{\underline\sigma}{2}
             \min\{d_Z,d_Q/Q_{\max}\}.
 \label{eq:domain-precision}
\end{equation}
Indeed, $|c-z_s^*(x)|\le\eta_0/\underline\sigma$ and
$\sup_{q\in\cQ}|a_sq-q|\le Q_{\max}\eta_0/\underline\sigma$.
Thus $c\in\mathcal Z^+$ and $a_s\cQ\subset\cQ^+$ at every locally
admissible pricing call.

\subsection{Calibration and local precision}

Require $q_S(n_u-1)\ge32\ell_T$ and $N_U\ge1$ before computing any
calibration radius; otherwise invoke fallback. Use the
mean-estimator failure level $\varepsilon_{\rm fit}=\delta/(8KT)$
for each state and each possible prefix. Require its deterministic
sample-size gate at every integer $N_U\le n\le n_u$, and let
\[
 \eta_m=\sup_{N_U\le n\le n_u}
          r_{\mathcal M}(n,\varepsilon_{\rm fit}),\qquad
 \eta_\sigma=
 \frac{2B\eta_m+3B^2\sqrt{2\ell_T/N_U}}{2\underline\sigma}.
\]
The second radius uses the state-scale estimator in
\eqref{eq:scale-estimator} on the same complete statewise sample as the mean fit.
The supremum gives a radius valid at every possible retained count.
For the concrete routines in Sections~\ref{sec:linear-valuations} and
\ref{sec:valuation-models}, a gate that passes at $N_U$ also passes
at every larger count, and the radii are nonincreasing. Checking
$N_U$ therefore suffices.
A failed regression gate, an empirical
design failure, or fewer than $N_U$ calibration visits in any state
invokes fallback.

The perturbation coefficient can be computed from the declared
envelopes. Let $a_-=\underline\sigma/\overline\sigma$,
$a_+=\overline\sigma/\underline\sigma$,
$Z_{\max}=\sup_{z\in\mathcal Z}|z|$ and
$\Psi_{\max}=\sup_{q\in\cQ}|q|$. Put
\begin{align*}
 C_{\rm LS}&=1\vee
 \frac{\max\{L_\psi,L_\psi Z_{\max}+\Psi_{\max}\}}
      {\underline\sigma},&
 L_D&=L_g/\underline\sigma,\qquad L_V=L_D(1+L_\psi),\\
 C_p&=\max\left\{1,\,
 4\overline\sigma L_ra_+C_{\rm LS}^2,\,
 \frac{L_V}{2}+\frac{\overline\sigma L_D^2}{2\lambda_r}\right\}.
\end{align*}
With $\eta_b$ denoting a uniform bound on the supplied shadow error,
the squared-input contribution to the pricing certificate is
$C_pN(\eta_m^2+\eta_\sigma^2+\eta_b^2)$; Appendix
\ref{app:regret-proofs} verifies the deterministic perturbation
steps separately from bandit optimization.
The persistent module uses the global prefix envelope
$\Phi_T(n)$ in \eqref{eq:pricing-map-envelope}, with the known
structural $C_{\rm map}$ in \eqref{eq:pricing-map-constant}.
Its leading time exponent is
$\alpha=(2\beta-1)/(4\beta-3)$.
These are deterministic settings, not fitted model parameters.

For $\vartheta=1-a_0$, $c=8C_p/q_S$ and the local radius $r$ below,
the active-block schedule is
\begin{equation}
 \lambda=\frac{\vartheta+\sqrt{\vartheta^2+4cr}}2,\qquad
 \varrho=\frac{1+\lambda^{-2}}2,\qquad
 L_k=\lceil L_0\varrho^k\rceil.
 \label{eq:block-schedule}
\end{equation}
The conditions on $r$ ensure $0<\lambda<1$ and
$1<\varrho<\lambda^{-2}$. This choice accounts for the lagged
shadow update, rather than only the Bellman contraction
$\vartheta$. With $C_\varrho=2\varrho/(\varrho-1)$, the number of
active calls through a block is at most $C_\varrho$ times its
planned length. Select
\begin{equation}
 0<r\le\min\{H_V,\eta_0/4,a_0/(4c)\},\qquad
 \epsilon=\frac{a_0r}{32},\qquad
 r_{b,{\rm init}}=
 \min\left\{\frac r8,
 \sqrt{\frac{q_S(1-\lambda)r}{128C_p}}\right\}.
 \label{eq:precision-settings}
\end{equation}
The number $\lambda$ is as in \eqref{eq:block-schedule}.
Require $\eta_m,\eta_\sigma\le\eta_0$ and
\begin{equation}
 \eta_m\le\frac{\epsilon}{16C_{\rm app}},\qquad
 \frac{4C_pC_\varrho}{q_S}(\eta_m^2+\eta_\sigma^2)
 \le\frac{(1-\lambda)r}{16}.
 \label{eq:pilot-gates}
\end{equation}
Here $C_\varrho=2\varrho/(\varrho-1)$ for the block schedule
\eqref{eq:block-schedule}. One conservative approximation constant is
\[
 C_{\rm app}=4\left\{1+\overline\sigma+
 \frac{(p_U+H_V)L_g(2+\overline\sigma)}{\underline\sigma}\right\}.
\]
It bounds the safe residual-price objective's Lipschitz and
fitted-mean approximation errors for $\spn(h)\le H_V$.
Thus \eqref{eq:pilot-gates} sets a precision requirement on the
regression budget.

\subsection{Finite initialization budgets}

Use fitted-mean bins and a residual grid with mesh sizes
$\Delta_u,\Delta_w\le\epsilon/(16C_{\rm app})$; include both endpoints
of $\cW$. Let their sizes be $M$ and $A$, respectively.
Set
\begin{equation}
 n_{\rm w}=
 \left\lceil\frac{128A^2R^2(M\log A+\ell_T)}{\epsilon^2}\right\rceil,
 \quad
 n_b=\left\lceil\frac{8H_V^2\ell_T}{r_{b,{\rm init}}^2}\right\rceil.
 \label{eq:warm-quotas}
\end{equation}
If the residual interval is a singleton, use $A=1$ and interpret
$M\log A=0$. Each initialization block uses every observation of
each state for its empirical Bellman maximum, with minimum count
$n_{\rm w}$. The final block uses all observations of each
state/outcome pair for the initial shadow, with minimum count $n_b$.

For a statewise quota $n$ and overlap factor $\kappa$, define
\[
 D(n,\kappa)=
 1+\left\lceil\frac{8n+32\ell_T}{q_S\kappa}\right\rceil.
\]
Put $D_{\rm w}=D(n_{\rm w},1)$ and
$D_*=\max\{D_{\rm w},D(n_b,\kappa_0)\}$.
The first $J-1$ initialization blocks have nominal length $D_{\rm w}$;
the final block has nominal length $D_*$. The quotas set minimum counts;
the estimators use the full realized counts. Every elapsed period
counts toward the horizon. A count shortfall is checked at block end
and invokes fallback, unless bounded count recovery is used. Under this
optional rule, a block with insufficient counts continues with the same
randomized prices, frozen continuation vector and accumulated records
until all its count gates pass, up to three times its nominal length
or the end of the horizon. An unresolved shortfall then invokes fallback
if periods remain. A block that passes at its nominal deadline is
unchanged; the deterministic startup check still uses nominal lengths.
No division by an empty bin is needed in
\eqref{eq:warm-ips}; an empty fitted-mean bin has zero scores and uses
the fixed tie-breaking arm.

Write $\vartheta=1-a_0$. The number of initialization iterations is
\begin{equation}
 J=
 \begin{cases}
 1+\left\lceil\log(8H_V/r)/(-\log\vartheta)\right\rceil,
          &0<\vartheta<1,\\
 2,       &\vartheta=0.
 \end{cases}
 \label{eq:warm-iterations}
\end{equation}
Every block maximizes empirical IPS values using the fixed pre-block
vector $v_j$. The last block uses $v_{J-1}$ in both its value maximum
and initial-shadow contrast.
The nominal initialization duration is
\[
 \tau_{\rm w}=(J-1)D_{\rm w}+D_*.
\]

\subsection{Pricing-module and active-block settings}

Let $D_{\mathcal Z}$ be the length of the nondegenerate input envelope.
If necessary, enlarge a singleton declared input envelope to a
nondegenerate compact interval inside $\mathcal Z^+$ that still
satisfies the declared geometry and domain bounds.
Write $Q_{\max}=1\vee\Psi_{\max}$ and $\bar L_\psi=1\vee L_\psi$.
Choose $\rho_{\rm loc}>0$, $c_{\rm grid}>0$ and a grid so that
\[
 \frac{3\rho_{\rm loc}}8
 \le\min\{\rho_0/a_+,\rho_Q\},\qquad
 \operatorname{mesh}(\mathcal G_{\cQ})
 \le c_{\rm grid}\rho_{\rm loc}^2,\qquad
 c_{\rm grid}\rho_{\rm loc}^2
 \le\frac1{16}\sqrt{\lambda_r/L_r}\,\rho_{\rm loc},
\]
including both endpoints of $\cQ$. Define
\begin{align*}
 C_{\rm drift}&=1\vee Q_{\max}L_g
 \max\left\{a_+,\frac2{\underline\sigma},
               \frac{2(Q_{\max}+Z_{\max})}{\underline\sigma}\right\},\\
 A_{\rm loc}&=\lambda_ra_-\rho_{\rm loc}^2/1024,\\
 h_{\rm gate}&=\min\left\{D_{\mathcal Z},
 \frac{A_{\rm loc}}{2C_{\rm drift}},
 \frac{\rho_{\rm loc}}{32\bar L_\psi},
 \left(\frac{\rho_{\rm loc}}{32\bar L_\psi}\right)^{1/(\beta-1)}
 \right\}.
\end{align*}
Take $0<\ell_{\rm loc}\le h_{\rm gate}$ and
\[
 m_0\ge(16Q_{\max}/A_{\rm loc})^2,\qquad
 \eta_m+\eta_\sigma\le
 \min\left\{\frac{A_{\rm loc}}{4C_{\rm drift}},
                  \frac{\rho_{\rm loc}}{16C_{\rm LS}}\right\}.
\]
These are additional deterministic precision and tuning checks.
Fix $c_h>0$, initialize one module with global budget $T$, and choose
\[
 h_T=\min\{D_{\mathcal Z},\ell_{\rm loc},
                    c_hT^{-1/(4\beta-3)}\},\qquad
 M_T=\lceil D_{\mathcal Z}/h_T\rceil.
\]
Partition the entire envelope into $M_T$ equal-width bins, with
unique half-open assignments and the rightmost endpoint included.
For every such width, the preceding settings give
\begin{align*}
 C_{\rm drift}(h_T+\eta_m+\eta_\sigma)
      +4Q_{\max}/\sqrt{m_0}&\le A_{\rm loc},\\
 L_\psi(h_T^{\beta-1}+h_T)
      +C_{\rm LS}(\eta_m+\eta_\sigma)&\le\rho_{\rm loc}/8.
\end{align*}
One may take
\[
 H_{\rm act}=\left\lceil\max\left\{2,
 \left(\frac{c_h}{\min\{D_{\mathcal Z},\ell_{\rm loc}\}}\right)^{4\beta-3}
 \right\}\right\rceil.
\]
Require $T\ge H_{\rm act}$; then $h_T=c_hT^{-1/(4\beta-3)}$.
Each bin uses
\[
 m_{\rm coarse}=
 \left\lceil m_0\log\{eT M_T|\mathcal G_{\cQ}|/\delta_A\}\right\rceil,
 \qquad \delta_A=\delta/16.
\]
All bins, anchors and local-copy histories are retained when the
outer block changes. Only the predictable shadow used in the scalar
input and physical-price conversion is updated.
The refinement generator uses affine normalization and local doubling
as in the theoretical ORBIT construction. Its declared dimension is
$d_{\rm act}+1$, and its observed normalized loss is
$1/2-q_tY_t/(2Q_{\max})$. Use the copy confidence allocation in
\eqref{eq:pricing-copy-allocation}, the explicit body positioning in
\eqref{eq:pricing-legendre-positioning}, and the tuning and short-copy
rule in \eqref{eq:bco-copy-tuning}. A short copy plays the anchor,
the centered body's origin. No copy queries the unknown map.
Use the symmetric polynomial trust region around the stored grid
maximizer as the coefficient body. Project each generated refinement
action onto $\cQ$, and form the
normalized loss from the executed action. On successful localization,
the entire trust region already takes values in $\cQ$, so this
projection is inactive. The fixed Legendre transformation in Appendix
\ref{app:pricing-certificate-proof} positions its coefficient body
between Euclidean balls of radii one and $d_{\rm act}+1$; it uses
no covariate-design assumption.
Membership is checked at the endpoints and, for a nonconstant
polynomial, at its real critical points in $[-1,1]$.

For complete-block calculations define the deterministic radii
\[
 r_b(L)=8H_V\sqrt{\frac{\ell_T}{q_S\kappa_0(L-1)}},
 \qquad
 r_v(L)=16R\sqrt{\frac{\ell_T}{q_S(L-1)}}\qquad(L\ge2).
\]
Choose $L_0\ge\max\{2,M_T\}$ so that
\begin{align}
 r_b(L_0)&\le r_{b,{\rm init}},\notag\\
 \frac{4\Phi_T(C_\varrho L_0)}{q_SL_0}
       &\le\frac{(1-\lambda)r}{16},
 &r_v(L_0)&\le\frac{(1-\lambda)r}{16}.
 \label{eq:block-gates}
\end{align}
Together with \eqref{eq:pilot-gates} and
\eqref{eq:precision-settings}, these inequalities allocate the
forcing terms in the delayed value recursion. The map gate includes
the nonvanishing term $C_{\rm map}\ell_T^{17}C_\varrho h_T^{2\beta-2}$;
increasing $L_0$ alone cannot make that term small. For complete
active blocks the other terms decrease with block length. The global
budget gate is
\[
 n_u+\tau_{\rm w}+L_0\le T.
\]
If any deterministic gate fails, use fallback for the entire horizon,
as in Algorithm~\ref{alg:shape} and Theorem~\ref{thm:conditional-master}.

For a complete active block of length $L$, require, for every state
and outcome,
\[
 N(s)\ge\max\{1,\lfloor q_S(L-1)/4\rfloor\},\qquad
 N_y(s)\ge\max\{1,\lfloor q_S\kappa_0(L-1)/4\rfloor\}.
\]
On failure, do not form \eqref{eq:joint-update}. Before each active
quote, check that $\widehat z_t$ is finite, project it onto
$\mathcal Z$, and check the module's numerical flag and
$p_t\in\cP$; a failed check rejects the candidate and invokes
the fallback rule in Section~\ref{sec:unified-policy}.
Projection changes neither the supplied shadow nor the
conversion from a standardized action to a physical price.
On the local-precision event, the maintained domains and model
assumptions give the purchase overlap used in the analysis.

A partial terminal block has no count check or continuation update.
The persistent module retains its global budget $T$ and all local
histories; stopping does not retune its bins or restart its generator.

For fixed structural constants and $\delta=T^{-3}$, the map floor
tends to zero. Writing $\chi=(1-\lambda)r/16$ and
$A_{\rm map}=4C_{\rm map}/q_S$, sufficient map settings are
\[
 A_{\rm map}C_\varrho\ell_T^{17}h_T^{2\beta-2}\le\chi/2,
 \qquad
 L_0\ge\max\{M_T,16A_{\rm map}^2C_\varrho\chi^{-2}
                           M_T\ell_T^{34}\}.
\]
Together with the sampling gates they permit
$L_0=O(M_T\ell_T^{34})$. This startup cost is retained explicitly.
For other confidence schedules, the displayed finite-sample gates
determine feasibility.

\section{Proofs for the regret analysis}
\label{app:regret-proofs}

Write $H=H_V$, $q=q_S$, $\vartheta=1-a_0$ and $\ell=\ell_T$ in this
appendix. At a period $t$, let $\mathscr G_t$ contain the entire past,
the current $S_t$ and all fixed block inputs, but not the current
covariate or fresh pricing randomization. After the quote and purchase,
let $\mathscr L_t$ additionally contain $X_t,p_t,Y_t$, before
$S_{t+1}$ is observed. These information times distinguish selection
of a state visit from selection of a state--outcome visit.

\subsection{Bellman structure and the benchmark}

Consider a price correspondence whose feasible prices have residuals
in $\mathcal I$. Assumption~\ref{ass:coverage} gives the common
minorization $a_0\nu$ for every induced transition row. The difference
between its Bellman operators at $h$ and $\widetilde h$ lies, at every state, in
\[
 a_0\nu^\top(h-\widetilde h)
 +(1-a_0)[\min(h-\widetilde h),\max(h-\widetilde h)].
\]
Indeed, every feasible transition row is
$a_0\nu+(1-a_0)R_{s,x,p}$ for a probability row $R_{s,x,p}$.
Apply the maximum-difference inequality at each $x$ and integrate.
Taking span gives contraction by $\vartheta$. On anchored vectors,
span is a norm, so Banach's theorem gives the unique anchored $h^*$
and its additive eigenvalue $\rho^*$. The span of $\mathcal T0$ is
at most $p_U$, which gives $a_0\spn(h^*)\le p_U$.
The true-shadow oracle in Assumption \ref{ass:utility-domain} makes
the restricted and full Bellman objectives agree at this fixed point.

\begin{proof}[Proof of Proposition \ref{prop:benchmark-bridge}]
By \eqref{eq:true-bellman-gap}, the conditional one-period identity is
\[
 \E[p_tY_t+h^*(S_{t+1})\mid\mathscr G_t]
 =\rho^*+h^*(S_t)
   -\E[\Delta^*(X_t,S_t,p_t)\mid\mathscr G_t].
\]
Telescoping gives
\[
 \E_s^\pi\sum_{t=1}^Tp_tY_t
 =T\rho^*+h^*(s)-\E_s^\pi h^*(S_{T+1})
  -\E_s^\pi\sum_{t=1}^T\Delta^*(X_t,S_t,p_t).
\]
Nonnegativity of the gaps bounds any policy's revenue above by
$T\rho^*+h^*(s)-\min h^*$. The stationary selector
\eqref{eq:physical-oracle} has zero gap and revenue at least
$T\rho^*+h^*(s)-\max h^*$. These two bounds also sandwich the
finite-horizon optimum. Subtract the learner's identity and use
$\max h^*-\min h^*\le H$ to obtain
\eqref{eq:regret-gap-bridge}.
If $P_0=P_1$, every policy has the same terminal-state distribution.
The zero-gap stationary selector attains the resulting revenue bound,
so subtracting its identity gives the exact comparison in this case.
\end{proof}

The span projection used throughout the method is nonexpansive.
To check this, let $m=\min(v-w)$ and $M=\max(v-w)$. Monotonicity
and translation equivariance of $C_H$ give
\[
 C_Hw+m\one\le C_Hv\le C_Hw+M\one.
\]
Thus $\spn(C_Hv-C_Hw)\le\spn(v-w)$.
If $\spn(w)\le H$, then $C_Hw=w$. Anchoring does not change span,
so projection cannot increase the error relative to $h^*$ when
the comparison vector is $\rho^*\one+h^*$.

\subsection{Coverage, calibration and finite initialization}

We first record the probability arguments used below. Suppose a
block's quoted prices preserve purchase overlap and its inputs are
fixed or predictable. For a block of length $L$, the $L-1$ successor
states after its first period have conditional visit probability
at least $q$ for each state. Multiplicative Bernoulli martingale
concentration gives
\begin{equation}
 \Pp\{N(s)<q(L-1)/2\mid\text{block start}\}
 \le e^{-q(L-1)/8}.
 \label{eq:state-count-tail}
\end{equation}
For completeness, if $I_i$ has conditional mean at least $p$,
then for $\zeta>0$,
$\E[e^{-\zeta I_i}\mid\text{past}]
 \le1-p(1-e^{-\zeta})$.
Iterating and applying Markov's inequality at $\zeta=\log2$
yields the usual lower-tail bound
$\Pp(\sum_{i=1}^nI_i<pn/2)\le e^{-pn/8}$.

To count outcomes, take $m=\lceil q(L-1)/2\rceil$ and consider
the first $m$ visits to state $s$. Before the purchase on each such
visit, both outcomes have conditional probability at least
$\kappa_0$ after averaging the current covariate and seed.
A second application of the same inequality gives
\begin{equation}
 \Pp\{N_y(s)<q\kappa_0(L-1)/4,\
            N(s)\ge m\mid\text{block start}\}
 \le e^{-q\kappa_0(L-1)/16}.
 \label{eq:outcome-count-tail}
\end{equation}
The visit sequence can be defined by continuing the frozen phase
virtually beyond its scheduled end. On the count event, its first
$m$ visits occur within the real block. Neither
\eqref{eq:state-count-tail} nor
\eqref{eq:outcome-count-tail} conditions on the final visit count.
In particular, no one-step lower bound is asserted for the pair
indicator when the current state has already been revealed.

For a bounded martingale difference with conditional range at most
$B_0$, the exponential bound
$\E[e^{uM}\mid\text{past}]\le e^{u^2B_0^2/8}$
implies a two-sided prefix-mean bound
$B_0\sqrt{v/(2n)}$ with failure at most $2e^{-v}$.
Predictable state selection, or selection after the purchase but
before the next state, preserves this argument by stopping the
exponential supermartingale. A union bound over $n\le T$ makes the
bound valid at every observed visit-prefix length.

\paragraph{Calibration event.}
Uniform prices have purchase probability
$\{\E[V_t\mid X_t,S_t]-p_L\}/B$.
For $V_t\in[p_L,p_U]$,
\[
 \operatorname{Var}(V_t\mid X_t,S_t)
 \le B\{\E[V_t\mid X_t,S_t]-p_L\},
 \quad
 \operatorname{Var}(V_t\mid X_t,S_t)
 \le B\{p_U-\E[V_t\mid X_t,S_t]\}.
\]
The scale lower bound therefore gives overlap
$\underline\sigma^2/B^2\ge\kappa_0$. With the deterministic
calibration count gate in Appendix \ref{app:operational-details},
\eqref{eq:state-count-tail} supplies $N_U$ visits to each state
except with probability $Ke^{-4\ell}$.
Continue the uniform-price phase virtually beyond its scheduled end.
A state visit is selected before its fresh covariate, quote and
valuation noise, whose conditional distribution is the same at every
visit to that state. Induction therefore gives the regression
routine's product experiment for every deterministic statewise
prefix. This experiment is not conditioned on its finite-block count.
Write $N_s^U$ for the actual number of calibration visits to state $s$.

Apply the regression contract at failure level
$\varepsilon_{\rm fit}=\delta/(8KT)$ to every prefix length
$N_U\le n\le n_u$. The deterministic gates in Appendix
\ref{app:operational-details} cover all these lengths. A union bound
over states and prefixes costs at most
$Kn_u\varepsilon_{\rm fit}\le\delta/8$. On this simultaneous event
and the count event, the fit at $n=N_s^U$ succeeds and has error at
most the deterministic envelope $\eta_m$. No independence
conditional on $N_s^U$ is required.

To analyze scale estimation on the same records, use the oracle response
\[
 W_t=Z_t^{(2)}-\{m_*(X_t,s)-p_L\}^2,
 \qquad -B^2\le W_t\le2B^2.
\]
Equation \eqref{eq:calibration-moments} gives conditional mean
$\sigma_s^2$ before the uniform quote and purchase. Apply the
prefix-mean bound with range $3B^2$ and exponent $4\ell$, then take
a union over states and $n\le n_u$. Except with probability
$2Kn_ue^{-4\ell}$, the sample mean of each prefix differs from
$\sigma_s^2$ by at most $3B^2\sqrt{2\ell/n}$. Evaluate this bound
at the actual $N_s^U\ge N_U$.
On the uniform mean-fit event, projection of both means into
$[p_L,p_U]$ gives the pathwise bound
\[
 \left|\{\widehat m(X_t,s)-p_L\}^2
        -\{m_*(X_t,s)-p_L\}^2\right|\le2B\eta_m.
\]
Variance projection and
$\widehat\sigma_s+\sigma_s\ge2\underline\sigma$ therefore give
\[
 \max_s|\widehat\sigma_s-\sigma_s|
 \le\frac{2B\eta_m+3B^2\sqrt{2\ell/N_U}}
           {2\underline\sigma}=\eta_\sigma.
\]
This argument intersects two concentration events after proving them;
it neither conditions on the fitted mean nor assumes independence
between the mean and scale estimates. Their squared scale error is
$O(\eta_m^2+\ell/N_U)$. Counts and scale means together fail with
probability at most
$K(1+2n_u)e^{-4\ell}\le3Ke^{-\ell}$, since $n_u\le T$.
Together with the simultaneous regression event, these events cover
all calibration failures.

Let $\cE_{\rm pil}$ be the event that calibration passes its empirical
checks and the frozen fits have errors at most $\eta_m$ and
$\eta_\sigma$ from Appendix \ref{app:operational-details}.
Conditional on the calibration history on this event, all residual
prices in the continuous family
$\widehat A_s(x)=\{\widehat m(x,s)+\widehat\sigma_s w:w\in\mathcal W\}$
are feasible. This family's fixed population Bellman operator,
denoted by $\mathcal T_{\widehat A}$,
contracts span by $\vartheta=1-a_0$ and retains the true fixed point:
\[
 \mathcal T_{\widehat A}h^*=\rho^*\one+h^*.
\]
The equality uses feasibility of $p^*$ at $h^*$, not optimality of
the scalar price map at every intermediate continuation vector.

\begin{proposition}[Initialization from binary feedback]
\label{prop:warm-start}
Use the grids, quotas, precision gates and iteration count in Appendix
\ref{app:operational-details}. If the deterministic budget gate passes,
then, conditional on the calibration history on $\cE_{\rm pil}$,
with probability at least $1-\delta/8$ all initialization quotas pass
and, for $0\le j\le J$,
\begin{equation}
 \spn(v_j-h^*)\le
 \vartheta^jH_V+\frac{2\epsilon(1-\vartheta^j)}{a_0}.
 \label{eq:warm-contraction}
\end{equation}
In particular, with $e_{-1}=\spn(h_{-1}-h^*)$ and
$e_0=\spn(h_0-h^*)$,
\[
 e_{-1},e_0\le r/4,\qquad
 \|b_{-1}-b^*\|_\infty\le e_{-1}+r_{b,{\rm init}}.
\]
\end{proposition}

The finite residual grid and fitted-mean bins are used only to reach
a fixed structural precision. Each residual-price block estimates
all candidate policy values and uses their empirical maximum to
update the continuation vector. The proof controls these values
using the total number of visits to each state, including bins of
arbitrarily small probability, without a separate evaluation block
or noise-tail estimate. At fixed structural constants, the
initialization budgets beyond calibration grow logarithmically
with the confidence parameter and horizon.

\begin{proof}[Proof of Proposition \ref{prop:warm-start}]
Condition on the calibration history on $\cE_{\rm pil}$.
The safe residual-price correspondence is now fixed. Every one of
its transition rows minorizes $a_0\nu$, so its true population
operator $\mathcal T_{\widehat A}$ contracts span by $\vartheta$.
Assumption \ref{ass:implementation-domains}(i) retains $p^*$,
hence $\mathcal T_{\widehat A}h^*=\rho^*\one+h^*$.

Fix an initialization iteration and a vector $h$ of span at most $H$.
For a state $s$, the true one-step return mean at residual $w$ is
\[
 F_h(x,w)=(P_0h)(s)+
 \{\widehat m(x,s)+\widehat\sigma_s w+b_s(h)\}
 g\!\left(\frac{\widehat\sigma_s w+\widehat m(x,s)-m_*(x,s)}
                   {\sigma_s}\right).
\]
For two contexts in one fitted-mean bin, uniformly in $w$,
\[
 |F_h(x,w)-F_h(x',w)|
 \le\Delta_u+2RL_g\eta_m/\underline\sigma.
\]
Rounding $w$ to the residual grid changes $F_h$ by at most
$\{\overline\sigma+RL_g\overline\sigma/\underline\sigma\}\Delta_w$.
Both endpoints in each comparison are actual safe residuals in
$\mathcal I$; its Lipschitz condition therefore applies on the intervening
interval. The difference between the expected maximum grid return
within a bin and the largest expected fixed-arm return is at most
twice the within-bin oscillation. Sum over bins, with zero contribution
from null bins. The best bin policy is within
\[
 C_{\rm app}(\Delta_u+\Delta_w+\eta_m)\le3\epsilon/16
\]
of $\mathcal T_{\widehat A}h$ at every state.

Now condition on the start of iteration $j$, so $v_j$ is fixed,
and write $L_j$ for its calendar length.
For a fixed bin policy $\pi$, its IPS response is
$A\one\{A_t=\pi(I(\widehat m(X_t,s)))\}\mathcal R_t^j$.
It lies in $[0,AR]$ and its conditional mean equals the policy's
expected shifted return. Continue the same frozen iteration
virtually, apply the visit-prefix mean bound, and take a union over
the $A^M$ policies and every prefix length $n\le L_j\le T$.
This costs at most $2Te^{-\ell}$ per state and iteration. On the
resulting simultaneous event, let $N_{j,s}$ be the actual number of
state-$s$ records in the block. When $N_{j,s}\ge n_{\rm w}$, the
uniform score error for the full-record scores is at most
\[
 AR\sqrt{\frac{M\log A+\ell}{2N_{j,s}}}
 \le AR\sqrt{\frac{M\log A+\ell}{2n_{\rm w}}}
 \le\epsilon/16.
\]
The empirical Bellman value $\bar Z_j(s)$ is the maximum of these
empirical policy values after restoring $\min v_j$. The difference
between two maxima is bounded by the largest pointwise difference,
so its error relative to the best population bin-policy value is
at most $\epsilon/16$, without a policy-selection loss or a new
evaluation sample. The grid and bin approximation then give
\[
 \|\bar Z_j-\mathcal T_{\widehat A}v_j\|_\infty
 \le3\epsilon/16+\epsilon/16=\epsilon/4.
\]
To verify the minimum counts used in these bounds, use
\eqref{eq:state-count-tail}--\eqref{eq:outcome-count-tail} with
$L=D(n,\kappa)$.
Here $q(L-1)\ge(8n+32\ell)/\kappa$. Thus every randomized block
of length $D_{\rm w}=D(n_{\rm w},1)$ supplies at least $n_{\rm w}$
visits to every state,
apart from the stated exponential failures. The final duration
$D_*=\max\{D_{\rm w},D(n_b,\kappa_0)\}$ also supplies $n_b$
observations of every state/outcome pair. These are intersections
with the simultaneous prefix events, not conditioning on future
counts. The estimates use all available records once these
minimum gates pass.

Projection nonexpansiveness now gives
$\spn(v_{j+1}-h^*)\le\vartheta\spn(v_j-h^*)+2\epsilon$.
Starting from $\spn(v_0-h^*)\le H$ proves
\eqref{eq:warm-contraction}. The chosen $J$ makes
$\vartheta^{J-1}H\le r/8$ when $\vartheta>0$, and
$2\epsilon/a_0=r/16$. Thus both final errors are at most $3r/16$.
For $\vartheta=0$ and $J=2$, the same conclusion follows after
one update.

In the final randomized block, every observation of a state/outcome
pair evaluates the already frozen $v_{J-1}$, not $v_J$.
Conditional on the purchase but before the next state, the mean
of $v_{J-1}(S_{t+1})$ is $(P_yv_{J-1})(s)$ and its range has
length at most $H$. Apply the outcome-selected martingale mean
bound to every prefix and take a union over states, outcomes and
$n\le T$. The failure cost is at most $4KTe^{-\ell}$. On this
event, write $N_{J-1,s,y}\ge n_b$ for the actual pair counts.
The unprojected contrast error is at most
\[
 H\sqrt{\frac{\ell}{2N_{J-1,s,1}}}
 +H\sqrt{\frac{\ell}{2N_{J-1,s,0}}}
 \le H\sqrt{\frac{2\ell}{n_b}}
 \le r_{b,{\rm init}}/2.
\]
Projection can be compared with $b^*\in\cB$, and
\[
 \|(P_1-P_0)(v_{J-1}-h^*)\|_\infty
 \le\spn(v_{J-1}-h^*).
\]
This gives the initial-shadow conclusion. Sharing the final block
requires no independence between the IPS values and outcome means:
their concentration bounds are derived under block-start information
and then intersected with each other and the count events.
All probability bounds are conditional on the information preceding
their respective randomized blocks. The passing horizon budget
implies $J\le T$. IPS errors over all iterations cost at most
$2KJTe^{-\ell}$, while the count errors are bounded by
$4K(J+1)e^{-\ell}$. Including the final-shadow means, their sum
is at most
\[
 \{2KJT+4KT+4K(J+1)\}e^{-\ell}
 \le8K(T+1)^2e^{-\ell}=\delta/8.
\]
\end{proof}

\subsection{One-block identities and delayed stability}

For a complete active block $k$, put
\[
 A_k=\sum_{t\in\mathcal I_k}\Delta^*(X_t,S_t,p_t),\qquad
 e_k=\spn(h_k-h^*),\qquad N_{k,\min}=\min_sN_k(s).
\]
The next lemma does not assume a pricing-regret bound. It quantifies
how the actual pricing loss enters the return-based update.

\begin{lemma}[One pricing block, two continuation estimates]
\label{lem:joint-block}
Condition on block-start information, including the fixed vector
$h_k$ with span at most $H_V$. Suppose the block has length $L\le T$,
all quotes are feasible, and the purchase probability after averaging
current covariates and policy randomization lies in
$[\kappa_0,1-\kappa_0]$. If $q_S\kappa_0(L-1)\ge16\ell_T$, there is an
event with conditional failure probability at most
\[
 8KL e^{-\ell_T}
 +K e^{-q_S(L-1)/8}
 +2K e^{-q_S\kappa_0(L-1)/16}
\]
on which all runtime count gates pass,
$N_{k,\min}\ge q_S(L-1)/2$, and the updates satisfy
\begin{align}
 e_{k+1}
 &\le\vartheta e_k+A_k/N_{k,\min}+r_v(L),\notag\\
 \|b_k-b^*\|_\infty&\le e_k+r_b(L).
 \label{eq:joint-block-bounds}
\end{align}
\end{lemma}

The value and shadow estimates may depend on one another. The proof
uses the fixed pre-block target $h_k$ and chronological martingales,
rather than held-out independence. Effective coverage applies to
the transition row after averaging the current policy's randomization;
it need not hold for either purchase-conditioned row separately.

\begin{proof}[Proof of Lemma \ref{lem:joint-block}]
Fix the block-start inputs and put $e=h_k-h^*$. Let
\[
 K_t=\Pp(S_{t+1}\in\cdot\mid\mathscr G_t),\qquad
 d_t=\E[\Delta^*(X_t,S_t,p_t)\mid\mathscr G_t].
\]
The Bellman equation and the independence of the fresh covariate give
the exact identity
\begin{equation}
 \E[Z_t^k\mid\mathscr G_t]
 =\rho^*+h^*(S_t)+K_te-d_t.
 \label{eq:shared-return-identity}
\end{equation}
This is an evaluation identity for the actual pricing decisions,
not an assertion that the empirical return mean equals
$\mathcal T h_k$.

The purchase probability averaged over the current covariate and
policy seed belongs to the effective-coverage interval. Thus
$K_t\ge a_0\nu$, and every $K_te$ lies in the same interval
$a_0\nu^\top e+\vartheta[\min e,\max e]$.
Its width is $\vartheta\spn(e)$.

For each state and every visit prefix, apply the martingale mean
bound to $Z_t^k-\E[Z_t^k\mid\mathscr G_t]$ and
$\Delta_t^*-d_t$. Each underlying variable has conditional range
at most $R$. For the outcome-selected continuation means, condition
instead on $\mathscr L_t$; the model gives
\[
 \E[h_k(S_{t+1})\mid\mathscr L_t]
 =(P_{Y_t}h_k)(S_t).
\]
Their conditional range is at most $H$. Union over both sides,
states, two outcomes and prefixes $n\le L$ costs at most
$8KL e^{-\ell}$. Let $\zeta_s$ be the return-mean noise and
$\eta_s$ the mean of $\Delta_t^*-d_t$ in state $s$.
On this event, average \eqref{eq:shared-return-identity} by state:
\[
 \bar Z_k(s)-\{\rho^*+h^*(s)\}
 =\overline{K_te}_s-\bar d_s+\zeta_s.
\]
Since $d_t\ge0$ and the realized gaps are nonnegative,
\[
 0\le\bar d_s\le A_k/N_{k,\min}
                  +R\sqrt{\ell/(2N_{k,\min})}.
\]
The common interval for $K_te$ and the span of $\zeta$ therefore give
\[
 \spn\{\bar Z_k-(\rho^*\one+h^*)\}
 \le\vartheta e_k+A_k/N_{k,\min}
                 +3R\sqrt{\ell/(2N_{k,\min})}.
\]
Use \eqref{eq:state-count-tail}--\eqref{eq:outcome-count-tail}.
On their success event,
$N_{k,\min}\ge q(L-1)/2$ and
$\min_{s,y}N_{y,k}(s)\ge q\kappa_0(L-1)/4$.
The first bound and span nonexpansiveness give the value inequality
with the stated, conservative $r_v(L)$.

For the shadow, interval projection is nonexpansive relative to $b_s^*$:
\[
 |b_k(s)-b_s^*|
 \le|\widehat\mu_{1,k}(s)-\widehat\mu_{0,k}(s)-b_s^*|
 \le e_k+H\sqrt{2\ell/\min_yN_{y,k}(s)}.
\]
The outcome count bound makes the last term at most $r_b(L)$.
The count and martingale failures sum to the lemma's stated bound.
All intersections are taken after deriving these bounds under
block-start information; no independence of the two estimates
or conditioning on a future pricing-success event was used.
\end{proof}

\begin{proposition}[The delayed continuation recursion]
\label{prop:delayed-stability}
Work on the calibration and initialization events of Proposition
\ref{prop:warm-start}, and suppose the single pricing event in
Proposition \ref{prop:pooled-pricing} and the block events in
Lemma \ref{lem:joint-block} hold up to each complete block considered.
Write $\eta^2=\eta_m^2+\eta_\sigma^2$, set
$r_{b,-1}=r_{b,{\rm init}}$ and $r_{b,k}=r_b(L_k)$.
Then, with $c=8C_p/q_S$,
\begin{equation}
 e_{k+1}\le\vartheta e_k+ce_{k-1}^2+u_k,\qquad
 u_k=\frac{4\Phi_T(C_\varrho L_k)}{q_SL_k}
       +\frac{4C_pC_\varrho\eta^2}{q_S}
       +\frac{8C_pr_{b,k-1}^2}{q_S}+r_v(L_k).
 \label{eq:delayed-recursion}
\end{equation}
For the schedule in Appendix \ref{app:operational-details},
$u_k\le(1-\lambda)r/4$ and
\[
 W_k:=e_k+(\lambda-\vartheta)e_{k-1}\le r/2
 \quad(k\ge0),\qquad
 W_{k+1}\le\lambda W_k+u_k.
\]
Consequently every inherited shadow stays inside the local domain.
\end{proposition}

\begin{proof}[Proof of Proposition \ref{prop:delayed-stability}]
For $L_k\ge2$, the strong state-count event implies
$N_{k,\min}\ge qL_k/4$. The previous shadow has error at most
$e_{k-1}+r_{b,k-1}$. Let $N_k^A$ be the number of active calls
through this complete block. The geometric schedule gives
\[
 N_k^A=\sum_{j=0}^kL_j
 \le2L_0\sum_{j=0}^k\varrho^j
 \le C_\varrho L_k.
\]
By nonnegativity of $\Gamma_t$, the persistent prefix event and
the pointwise shadow transfer imply
\begin{equation}
 A_k\le\Phi_T(C_\varrho L_k)
       +C_pC_\varrho L_k\eta^2
       +C_pL_k(e_{k-1}+r_{b,k-1})^2.
 \label{eq:persistent-block-loss}
\end{equation}
The prefix event is applied only at the actual $N_k^A\le T$;
monotonicity of $\Phi_T$ then permits evaluation at
$C_\varrho L_k$, even when this deterministic argument exceeds $T$.
Substitute this bound into \eqref{eq:joint-block-bounds} and use
$(e_{k-1}+r_{b,k-1})^2\le2e_{k-1}^2+2r_{b,k-1}^2$.
This gives \eqref{eq:delayed-recursion} with $c=8C_p/q$.

The pilot, map, previous-shadow sampling and current-value sampling
terms in $u_k$ are each at most $(1-\lambda)r/16$.
For $k=0$ this follows from $r_{b,-1}=r_{b,{\rm init}}$;
for later blocks $r_b(L_{k-1})\le r_b(L_0)$.
The length-dependent terms decrease with complete-block length;
the fixed map-approximation floor is included in the map gate.
Also $\vartheta+cr<1$, so $0<\lambda<1$.
Put $a=\lambda-\vartheta>0$; then $\lambda a=cr$.
Whenever $e_{k-1}\le r$,
\[
 e_{k+1}+ae_k
 \le(\vartheta+a)e_k+cr e_{k-1}+u_k
 =\lambda W_k+u_k.
\]
Initialization gives $W_0\le(1+a)r/4<r/2$.
Induction and $u_k\le(1-\lambda)r/4$ preserve $W_k\le r/2$.
The inherited shadow error is consequently at most
$r/2+r_{b,{\rm init}}\le5r/8<\eta_0$.
Thus the induction keeps the domain assumptions valid at every
active block start; it does not test this true error in the algorithm.
\end{proof}

\subsection{Deterministic pricing-input perturbations}

Fix a pricing call with valid frozen mean, scale and shadow estimates,
and condition on its current state $s$ and covariates $x$. Write
$a_s=\widehat\sigma_s/\sigma_s$,
$c_s(x)=\{m_*(x,s)+\widehat b_s\}/\sigma_s$, and
$F_t(q)=qg(a_sq-c_s(x))$.
The maximizer is $q^\dagger=a_s^{-1}\psi(c_s(x))$.
The exact relation
$c_s(x)-\widehat z=(a_s-1)\widehat z+
 \{m_*(x,s)-\widehat m(x,s)\}/\sigma_s$
gives
\[
 |q^\dagger-\psi(\widehat z)|
 \le\frac{L_\psi}{\widehat\sigma_s}|m_*-\widehat m|
 +\frac{L_\psi Z_{\max}+\Psi_{\max}}{\widehat\sigma_s}
       |\sigma_s-\widehat\sigma_s|
 \le C_{\rm LS}(\eta_m+\eta_\sigma).
\]
A degree-$d_{\rm act}$ Taylor comparator for $\psi$ on the current
bin has additional action error at most
$C_{\rm poly}h_T^{\beta-1}$.
Quadratic growth for $f_{c_s(x)}(a_sq)$ therefore bounds its
normalized loss by
$2L_ra_+C_{\rm LS}^2(\eta_m^2+\eta_\sigma^2)
 +Ch_T^{2\beta-2}$.
Successful coarse localization and the stated precision settings place
this comparator in the local coefficient body.

We next transfer the scalar loss to the true shadow. Fix $(x,s)$
and let $b$ range over the interval
$\cB\cap[b_s^*-\eta_0,b_s^*+\eta_0]$. Define the scalar branch value
\[
 z_b=\frac{m_*(x,s)+b}{\sigma_s},\qquad
 p_b=\sigma_s\psi(z_b)-b,\qquad
 W(b)=\sigma_s f_{z_b}\{\psi(z_b)\}
     =(p_b+b)D_s(p_b\mid x).
\]
At $b_s^*$, Assumption \ref{ass:utility-domain} identifies
$W(b_s^*)$ with the maximum over the full physical price interval.
If the local shadow interval is a singleton, there is no shadow
perturbation and the transfer bound follows directly from this
identity. Otherwise, we establish a smooth envelope on that interval.
The local domain contract places $z_b$ in $\mathcal Z^+$ and
$(\sigma_s/\widehat\sigma_s)\psi(z_b)$ in the interior of $\cQ$.
Thus $p_b$ is feasible and its residual has a neighborhood inside
$\mathcal I$. The first-order condition for the interior scalar maximizer is
\[
 g\{\psi(z_b)-z_b\}
 +\psi(z_b)g'\{\psi(z_b)-z_b\}=0.
\]
Because $\psi$ is Lipschitz, $W$ is absolutely continuous on the
local shadow interval. At almost every $b$, differentiation and the
first-order condition give
\begin{align*}
 W'(b)
 &=\psi'(z_b)\left[
       g\{\psi(z_b)-z_b\}
       +\psi(z_b)g'\{\psi(z_b)-z_b\}\right]
       -\psi(z_b)g'\{\psi(z_b)-z_b\}\\
 &=D_s(p_b\mid x).
\end{align*}
The right-hand side is continuous, so this identity holds throughout
the interval, with one-sided derivatives at its endpoints. Moreover,
$p_b$ is $(1+L_\psi)$-Lipschitz in $b$ and purchase probability is
$L_D$-Lipschitz between these safe prices. Hence $W'$ is
$L_V=L_D(1+L_\psi)$-Lipschitz. This envelope calculation does not
identify $W(b)$ with a full-interval maximum at perturbed shadows.

For a price $p=\widehat\sigma_s q-\widehat b_s$ proposed by the
pricing module, with $q\in\cQ$,
write $\widehat b=\widehat b_s$, $u=\widehat b-b_s^*$ and
$G=W(\widehat b)-(p+\widehat b)D_s(p\mid x)$.
Scalar quadratic growth gives
$G\ge\lambda_r|p-p_{\widehat b}|^2/(2\sigma_s)\ge0$.
The envelope remainder and the safe-price Lipschitz bound yield
\begin{align*}
 \Delta^*(x,s,p)
 &=G+W(b_s^*)-W(\widehat b)+uW'(\widehat b)
       +u\{D_s(p\mid x)-D_s(p_{\widehat b}\mid x)\}\\
 &\le G+\frac{L_V}{2}|u|^2
       +L_D|u|\,|p-p_{\widehat b}|.
\end{align*}
Young's inequality bounds the last term by
$G+\sigma_s L_D^2|u|^2/(2\lambda_r)$, and therefore
\[
 \Delta^*(x,s,p)
 \le2G+
 \left\{\frac{L_V}{2}
       +\frac{\overline\sigma L_D^2}{2\lambda_r}\right\}
           |\widehat b-b_s^*|^2.
\]
Since $G=\widehat\sigma_s\{F_t(q^\dagger)-F_t(q)\}$,
the displayed definition of $C_p$ absorbs the pilot-comparator
and shadow terms. Appendix \ref{app:pricing-certificate-proof}
controls the remaining bandit-learning loss.

\subsection{Whole-horizon aggregation}

\begin{proof}[Proof of Theorem \ref{thm:conditional-master}]
First construct the good event chronologically. The simultaneous
regression contract costs $\delta/8$; calibration counts and scale means
cost at most $3Ke^{-\ell}$. Proposition \ref{prop:warm-start}
costs at most $\delta/8$. Condition next on initialization of the pricing module
with valid pilot and warm-start outputs. Use the predictable
analytical extension in Appendix \ref{app:pricing-certificate-proof}:
it agrees with the policy until a first block-start local-shadow
exit and thereafter uses $b^*$. The fits and all pricing-module histories
are retained. If an optimizer flag would terminate learning,
use a fixed $q\in\cQ$ thereafter in this probability experiment;
on valid local domains this preserves feasibility and overlap.
Assign bounded-span continuation vectors if an empirical update
cannot be formed. All such replacements are predictable.

The single global pricing event costs at most $\delta_A=\delta/16$.
Lemma \ref{lem:joint-block} supplies block concentration under
successive block-start histories in the extended experiment,
before intersection with the pricing event. Its estimates need
no independence from the same observations that update the pricing module.
The delayed induction below excludes the first local exit on
these events, and the pricing event excludes an optimizer flag.
The extended and actual processes then agree throughout learning.
Neither probability argument conditions on eventual locality
or future optimizer success.

The input and action projections preserve this construction. On valid frozen
fits, Assumption \ref{ass:implementation-domains}(ii) places every
raw fitted input in $\mathcal Z$ before the first local-shadow
exit, so input projection is the identity on that stopped history.
The coarse localization event in Appendix
\ref{app:pricing-certificate-proof} places the entire symmetric
coefficient body inside $\cQ$, making action projection inactive as well.
For comparison, define an auxiliary guarded policy that uses the
same bounded fallback when a raw input is outside $\mathcal Z$ or
$\operatorname{dist}(\widetilde q_j,\partial\cQ)<\rho_{\rm loc}/4$,
and otherwise uses the unprojected action rule.
Couple it to the projected policy using the same optimizer seeds,
covariates, valuation noise and transition uniforms.
Their complete histories agree before a first differing decision;
the anchor rejection cannot occur and neither projection changes an
action on the good event. The chronological event construction and delayed
induction exclude such a difference on the good event. On its
complement, the retained physical-price checks preserve the $RT$
bound on the sum of conditional gaps, giving the same $RT\delta$
failure charge.

The active count tails are negligible at the declared starting
length: $r_b(L_0)\le r_{b,{\rm init}}\le H/8$ implies
$q\kappa_0(L_0-1)\ge4096\ell$.
Over at most $T$ blocks, their sum is at most $3KTe^{-\ell}$.
The prefix-mean events cost at most $8KT^2e^{-\ell}$.
All failure costs therefore sum to less than $\delta$, using
$e^{-\ell}=\delta/\{64K(T+1)^2\}$.
On the intersection, Proposition \ref{prop:delayed-stability}
applies inductively and no runtime failure is triggered.
The concentration bounds were proved conditional on successive
block-start histories. The induction is pathwise on their intersection
and does not condition the data on future success.

We next sum the deterministic errors on this event. Put
$w_k=L_0\varrho^k$, $x_k=\sqrt{w_k}W_k$ and
$d=\lambda\sqrt\varrho<1$. Then
\[
 x_{k+1}\le d x_k+\sqrt{\varrho w_k}\,u_k.
\]
The geometric convolution has summable kernel. On any finite
prefix its Euclidean sequence norm therefore satisfies
\[
 \|x\|_2\le\frac{\sqrt{L_0}W_0}{\sqrt{1-d^2}}
        +\frac{\sqrt\varrho}{1-d}\|\sqrt w\,u\|_2.
\]
This follows by separating the initial sequence $d^kx_0$ and
applying the triangle inequality to the shifted forcing sequences;
the sum of their kernel coefficients is $(1-d)^{-1}$.
Since $w_k\le L_k\le2w_k$, it gives
\begin{equation}
 \sum_k n_ke_{k-1}^2
 \le C\left\{L_0+\sum_{k\text{ complete}}L_ku_k^2\right\},
 \label{eq:weighted-value-energy}
\end{equation}
where $n_k$ is the actual number of calls in block $k$.
The first shadow index is charged separately to $L_0$.
For the terminal block, $n_k\le L_k\le\varrho L_{k-1}+1$,
so the same estimate applies. In particular, it requires
$\lambda\sqrt\varrho<1$, not the stronger condition
$\lambda\varrho<1$.

Let $s=\beta-1$ and
$z_T=\ell^{17}h_T^{2s}+\eta^2$. For $k\ge1$, the forcing bound,
$L_k\ge L_0\ge M_T$ and the preceding block's sampling radius
give
\[
 u_k\le C\left\{
 \ell^{17}\sqrt{M_T/L_k}+z_T
              +\ell/L_k+\sqrt{\ell/L_k}\right\}.
\]
The $k=0$ forcing is bounded by the precision gates and costs
at most $CL_0$ in \eqref{eq:weighted-value-energy}.
There are at most $C\log(eT)$ blocks. Their complete lengths
sum to at most $T$, and their reciprocal lengths form a
geometrically summable sequence. Consequently,
\begin{align}
 \sum_k n_ke_{k-1}^2
 &\le C\{L_0+\ell^{34}M_T\log(eT)
             +\ell^2\log(eT)+Tz_T^2\},\notag\\
 \sum_k n_kr_{b,k-1}^2
 &\le C\{L_0+\ell\log(eT)\}.
 \label{eq:persistent-shadow-energy}
\end{align}
The second bound follows directly from
$L_k/L_{k-1}\le\varrho+1/L_{k-1}$ and treats $k=0$ separately.

The persistent pricing event is used only once for aggregation.
If $N_A\le T$ is the total active-call count, then
\[
 \sum_{t\text{ active}}\Delta_t^*
 \le\Phi_T(N_A)+C_pN_A\eta^2
       +C_p\sum_k n_k(e_{k-1}+r_{b,k-1})^2.
\]
Thus the prefix bound is not summed again at every block.
For $h_T=c_hT^{-1/(4\beta-3)}$ and fixed structural constants,
\[
 M_T\le CT^{2\alpha-1},\quad
 \Phi_T(T)\le C\ell^{17}T^\alpha,\quad
 T\ell^{34}h_T^{4s}\le C\ell^{34}T^{2\alpha-1}.
\]
Together with $Tz_T^2\le
2T\ell^{34}h_T^{4s}+2T\eta^4$, these bounds account for
every term of \eqref{eq:conditional-master} on the good event.
Calibration and warm initialization cost at most
$R(n_u+\tau_{\rm w})$; $L_0$ also covers the initial value transient.
Outside this event, the sum of conditional gaps is at most $RT$
even on a fallback path. Taking expectations adds at most $RT\delta$.
Finally, Proposition \ref{prop:benchmark-bridge} adds at most $H$.
This proves \eqref{eq:conditional-master} uniformly in the initial state.
If the deterministic budget or precision gate fails, both oracle
and policy revenue lie in $[0,p_UT]$, which gives the fallback bound.

The same guarantee holds with the optional initialization count recovery
in Appendix \ref{app:operational-details}. Couple both policies using
the same covariates, noises, transition innovations and private random
streams. Their first possible difference is a nominal initialization
deadline with insufficient counts. Every such deadline passes on the
good event above, so no first difference occurs there. The initialization
durations and all good-event bounds are therefore unchanged. On the
complement, every executed price remains feasible and the conditional
gap is still at most $R$. The same $RT\delta$ charge and $H$ boundary
term complete the proof; no accuracy claim on recovery paths is needed.

\end{proof}

\section{The predictable convex-bandit module}
\label{app:predictable-bandit}

The refinement losses change with the current state and covariates.
We therefore use the adversarial, rather than the stochastic, version
of the online-Newton method of \citet{fokkema}. This appendix verifies
the information and concentration interfaces needed on this path.
All source labels below refer to arXiv:2406.06506v1. Gaussian
surrogate identities and their deterministic comparison inequalities
are used as in that source; the conditional concentration argument
is given here because the physical feedback is projected before it
enters the Gaussian method.

\subsection{Information, feedback and copy settings}

Consider a single local copy on a known convex body $\mathcal D$ in
$d=d_{\rm act}+1$ coefficient dimensions. In this appendix,
$\mathbb B_2$ denotes the Euclidean unit ball in $\R^d$.
An invertible affine transformation positions this body so that
$\mathbb B_2\subseteq\mathcal D\subseteq2(d+1)\mathbb B_2$.
Its gauge is $\pi_{\mathcal D}(x)=\inf\{r>0:x\in r\mathcal D\}$.
For a positive semidefinite matrix $M$, write
$\|v\|_M=(v^\top Mv)^{1/2}$.
At call $i$, let $\mathscr G_{i-1}$ contain all information available
before the copy draws its current seed. This includes observations
received while the copy was idle and the current covariates. The
convex loss $L_i:\mathcal D\to[0,1]$ is measurable at this time.
For the complete fresh seed $U_i$, put
$\mathscr A_i=\mathscr G_{i-1}\vee\sigma(U_i)$. The queried action
$A_i$ is $\mathscr A_i$-measurable and the observed feedback satisfies
\begin{equation}
 0\le\ell_i\le1,\qquad
 \E(\ell_i\mid\mathscr A_i)=L_i(A_i).
 \label{eq:bco-feedback-contract}
\end{equation}
Thus the physical noise $\varepsilon_i=\ell_i-L_i(A_i)$ is centered
after the complete seed and bounded by one. No independence of future
losses from past actions is required.

Write $n$ for the announced copy length and $\zeta$ for its failure
allocation. With sufficiently large universal constants fixed once,
set
\begin{align}
 \zeta_0&=\zeta/(64n^2),&
 \ell&=c_0\{1+\log(en d/\zeta_0)\},&
 \Lambda&=C_0\ell^2,\notag\\
 \lambda&=d^{-3}\Lambda^{-5},&
 \gamma&=(4d\Lambda)^{-1},&
 \eta&=\sqrt{d/(n\Lambda^3)},\notag\\
 \sigma_F^2&=(16d^2)^{-1},&
 F&=d^5\Lambda^8,&
 \epsilon_F&=d^{7/2}\Lambda^{17/2}/\sqrt n.
 \label{eq:bco-copy-tuning}
\end{align}
The symbol $\sigma_F$ is an optimizer setting, not a valuation scale.
If $\epsilon_F>1/2$, play the body's center throughout this copy.
Its regret against every fixed action is at most $n$, and
\begin{equation}
 n<2d^{7/2}\sqrt n\,\Lambda^{17/2}.
 \label{eq:bco-short-copy}
\end{equation}
Otherwise use Algorithm 1 of \citet{fokkema}, including its bonus
rule and convexified restart implementation, with the settings
\eqref{eq:bco-copy-tuning}. Quadratic iterate minimizations and
Gaussian draws use the exact real-arithmetic/oracle model; the
restart minimum may instead be certified within the tolerance
specified below. These are the optimization operations meant by
the certified model in Proposition \ref{prop:pooled-pricing}.
The guarantee does not cover an arbitrary floating-point
local solver. If an iterate, covariance or optimization certificate
cannot be formed, a stand-alone copy plays the center on its remaining
calls. Within \SHAPE, the copy reports the failure to the outer policy,
which applies its global fallback rule. The proof below bounds the
probability of this failure.

For clarity, write $\pi_+(x)=\max\{1,\pi_{\mathcal D}(x)/(1-\epsilon_F)\}$
and $v(x)=\{\pi_+(x)-1\}/\epsilon_F$. The copy draws
$G_i\sim N(\mu_i,\Sigma_i)$ conditional on $\mathscr G_{i-1}$,
queries $A_i=G_i/\pi_+(G_i)$, and constructs the observable return
\begin{align}
 Y_i^F&=\pi_+(G_i)\ell_i+2v(G_i),\notag\\
 f_i(x)&=\pi_+(x)L_i\{x/\pi_+(x)\}+2v(x).
 \label{eq:bco-extension}
\end{align}
Conditional on the full Gaussian seed, $\E(Y_i^F\mid\mathscr A_i)=f_i(G_i)$.
The noise of this extended return is $\pi_+(G_i)\varepsilon_i$,
so its concentration must account for the Gaussian projection factor.

\subsection{Gaussian identities and conditional concentration}

Let $p_i$ be the density of $N(\mu_i,\Sigma_i)$. Define
\begin{align}
 s_i(z)&=\E_{i-1}\left[(1-\lambda^{-1})f_i(G_i)
       +\lambda^{-1}f_i\{(1-\lambda)G_i+\lambda z\}\right],\notag\\
 R_i(z)&=\frac{p_i\{(G_i-\lambda z)/(1-\lambda)\}}
                 {(1-\lambda)^d p_i(G_i)},\qquad
 \widehat s_i(z)=Y_i^F\{1-\lambda^{-1}+\lambda^{-1}R_i(z)\},
 \label{eq:bco-surrogate}
\end{align}
where $\E_{i-1}$ conditions on $\mathscr G_{i-1}$.
Conditional subexponential norms below use the convention
\[
 \|Z\|_{\psi_1}
 =\inf\{c>0:\E_{i-1}\exp(|Z|/c)\le2\ \text{almost surely}\}.
\]
Conditional change of variables gives
$\E_{i-1}\widehat s_i(z)=s_i(z)$ and the corresponding gradient
and Hessian identities. These identities hold round by round for
predictable $f_i$; they do not condition on future inputs. Put
$g_i=\nabla\widehat s_i(\mu_i)$, $H_i=\nabla^2\widehat s_i(\mu_i)$ and
\[
 \widehat q_i(z)=g_i^\top(z-\mu_i)
           +\tfrac14(z-\mu_i)^\top H_i(z-\mu_i),\qquad
 q_i=\E_{i-1}\widehat q_i.
\]
The actual covariance recursion is
\begin{equation}
 \Sigma_{i+1}^{-1}
 =(1-2\gamma_i)\Sigma_i^{-1}+\tfrac\eta2H_i,
 \label{eq:bco-correct-covariance}
\end{equation}
with $\gamma_i\in\{0,\gamma\}$. Write $\bar\Sigma_i^{-1}$ for the
same recursion with $H_i$ replaced by its conditional mean
$\E_{i-1}H_i=\nabla^2s_i(\mu_i)$ and with initialization
$\bar\Sigma_1=\Sigma_1=\sigma_F^2I$.

We work in the source stopped experiment (Definition 7), stopping
before a covariance comparison, focus or restart condition fails.
Also stop analytically if the number of nonzero bonuses exceeds
$d\Lambda$. Up to that first possible stop,
$w_i=\prod_{j\le i}(1-2\gamma_j)\in[1/2,1]$.
The basic covariance envelope there is
$\Sigma_i\preceq4\sigma_F^2I\preceq I$.
There is also a deterministic upper spectral envelope. If
$L_f=Cd/\epsilon_F$ is the source extension's Lipschitz bound,
the Gaussian Hessian identity and stopped covariance comparison give
\[
 A\le\sigma_F^{-2}+C\eta n\lambda L_f\sqrt{dA},
 \qquad A=\max_{i\le\tau+1}\|\bar\Sigma_i^{-1}\|.
\]
Solving this scalar quadratic and using the comparison once more
gives
\begin{equation}
 \|\Sigma_i^{-1}\|\le
 \Gamma:=4\{\sigma_F^{-1}+C\eta n\lambda L_f\sqrt d\}^{2}
 \le C(d^2+n^2).
 \label{eq:bco-spectral-envelope}
\end{equation}
The source bonus-count argument now gives at most
$Cd\log\{en d\Gamma/(\lambda\epsilon_F)\}\le Cd\ell<d\Lambda$
nonzero bonuses. Thus the extra analytical stop cannot occur when
the other comparison events hold.
This envelope controls the spatial nets used below. Write
$\mathcal D_{\epsilon_F}=(1-\epsilon_F)\mathcal D$ for the shrunken
body. The source focus set through a stopping time $\tau$ is
\[
 K_\tau=\mathcal D_{\epsilon_F}\cap
 \bigcap_{i\le\tau}\{z:\|z-\mu_i\|_{\Sigma_i^{-1}}^2\le F\}.
\]
Every $z\in K_\tau$ therefore satisfies
\begin{equation}
 \lambda\|z-\mu_i\|_{\Sigma_i^{-1}}\le r_F,
 \qquad r_F=\lambda\sqrt F=(\sqrt d\,\Lambda)^{-1},
 \quad i\le\tau.
 \label{eq:bco-small-focus}
\end{equation}
Write $V_\tau=\sum_{i\le\tau}\E_{i-1}(Y_i^F)^2$.

\begin{lemma}[Concentration after projected feedback]
\label{lem:bco-amplified-concentration}
In this stopped experiment there is a universal $C$ such that the
following bounds hold, with conditional failure probability at most
$6n\zeta_0$, simultaneously over $\tau\le n$. The constant $c_0$
is chosen so that each of the six families below costs at most
$n\zeta_0$. Put
$Y_*=C\ell/\epsilon_F$ and
\[
 B_\tau=1+\frac C\lambda
       \{\sqrt{d\ell V_\tau}+d\ell Y_*\},\qquad
 D_\tau=C\lambda\ell^2
       \{1+\sqrt{dV_\tau}+d^2Y_*\}.
\]
Then
\begin{align}
 \sup_{z\in K_\tau}
 \left|\sum_{i\le\tau}(\widehat s_i-s_i)(z)\right|
 &\le B_\tau,\notag\\
 \left|\sum_{i\le\tau}
       \{\widehat s_i(\mu_i)-s_i(\mu_i)\}\right|
 &\le B_\tau,\notag\\
 \sup_{z\in K_\tau}
 \left|\sum_{i\le\tau}(\widehat q_i-q_i)(z)\right|
 &\le B_\tau.
 \label{eq:bco-scalar-concentration}
\end{align}
For every $M\succeq\Sigma_i^{-1}$, $i\le\tau$,
\begin{equation}
 -D_\tau M\preceq
 \sum_{i\le\tau}\{\nabla^2\widehat s_i(z)-\nabla^2s_i(z)\}
 \preceq D_\tau M
 \label{eq:bco-spatial-hessian}
\end{equation}
uniformly over $z\in K_\tau$. The same bound, up to a universal
factor, holds for Hessians evaluated at the predictable $\mu_i$
and for any one specified predictable weight process in $[1,2]$.
The statement does not take a supremum over weight processes.
\end{lemma}

\begin{proof}
All expectations in this proof are conditional before the current
Gaussian seed. Put $W_i=\Sigma_i^{-1/2}(G_i-\mu_i)$.
Gaussian norm concentration and the Lipschitz property of $\pi_+$
give common events with failure at most $C\zeta_0$ per round on
which $\|W_i\|\le C\sqrt{d\ell}$ and $\pi_+(G_i)\le C\sqrt\ell$.
Because $0\le\ell_i\le1$, these events imply
$0\le Y_i^F\le Y_*$. They are common to all spatial and directional
arguments and are not unioned over a net. Gaussian moments and
$\pi_+(G_i)\le1+2\|G_i-\mu_i\|$ also give a polynomial conditional
fourth-moment envelope for $Y_i^F$.

For $u=\Sigma_i^{-1/2}(z-\mu_i)$, the ratio in
\eqref{eq:bco-surrogate} is
\[
 R_i(z)=(1-\lambda)^{-d}
 \exp\left\{-\frac{\|W_i-\lambda u\|^2}{2(1-\lambda)^2}
                    +\frac{\|W_i\|^2}2\right\}.
\]
On the common event and the slightly enlarged focus region
$\lambda\|u\|\le2r_F$, this ratio is bounded by a universal
constant. The multiplier of $Y_i^F$ in $\widehat s_i(z)$ is
therefore at most $C/\lambda$. Direct differentiation shows that
the multiplier in $\widehat q_i(z)$ is bounded by
\[
 C\{|W_i^\top u|+\lambda[(W_i^\top u)^2+\|u\|^2]\}
 \le C/\lambda.
\]
The last inequality uses
$r_F\sqrt{d\ell}\le C$ and $r_F^2d\ell\le C$.
Thus each commonly truncated scalar score has conditional variance
at most $C\lambda^{-2}\E_{i-1}(Y_i^F)^2$ and magnitude at most
$CY_*/\lambda$. Bernstein's martingale inequality gives the first
and third bounds in \eqref{eq:bco-scalar-concentration} after the
spatial union described below. The moving-reference bound uses the
same scalar argument directly at the predictable $z_i=\mu_i$.
A spatial supremum alone would not imply this reference bound.

The Hessian requires separate variance and magnitude estimates.
For a fixed direction $a$, differentiation gives
\begin{align*}
 a^\top\nabla^2\widehat s_i(z)a
 &=Q_{i,a}(z)Y_i^F,\\
 Q_{i,a}(z)
 &=\frac{\lambda R_i(z)}{(1-\lambda)^2}
 \left\{\frac{[a^\top\Sigma_i^{-1/2}(W_i-\lambda u)]^2}
                    {(1-\lambda)^2}
                  -\|a\|_{\Sigma_i^{-1}}^2\right\}.
\end{align*}
Let $q_{i,a}=\lambda\|a\|_{\Sigma_i^{-1}}^2$ and truncate
$Q_{i,a}$ on the common Gaussian event, obtaining $\bar Q_{i,a}$.
The one-dimensional Gaussian quadratic in this formula implies
the conditional subexponential bound
$\|\bar Q_{i,a}/q_{i,a}\|_{\psi_1}\le C$.
Splitting a conditional expectation at $Cq_{i,a}\ell$ yields
\begin{align}
 \E_{i-1}[\bar Q_{i,a}^{\,2}(Y_i^F)^2]
 &\le Cq_{i,a}^2\ell^2\E_{i-1}(Y_i^F)^2\notag\\
 &\quad+
 \{\E_{i-1}(Y_i^F)^4\}^{1/2}
 \{\E_{i-1}[\bar Q_{i,a}^{\,4}
      \one_{\{|\bar Q_{i,a}|>Cq_{i,a}\ell\}}]\}^{1/2}\notag\\
 &\le Cq_{i,a}^2\ell^2\E_{i-1}(Y_i^F)^2+q_{i,a}^2n^{-12}.
 \label{eq:bco-variance-preserved}
\end{align}
Choose $c_0$ in \eqref{eq:bco-copy-tuning} to dominate the fixed
polynomial fourth-moment envelope. This split is inside an
expectation: it does not assert that the threshold $Cq_{i,a}\ell$
holds for every direction simultaneously. For a uniform magnitude
bound use the larger threshold $Cq_{i,a}d\ell$, or the common
Gaussian norm bound, and $Y_*$. Conditional centering then costs
only a universal factor. If
$\Gamma_a=\max_{i\le\tau}\|a\|_{\Sigma_i^{-1}}^2$,
Bernstein with a net logarithm $O(d\ell)$ gives
\[
 C\lambda\Gamma_a\ell^2
       \{1+\sqrt{dV_\tau}+d^2Y_*\}.
\]
Dyadic peeling over the stopped polynomial spectral range handles
$\Gamma_a$ and $V_\tau$.

Here are the uniformity and truncation details. Set
$\mathfrak P=2+n+d+\Gamma+\lambda^{-1}+\epsilon_F^{-1}+\zeta_0^{-1}$.
At copy initialization take deterministic spatial nets of the
bounded body and directional nets of the unit sphere with mesh
$\mathfrak P^{-100}$. Their total logarithmic cardinality is
$O(d\log\mathfrak P)=O(d\ell)$. On the common Gaussian event,
the displayed ratio and its first three spatial derivatives are
bounded by $C\mathfrak P^{10}$ on the enlarged focus region;
this follows by differentiating its exponential quadratic and
using \eqref{eq:bco-spectral-envelope} and $Y_i^F\le Y_*$.
The same polynomial bound applies to conditional derivatives by
Gaussian integration. Stop a spatial net point
before it leaves the enlarged focus region. Every point of the
final focus set has a nearby net point still in that region at
all preceding rounds, by \eqref{eq:bco-small-focus}. Differentiating
the displayed ratio gives these derivative bounds there, so
interpolation errors are at most $n^{-3}$ (and smaller than the
corresponding spectral tolerance for directional interpolation).
Gaussian ratio moments outside the common event are
polynomial uniformly on the enlarged focus region. Conditional
Cauchy--Schwarz therefore bounds every discarded-tail compensator
by an arbitrary inverse polynomial when $c_0$ is large enough.
This is a deterministic uniform bound, not a net union of tail
events. The errors are absorbed by the displayed additive constants.
The same construction applies to the scalar scores above.

Finally, $\Gamma_a\le a^\top Ma$ for every allowed $M$.
Interpolation of the fixed-direction inequalities proves
\eqref{eq:bco-spatial-hessian} without a data-dependent matrix net.
Predictable weights in $[1,2]$ change the variances and magnitudes
by at most fixed factors. Evaluation at $\mu_i$ is predictable and
requires no spatial net. Union over the finitely many score
families and endpoints gives the stated probability.
\end{proof}

\subsection{Transfer of the online-Newton guarantee}

\begin{proposition}[Predictable convex-bandit guarantee]
\label{prop:predictable-bco}
Under \eqref{eq:bco-feedback-contract}, the copy specified above has,
conditionally on its initialization and with probability at least
$1-\zeta$, the simultaneous comparator guarantee
\begin{equation}
 \sup_{a\in\mathcal D}\sum_{i=1}^n\{L_i(A_i)-L_i(a)\}
 \le C d^{7/2}\sqrt n\{\log(en d/\zeta)\}^{17}.
 \label{eq:bco-predictable-guarantee}
\end{equation}
It uses one observed feedback per call. In
\eqref{eq:pricing-primitive-interface}, one may consequently take
$p_B=17$ and $C_B=C d^{7/2}(1+\log d)^{17}$.
\end{proposition}

\begin{proof}
The short-copy assertion follows from \eqref{eq:bco-short-copy}.
Consider a nonshort copy. We give the source interfaces, including
the moving-reference and restart-comparator corrections, before
combining its deterministic inequalities.

First, the square-return argument of source Lemma 10 remains
conditional roundwise. Indeed,
$\E_{i-1}(Y_i^F)^2\le C\{1+\E_{i-1}v(G_i)^2\}$.
The source Gaussian convex-moment inequality and its deterministic
trace/log-determinant calculation give, until the stopping time,
\[
 \sum_{i\le\tau}\E_{i-1}v(G_i)^2
 \le Cn+
 \frac{C\sigma_F d^{3/2}\Lambda}{\epsilon_F\lambda\eta}
 \le Cn.
\]
This step uses $f_i''\succeq v''$ for each realized convex
extension, not a common loss across rounds. The deterministic
Gaussian smoothing tails are controlled at analytic tolerance
$\rho=\zeta_0$; their accumulated remainders are included in the
constant in this display. Truncating the
nonnegative $v(G_i)^2$ at $Cd\sigma_F^2\ell/\epsilon_F^2$ and
applying the scalar exponential-supermartingale inequality also
gives, with the allocated failure,
\begin{equation}
 \sum_{i\le\tau}(Y_i^F)^2\le Cn\Lambda^2,
 \qquad V_\tau\le Cn\Lambda^2.
 \label{eq:bco-square-budget}
\end{equation}
The truncation remainder costs
$Cd\sigma_F^2\ell^2/\epsilon_F^2\le Cn$.
All preceding identities hold with $\mathscr G_{i-1}$ conditioning.

Apply Lemma \ref{lem:bco-amplified-concentration} in this stopped
experiment. Its scalar bounds control both the uniform surrogate
regret error and the quadratic error by
\[
 C_\tau=\frac{C\Lambda}{\lambda}
             \{\sqrt{dV_\tau}+\sqrt n\}
 \le\frac{C\Lambda^2}{\lambda}\sqrt{nd}.
\]
The surrogate regret error includes the moving-reference sum in
\eqref{eq:bco-scalar-concentration} as well as the spatial error.
For the covariance event, use the weighted Hessian inequality
with $1/w_i\in[1,2]$, where $w_i=\prod_{j\le i}(1-2\gamma_j)$.
The bonus rule makes $\gamma_i$ predictable, even though its line
is printed after feedback in the source pseudocode. The rule has
at most $d\Lambda$ nonzero bonuses, so $w_i\in[1/2,1]$.
Substitution into \eqref{eq:bco-correct-covariance} gives the
source comparison
$\tfrac12\bar\Sigma_{\tau+1}^{-1}\preceq\Sigma_{\tau+1}^{-1}
\preceq\tfrac32\bar\Sigma_{\tau+1}^{-1}$.
This is the comparison for the conditional-mean recursion defined above.

The spatial Hessian event also certifies the restart calculation.
For a sufficiently large fixed $C_Q$, the known quadratic
\[
 Q_\tau(z)=C_Q\lambda\Lambda^3\sqrt{nd}
                  \|z-\mu_\tau\|_{\Sigma_\tau^{-1}}^2
\]
makes $\sum_{i\le\tau}\widehat s_i(z)+Q_\tau(z)$ convex on
$K_\tau$. This follows from
$\Sigma_i^{-1}\preceq8\Sigma_\tau^{-1}$ before stopping and
$D_\tau\le C\lambda\Lambda^3\sqrt{nd}$.
Its minimum is obtained in the source certified convex-oracle
model. Use the regularized restart statistic
\[
 \widehat R_\tau^{\,Q}=
 \eta\left[\sum_{i\le\tau}\widehat s_i(\mu_i)
 -\min_{z\in K_\tau}
       \left\{\sum_{i\le\tau}\widehat s_i(z)+Q_\tau(z)\right\}\right]
\]
and restart when it is at most $-\gamma F/32$.
For certified approximate minima require error at most
$\gamma F/(4096\eta)$ and charge
$E_{\rm opt}=\gamma F/4096$ below; for exact minima set
$E_{\rm opt}=0$.
All stored objectives depend only on known Gaussian densities
and observed feedback, not on evaluations of $L_i$ elsewhere.

We next record enough slack for this implementation. Let
$E_\tau^{\rm grad}=\sum_{i\le\tau}\|g_i\|_{\Sigma_{i+1}}^2$ and let
$B_{\rm tr}\le C_{\rm tr}d\Lambda/\lambda$ be the deterministic
smoothing/trace conversion term from source Lemmas 24, 26 and 11,
which give surrogate optimism, Gaussian smoothing and the trace bound.
This conversion term includes their deterministic smoothing-tail
remainders. Quadratic optimism has its own additive tolerance;
reserve $E_{\rm sm}=C\eta n\rho/\lambda^2$ for its cumulative
use in the FTRL and boundary arguments. In the nonshort branch,
\[
 n\rho/\lambda^2
 =\zeta d^6\Lambda^{10}/(64n)
 \le\zeta/(256d\Lambda^7).
\]
Specifically, put
\[
 R^f_\tau(a)=\sum_{i\le\tau}
       \{\E_{i-1}f_i(G_i)-f_i(a)\},\qquad
 R^s_\tau(a)=\sum_{i\le\tau}\{s_i(\mu_i)-s_i(a)\}.
\]
Optimism and the Gaussian smoothing comparison give
$\eta R^f_\tau(a)\le\eta R^s_\tau(a)+B_{\rm tr}$ for every
$a$ in the shrunken body. This comparison is uniform and is
valid before selecting either hindsight minimizer.
The Gaussian event and \eqref{eq:bco-square-budget} give
$E_\tau^{\rm grad}\le Cnd\Lambda^3\le nd\Lambda^4/2$.
For a sufficiently large single choice of $C_0$,
\begin{align}
 \eta D_\tau&\le1/64,\notag\\
 \frac{\operatorname{diam}(\mathcal D)^2}{\sigma_F^2}
       +2\eta^2E_\tau^{\rm grad}+3\eta C_\tau+E_{\rm sm}
       &\le\gamma F/192,\notag\\
 B_{\rm tr}+2\eta C_\tau+\eta S
                    +\eta\max_{K_\tau}Q_\tau+E_{\rm sm}+E_{\rm opt}
       &\le\gamma F/64,
 \qquad S=C\,d^2/\epsilon_F.
 \label{eq:bco-slack-inequalities}
\end{align}
The first four ratios in the last inequality are bounded
by $C/\Lambda$, $C/\Lambda^{3/2}$,
$C/(d\Lambda^{10}\gamma F)$ and $C/(d\Lambda^{5/2})$,
respectively. The diameter, gradient and scalar-error ratios
in the preceding line are bounded by
$C\Lambda^{-7}$, $C\Lambda^{-6}$ and $C\Lambda^{-3/2}$.
Thus these requirements fix universal tuning constants and do
not introduce an unknown model input.

It remains to check the comparator at restarts. Write $x^s_j$ for
a minimizer of $\sum_{i\le j}s_i$ over the shrunken body.
Before the first stopped round, $x^s_{\tau-1}\in K_\tau$.
The source FTRL, bonus and optimism inequalities are pathwise.
With \eqref{eq:bco-slack-inequalities} they show that a segment
without a restart has extended regret at most $C\gamma F/\eta$.
More explicitly, if $x^s_\tau\in K_\tau$ and there is no restart,
the source FTRL and optimism comparisons give
\[
 \tfrac12\|x^s_\tau-\mu_{\tau+1}\|_{\Sigma_{\tau+1}^{-1}}^2
 \le A_0-\eta\sum_{i\le\tau}\{s_i(\mu_i)-s_i(x^s_\tau)\}
 \le A_0+\gamma F/32+\eta C_\tau+E_{\rm opt},
\]
where $A_0=\operatorname{diam}(\mathcal D)^2/\sigma_F^2
+2\eta^2E_\tau^{\rm grad}+\eta C_\tau+E_{\rm sm}$.
The slack makes this smaller than $F/2$, preserving the next
focus condition. It also bounds maximal surrogate regret and,
after the trace conversion, maximal extended regret.
If $x^s_\tau$ leaves $K_\tau$, the restricted surrogate minimum
has a boundary representative $y$. Source Lemma 12 gives bonus
at most $-\gamma F/24$ there.
Consequently the restart statistic is at most
\[
 \gamma F/192-\gamma F/24+E_{\rm opt}
 \le-7\gamma F/192+\gamma F/4096<-\gamma F/32.
\]
Regularization can only decrease that statistic, so the
computable rule detects this escape.

For a restarted segment, use consecutive surrogate minimizers,
not an unproved focus property of an extended-loss minimizer.
Optimality at $\tau-1$ gives
\[
 \sum_{i\le\tau}s_i(x^s_{\tau-1})
       -\sum_{i\le\tau}s_i(x^s_\tau)
 \le s_\tau(x^s_{\tau-1})-s_\tau(x^s_\tau)\le S.
\]
Indeed, the source extension bound implies
$\operatorname{Lip}(s_\tau)\operatorname{diam}(\mathcal D)
\le C d^2/\epsilon_F$. The deterministic smoothing inequality
first bounds extended regret by maximal surrogate regret plus
$B_{\rm tr}/\eta$. The last display then replaces that global
surrogate minimizer by $x^s_{\tau-1}\in K_\tau$.
Uniform concentration and regularization therefore bound the
maximal extended regret by
\[
 \eta^{-1}\left\{
 \widehat R_\tau^{\,Q}+B_{\rm tr}
       +2\eta C_\tau+\eta S+\eta\max_{K_\tau}Q_\tau
       +E_{\rm sm}+E_{\rm opt}\right\}<0.
\]
This proves nonpositive regret against every fixed comparator on
each restarted segment. It avoids identifying a minimizer of the
extended losses with a minimizer of the surrogate losses.

Condition at each restart history and allocate the concentration
events at every endpoint. There are six concentration families
in Lemma \ref{lem:bco-amplified-concentration}, one common
Gaussian/feedback event and one square-return event. Choose
$c_0$ so that each family costs at most $n\zeta_0$ per segment.
There are at most $n$ segments, so the total is at most
$8n^2\zeta_0=\zeta/8$. The fallback conditions cannot fail on
this event: covariance positivity and the focus induction hold,
and the regularized programs have the stated certificates.
The final nonrestarted segment contributes at most
$C\gamma F/\eta$; all earlier segments contribute nonpositive
regret. Finally, conditional Hoeffding for the bounded physical
losses and the source extension comparison give, uniformly in
the comparator,
\[
 \sum_{i=1}^n\{L_i(A_i)-L_i(a)\}
 \le C\sqrt{n\ell}+n\epsilon_F+C\gamma F/\eta.
\]
Both latter terms are at most
$Cd^{7/2}\sqrt n\,\Lambda^{17/2}$. Substitution of
\eqref{eq:bco-copy-tuning}, with the remaining failure allocation
for this last martingale, proves
\eqref{eq:bco-predictable-guarantee}.
\end{proof}

\section{The persistent pooled pricing certificate}
\label{app:pricing-certificate-proof}

This appendix proves the map-learning certificate for the single
pooled pricing module initialized after warm start. Its bins, coarse anchors,
local-copy counts and refinement histories persist across all outer
blocks. Only the predictable shadow input changes at block boundaries.
The certificate controls learning of the shared map; transfer to the
true Bellman loss is a separate, pointwise inequality.

We use Proposition \ref{prop:predictable-bco} conditionally at the
initialization of each local copy. For normalized losses and feedback
in $[0,1]$, write its fixed-horizon interface as
\begin{equation}
 \sum_{i=1}^n\{L_i(A_i)-L_i(a)\}
 \le C_B\sqrt n\{\log(en/\zeta)\}^{17}.
 \label{eq:pricing-primitive-interface}
\end{equation}
The conditional failure probability is at most $\zeta$, simultaneously
over comparators $a$ in the copy's body. Here
$C_B=C d^{7/2}(1+\log d)^{17}$, $d=d_{\rm act}+1$, is supplied by
Appendix \ref{app:predictable-bandit}; thus $p_B=17$ throughout.
The loss at a call is fixed before its fresh seed, and feedback is
conditionally unbiased after that complete seed. The short-copy
convention and exact real-arithmetic/certified-oracle model are those
of that appendix.

\subsection{A single initialization and a finite-budget bound}

Fix the global budget $T\ge H_{\rm act}$ when the pricing module is initialized.
Abbreviate
\[
 \begin{gathered}
 h=h_T=\min\{D_{\mathcal Z},\ell_{\rm loc},c_hT^{-\kappa_\beta}\},
 \qquad \kappa_\beta=\frac1{4\beta-3},\\
 M=M_T=\lceil D_{\mathcal Z}/h\rceil,\qquad
 G=|\mathcal G_{\cQ}|,\quad Q=Q_{\max},
 \quad J_T=1+\lceil\log_2T\rceil.
 \end{gathered}
\]
Partition the full input envelope into these $M$ fixed bins, denoted by
$B_j$ with closures $\overline B_j$. Each bin
uses successive refinement-copy budgets $1,2,4,\ldots$, without
restarting at outer-block boundaries. Allocate confidence only once:
\begin{equation}
 \begin{gathered}
 \delta_A=\frac\delta{16},\qquad
 \zeta_T=\frac{\delta_A}{4MTJ_T},\qquad
 L_c=\log(eTMG/\delta_A),\\
 L_B=\log(eT/\zeta_T)
     =\log(4eT^2MJ_T/\delta_A).
 \end{gathered}
 \label{eq:pricing-copy-allocation}
\end{equation}
The coarse quota for each bin/grid pair is
$m_{\rm coarse}=\lceil m_0L_c\rceil$. A copy is updated only on its
own bin's refinement calls. Observations received while it is idle
remain part of the chronological information at its next call.

\begin{proposition}[All-prefix certificate for the persistent pricing module]
\label{prop:full-cover-pricing-certificate}
Condition on the pricing module's initialization information, with frozen
$\widehat m,\widehat\sigma$ satisfying the local domain and precision
conditions of Appendix \ref{app:operational-details} and actual error
bounds $\eta_m,\eta_\sigma$. Let the shadow vectors be predictable,
constant within each outer block, projected into $\cB$, and locally admissible:
$|\widehat b_{t,s}-b_s^*|\le\eta_0$ for all states and pricing calls.
Index pricing calls chronologically by $t\le T$, and put
\[
 \begin{gathered}
 a_s=\widehat\sigma_s/\sigma_s,\quad
 c_t=\{m_*(X_t,S_t)+\widehat b_{t,S_t}\}/\sigma_{S_t},\\
 F_t(q)=qg(a_{S_t}q-c_t),\qquad
 q_t^\dagger=a_{S_t}^{-1}\psi(c_t),\\
 \Gamma_t=2\widehat\sigma_{S_t}
              \{F_t(q_t^\dagger)-F_t(q_t)\}\ge0.
 \end{gathered}
\]
The physical comparison price
$p_t^\dagger=\widehat\sigma_{S_t}q_t^\dagger-\widehat b_{t,S_t}$
is the scalar branch price in \eqref{eq:adjusted-pricing-loss}.
Thus the two definitions of $\Gamma_t$ agree.
Run the single persistent pricing module with
\eqref{eq:pricing-copy-allocation}. There is an event of conditional
probability at least $1-\delta_A$ on which no pricing-module or domain
flag occurs and, simultaneously for all realized pricing prefixes
$1\le N\le T$,
\begin{align}
 \sum_{t=1}^N\Gamma_t
 \le {}&4\overline\sigma QG(m_0L_c+1)M\notag\\
 &+4\overline\sigma Q(2+\sqrt2)C_BL_B^{17}\sqrt{MN}\notag\\
 &+2\overline\sigma L_ra_+L_\psi^2N h^{2\beta-2}
   +C_pN(\eta_m^2+\eta_\sigma^2).
 \label{eq:pricing-finite-budget-bound}
\end{align}
The probability statement conditions only on information at initialization,
not on future shadows, states, covariates or bin counts.
Moreover, at each locally admissible pricing call,
\begin{equation}
 \Delta^*(X_t,S_t,p_t)
 \le\Gamma_t+C_p|\widehat b_{t,S_t}-b_{S_t}^*|^2.
 \label{eq:pricing-pointwise-shadow-transfer}
\end{equation}
For a process that may leave the local shadow domain, the same
probability guarantee controls every actual prefix strictly before
its first local exit, as formalized at the end of the proof.
\end{proposition}

\begin{proof}
Fix an initialization history satisfying the premises. The fits,
full-range bins and coarse schedule are now fixed, but future shadows
need not be. For a locally admissible triple $(x,s,b)$, write
\[
 \widehat z(x,s,b)=\{\widehat m(x,s)+b\}/\widehat\sigma_s,
 \quad c(x,s,b)=\{m_*(x,s)+b\}/\sigma_s,
 \quad F_{x,s,b}(q)=qg\{a_sq-c(x,s,b)\}.
\]
The local domains give $c(x,s,b)\in\mathcal Z^+$,
$a_s\cQ\subset\cQ^+$, and
$q^\dagger(x,s,b)=a_s^{-1}\psi\{c(x,s,b)\}$ in the interior of $\cQ$,
with distance at least $\rho_Q$ from its boundary. The exact identity
$F_{x,s,b}(q)=a_s^{-1}f_{c(x,s,b)}(a_sq)$ gives
\begin{equation}
 \frac{\lambda_ra_s}{2}|q-q^\dagger|^2
 \le F_{x,s,b}(q^\dagger)-F_{x,s,b}(q)
 \le\frac{L_ra_s}{2}|q-q^\dagger|^2.
 \label{eq:pricing-transformed-growth}
\end{equation}
In particular every transformed gap is nonnegative. We first obtain
anchors valid for all locally admissible future inputs, then construct
a fixed polynomial comparator in each bin, and aggregate local-copy
regret over the single global pricing history.

\paragraph{Coarse localization at actual occurrences.}
For bin $j$ and grid action $q$, let $I_t^{j,q}$ indicate that the
coarse schedule requests this pair. Once the pair's quota is complete,
write $\widehat F_j(q)$ for its mean observed payoff $qY_t$.
Immediately before the purchase,
the indicator, the current state, covariates, shadow and $q$ are known.
Thus
\[
 I_t^{j,q}\{qY_t-F_t(q)\}
\]
is a martingale difference bounded in absolute value by $2Q$.
Stop its exponential supermartingale when the pair's count reaches
$m_{\rm coarse}$, or at $T$ if it does not. For
$u=4Q/\sqrt{m_0}$, the bounded-increment estimate gives
\[
 \Pp\left(\left|
 \widehat F_j(q)-\frac1{m_{\rm coarse}}
       \sum_{i=1}^{m_{\rm coarse}}F_{\tau_i}(q)
 \right|>u,\ \tau_{m_{\rm coarse}}\le T
 \right)
 \le2\exp\{-2m_{\rm coarse}/m_0\}.
\]
Here $\tau_i$ are this pair's actual occurrences, possibly in different
outer blocks. No conditioning on their future count has been made. Since
$m_{\rm coarse}\ge m_0L_c$, a union over the $MG$ pairs costs at most
\[
 2MG e^{-2L_c}
 =\frac{2\delta_A^2}{e^2T^2MG}\le\delta_A/4.
\]
Call the resulting event $\cC_T$. Incomplete pairs impose no
concentration obligation and their bins do not enter refinement.

Changing shadows do not add a within-bin drift term. Indeed, the exact
identity
\[
 c(x,s,b)=a_s\widehat z(x,s,b)
             +\{m_*(x,s)-\widehat m(x,s)\}/\sigma_s
\]
holds for every locally admissible $b$. For any two such triples whose
fitted inputs lie in the same bin, it gives
\begin{align*}
 |a_s-a_{s'}|&\le2\eta_\sigma/\underline\sigma,\\
 |c(x,s,b)-c(x',s',b')|
 &\le a_+h+2(\eta_m+Z_{\max}\eta_\sigma)/\underline\sigma.
\end{align*}
Their residuals lie in $\mathcal I$. The declared Lipschitz bound
for $g$ and the definition of $C_{\rm drift}$ therefore imply
\begin{equation}
 \sup_{q\in\cQ}|F_{x,s,b}(q)-F_{x',s',b'}(q)|
 \le C_{\rm drift}(h+\eta_m+\eta_\sigma).
 \label{eq:pricing-within-bin-drift}
\end{equation}
On $\cC_T$, every completed empirical grid objective is within
\[
 u_T=C_{\rm drift}(h+\eta_m+\eta_\sigma)
                       +4Q/\sqrt{m_0}\le A_{\rm loc}
\]
of every locally admissible objective in that bin, including those
with shadow values not yet used. Let $q_g$ be a nearest grid point to
an arbitrary such objective's maximizer. Empirical
maximization and \eqref{eq:pricing-transformed-growth} yield
\[
 \frac{\lambda_ra_s}{2}|\widetilde q_j-q^\dagger|^2
 \le \frac{L_ra_s}{2}\operatorname{mesh}(\mathcal G_{\cQ})^2
       +2u_T.
\]
The grid and coarse gates consequently give
\begin{equation}
 \sup_{\substack{x,s,b:\ |b-b_s^*|\le\eta_0,\ b\in\cB,\\
                    \widehat z(x,s,b)\in\overline B_j}}
 |\widetilde q_j-q^\dagger(x,s,b)|^2
 \le\left(\frac1{256}+\frac4{1024}\right)\rho_{\rm loc}^2
 =\frac{\rho_{\rm loc}^2}{128}
 <\frac{\rho_{\rm loc}^2}{64}.
 \label{eq:pricing-uniform-anchor}
\end{equation}
The supremum concerns the fixed model, frozen fits and observed
anchor, not realized future occurrences. Its event is therefore
measurable when bin $j$ first enters refinement. Although not
observable by the policy, the supremum over this fixed set of
possible maximizers is a continuous function of the observed anchor.
In particular, the anchor has a uniform interior margin:
\[
 \operatorname{dist}(\widetilde q_j,\partial\cQ)
 \ge\rho_Q-\rho_{\rm loc}/8\ge\rho_{\rm loc}/4.
\]
Every polynomial in the symmetric coefficient trust region therefore
takes values in $\cQ$, making action projection inactive for the whole
body on this event.

\paragraph{A fixed comparator and known body positioning.}
Let $T_j$ be the midpoint Taylor polynomial of the shared map $\psi$
of degree $d_{\rm act}$ on the closed bin. It is the same comparator
throughout all outer blocks. The integral Taylor remainder
(the Lipschitz remainder at integer smoothness) gives
\[
 \sup_{z\in\overline B_j}|T_j(z)-\psi(z)|
 \le L_\psi h^{\beta-1}.
\]
The use of $h$, rather than the bin's possibly smaller actual width,
only enlarges this bound. The deterministic perturbation calculation
in Appendix \ref{app:regret-proofs} holds uniformly in locally
admissible $b$ and gives
$|q^\dagger-\psi(\widehat z)|
 \le C_{\rm LS}(\eta_m+\eta_\sigma)$.
Choose one actual coarse occurrence of the completed bin. For all
$z\in\overline B_j$, the triangle inequality, its Lipschitz bound and
\eqref{eq:pricing-uniform-anchor} give
\begin{align*}
 |T_j(z)-\widetilde q_j|
 &\le L_\psi h^{\beta-1}+L_\psi h
       +C_{\rm LS}(\eta_m+\eta_\sigma)+\rho_{\rm loc}/8\\
 &\le\rho_{\rm loc}/4.
\end{align*}
Thus its coefficients define a comparator in the actual whole-bin
body. At every future locally admissible refinement call, each
polynomial in that body obeys $|q-q_t^\dagger|\le3\rho_{\rm loc}/8$
by the uniform anchor bound. The radius gate gives
$a_{S_t}|q-q_t^\dagger|\le\rho_0$, so $-F_t(q)$, and hence its
composition with the affine coefficient map, is convex on the same
fixed body despite changes of shadow.

For completeness, the body has a deterministic positioning that does
not use a design distribution. Put $d=d_{\rm act}+1$, let $P_i$ be
the Legendre polynomial with $P_i(1)=1$, and set
$\Lambda_i(v)=\sqrt{2i+1}P_i(v)$ for $0\le i<d$. Express a centered
pricing polynomial as $\sum_i\alpha_i\Lambda_i(v)$ and use the known
invertible transform $u=4d\alpha/\rho_{\rm loc}$. Orthogonality for
$dv/2$ and $|P_i(v)|\le1$ give
\[
 \|\alpha\|_2
 \le\left\|\sum_i\alpha_i\Lambda_i\right\|_\infty
 \le d\|\alpha\|_2.
\]
The transformed body is consequently
\begin{equation}
 K_d=\{u:\sup_{|v|\le1}|u^\top\Lambda(v)|\le d\},
 \qquad B_2(0,1)\subseteq K_d\subseteq B_2(0,d).
 \label{eq:pricing-legendre-positioning}
\end{equation}
Translation by the observed anchor and this fixed basis change
preserve the available prices and losses. Membership is determined
by the endpoints and, for a nonconstant polynomial, its real critical
points in $[-1,1]$. A
maximizing point of a violated constraint supplies a separating
hyperplane, and the Minkowski functional is
$\sup_{|v|\le1}|u^\top\Lambda(v)|/d$. These formulas supply the
known body operations required in the primitive's declared oracle
model.

Finally, quadratic growth bounds the comparator bias at every
refinement call by
\begin{equation}
 F_t(q_t^\dagger)-F_t\{T_j(\widehat z_t)\}
 \le L_ra_+L_\psi^2h^{2\beta-2}
       +2L_ra_+C_{\rm LS}^2(\eta_m^2+\eta_\sigma^2).
 \label{eq:pricing-polynomial-bias}
\end{equation}
This follows by squaring the sum of the Taylor and fitted-input
action errors in the upper bound of
\eqref{eq:pricing-transformed-growth}. There is no shadow-estimation
error term in this comparison with the current shadow objective.

\paragraph{Predictable refinement and global terminal prefixes.}
At a bin's refinement call, include the complete chronological past,
its local history, and the current state, covariates and shadow in the
pre-seed information. Conditional on that information, define the
loss and observable feedback by
\begin{equation}
 L_t(\theta)=\frac12-
 \frac{F_t[\Pi_{\cQ}\{q_j(\bar z_t;\theta)\}]}{2Q},
 \qquad
 \ell_t=\frac12-\frac{q_tY_t}{2Q}.
 \label{eq:pricing-normalized-feedback}
\end{equation}
Both belong to $[0,1]$. The conditional feedback relation in
\eqref{eq:predictable-feedback}, after conditioning on the complete
current seed, gives $\E(\ell_t\mid\text{complete seed and pre-seed
information})=L_t(\theta_t)$. Their difference is bounded. On the
bin's uniform anchor event \eqref{eq:pricing-uniform-anchor}, both
input and action projections are inactive, and the losses are convex on the fixed body
\eqref{eq:pricing-legendre-positioning} for every locally admissible
future shadow input. This event is measurable at the bin's first
refinement initialization and hence at every later copy initialization.
Apply the conditional primitive at those local initializations.
Formally, outside that initial-information event, replace the local
protocol by constant convex losses and their exact feedback. This
analytical replacement is unused on coarse success; it does not
condition the primitive on a future local-success event.

Fix a deterministic global pricing prefix $N\le T$. If a budget-$m$
copy has received only $r\le m$ real calls by that prefix, terminate
the real protocol there and append constant virtual losses, declared
before each remaining fresh seed, with their exact constant feedback.
Causality leaves the real
actions unchanged, and for every fixed comparator $a$,
\[
 \sum_{i=1}^m\{\widetilde L_i(\widetilde A_i)
                         -\widetilde L_i(a)\}
 =\sum_{i=1}^r\{L_i(A_i)-L_i(a)\}.
\]
The global-prefix terminal signal and observations received while a
copy is idle are allowed in its pre-seed information. The completed protocol
therefore satisfies the same predictable interface. In particular,
the real-prefix violation probability is bounded by the completed
copy's failure probability, without conditioning on the random
value of $r$, any future state path, or any future shadow value.

There are at most $J_T$ copies per bin by any prefix, since their
budgets double. A union over bins, copies and the $T$ deterministic
prefixes, using \eqref{eq:pricing-copy-allocation}, costs at most
$MTJ_T\zeta_T=\delta_A/4$ on the respective initial-information
anchor events. On the resulting event, write $n_j^{\rm ref}(N)$ for bin
$j$'s real refinement count. If this is positive, the budget of its
last used copy is at most $n_j^{\rm ref}(N)$, because the preceding
budgets sum to one less than that budget. The geometric sum of
square roots is consequently at most
$(2+\sqrt2)\sqrt{n_j^{\rm ref}(N)}$. The primitive gives normalized
refinement regret at most
\[
 (2+\sqrt2)C_BL_B^{17}\sqrt{n_j^{\rm ref}(N)}.
\]
For a bin with no refinement calls the regret is zero. Summing over
bins and undoing the factor $2Q$ in
\eqref{eq:pricing-normalized-feedback} bounds the unnormalized
learning term by
\[
 2Q(2+\sqrt2)C_BL_B^{17}\sqrt{MN},
\]
since $\sum_j\sqrt{n_j^{\rm ref}(N)}\le\sqrt{MN}$.

Each coarse call loses at most $2Q$ in the transformed objective.
Let $n_j(N)$ count all calls to bin $j$ through prefix $N$, including
coarse and refinement calls. The coarse-first schedule, whose counts persist across outer blocks,
gives the exact accounting over the whole prefix
\[
 N_{\rm coarse}(N)
 =\sum_j\min\{n_j(N),Gm_{\rm coarse}\}
 \le MGm_{\rm coarse}\le MG(m_0L_c+1).
\]
Adding this cost, the refinement learning term and
\eqref{eq:pricing-polynomial-bias} proves the transformed prefix
bound for every $N\le T$ on one event. Since each transformed gap is
nonnegative,
\[
 \sum_{t=1}^N\Gamma_t
 \le2\overline\sigma\sum_{t=1}^N
                  \{F_t(q_t^\dagger)-F_t(q_t)\}.
\]
The declared $C_p\ge4\overline\sigma L_ra_+C_{\rm LS}^2$ absorbs
the fitted-input term. This proves
\eqref{eq:pricing-finite-budget-bound}, even when state scales and
shadows vary between calls.

\paragraph{True-loss transfer, flags and local stopping.}
The deterministic shadow transfer in Appendix
\ref{app:regret-proofs} holds at every locally admissible call:
\[
 \Delta_t^*\le\Gamma_t+
 \left\{\frac{L_V}{2}
       +\frac{\overline\sigma L_D^2}{2\lambda_r}\right\}
       |\widehat b_{t,S_t}-b_{S_t}^*|^2.
\]
The definition of $C_p$ proves
\eqref{eq:pricing-pointwise-shadow-transfer}. This pointwise bound is
not included in the global map-learning charge: a later block uses
its own current shadow errors, not those of all preceding blocks.

The coarse event costs at most $\delta_A/4$, and the refinement
union on its local anchor events costs at most $\delta_A/4$. Their
intersection thus has conditional probability at least $1-\delta_A$.
On that intersection,
the primitive supplies feasible coefficient actions, the anchor's
interior margin makes action projection inactive, and the local
domain assumptions place every raw input in $\mathcal Z$ and every
physical quote in $\cP$. The
certified primitive model excludes an additional numerical failure
on its success event. Hence none of the declared pricing flags is
triggered. Only this one initialization probability allocation
is used.

For completeness consider the actual predictable shadow process
without assuming that it remains local forever. Let $\tau$ be its
first pricing call at which a shadow vector violates the local radius;
it is detected before that call's fresh seed. Couple the process to
an analytical extension that agrees before $\tau$ and, from $\tau$
onward, replaces every shadow vector by $b^*$. Retain the frozen fits,
all existing bin and copy histories, and the same future randomization
mechanism; after $\tau$ the states evolve under this extended policy.
This is a predictable, everywhere locally admissible process, so the
proved certificate supplies a single event of conditional probability
at least $1-\delta_A$. On that event the actual and extended
histories agree for every prefix $N<\tau$, and the certificate and
absence of pricing flags hold for those actual prefixes. No
conditioning on $\{\tau>T\}$ or on future local success is involved.
On valid frozen fits, Assumption \ref{ass:implementation-domains}(ii)
also makes input projection the identity before $\tau$; the raw-input
identities used above therefore remain exact on these stopped prefixes.
On the same coarse event, the entire coefficient body lies in $\cQ$,
so action projection is also the identity on every stopped refinement
call. Outside a bin's initial-information anchor event, the constant-loss
analytical replacement above is used; no convexity of a genuinely
projected loss is asserted.
The outer local-stability induction can therefore use this stopped
certificate to show that the two processes agree throughout the
actual active horizon. This completes the proof.
\end{proof}

\subsection{A known all-prefix map-learning envelope}

The finite-budget coefficients give a single known envelope without
resetting the budget at an outer-block boundary. Define
\begin{equation}
 C_M=1+\frac{D_{\mathcal Z}}{c_h},\qquad
 C_{\log}=6+\log(64eC_MG),\qquad
 \eta^2=\eta_m^2+\eta_\sigma^2,
 \label{eq:pricing-map-log-constants}
\end{equation}
and choose the structural constant
\begin{align}
 C_{\rm map}=1\vee\big\{&
 4\overline\sigma QG(m_0C_{\log}+1)\notag\\
 &+4\overline\sigma Q(2+\sqrt2)C_BC_{\log}^{17}
 +2\overline\sigma L_ra_+L_\psi^2\big\}.
 \label{eq:pricing-map-constant}
\end{align}
This choice depends only on the declared structural settings and the
primitive constant $C_B$, not on $T,n_u,\delta$ or realized visits.
With the $\ell_T$ of Appendix \ref{app:operational-details}, put
\begin{equation}
 \Phi_T(n)=C_{\rm map}\ell_T^{17}
       \{M_T+\sqrt{M_Tn}+n h_T^{2\beta-2}\},\qquad
 \sum_{t=1}^N\Gamma_t\le\Phi_T(N)+C_pN\eta^2,
 \quad 1\le N\le T.
 \label{eq:pricing-map-envelope}
\end{equation}
The final inequality holds on the event of Proposition
\ref{prop:full-cover-pricing-certificate}, or before local exit in its
stopped version.

To verify this explicit choice, $T\ge H_{\rm act}$ implies
$h_T=c_hT^{-\kappa_\beta}$ and
$M_T\le C_MT^{\kappa_\beta}$. Since $\beta>1$,
$\kappa_\beta<1$, and $J_T\le T$ for integer $T\ge2$.
Consequently the single confidence allocation gives
\[
 L_c\le\log(16eC_MGT^2/\delta)\le C_{\log}\ell_T,
 \qquad
 L_B\le\log(64eC_MT^4/\delta)\le C_{\log}\ell_T.
\]
As $\ell_T\ge1$, each term in
\eqref{eq:pricing-finite-budget-bound} is bounded by the corresponding
term in \eqref{eq:pricing-map-envelope}, with the sum of the three
coefficients in \eqref{eq:pricing-map-constant}. This proves the
announced global envelope. A terminal pricing prefix uses the same
$h_T,M_T$ and persistent histories; it is not analyzed by restarting
a pricing module with budget equal to its realized length.

\section{Valuation-model guarantees and proofs}
\label{app:valuation-models}

The regression statements below concern independent bounded-response
pairs with a common covariate distribution. They apply to the full
statewise calibration samples as follows. Continue uniform-price
calibration virtually beyond its planned end. A state visit is
selected before the fresh covariate, quote and valuation noise, so
successive visits to a fixed state give the same independent-pair
experiment at every deterministic prefix length. Apply each routine
at every $N_U\le n\le n_u$ with failure level
$\varepsilon_T=\delta/(8KT)$, and take a union over states and prefixes.
This costs at most $\delta/8$, since $n_u\le T$.
On that simultaneous event and the separate count event, substitute
the actual $N_s^U\ge N_U$ without conditioning the regression
distribution on its terminal count. For the routines below, a
sample-size gate that passes at $N_U$ also passes at larger counts,
and the error radius is nonincreasing. Its deterministic worst-case
value is therefore attained at $N_U$.
The same-record analysis in Appendix~\ref{app:regret-proofs} gives
the scale-error bound for the estimator \eqref{eq:scale-estimator}.

\subsection{Uniform nonlinear regression}
\label{app:holder-cell-regression}

We first specify the multivariate H\"older convention used in
Section~\ref{sec:holder-mean-specialization}. For each state,
write $u_s(x)=m_*(x,s)-p_L$, so that $0\le u_s\le B=p_U-p_L$.
Fix $\gamma>0$ and a common radius $L_m$. With
$j=\lceil\gamma\rceil-1$, the mixed derivatives of $u_s$ through
order $j$ are continuous, and its positive-order derivatives through
$j$ are bounded by $L_m$. For a multi-index $a\in\mathbb N_0^d$, use
$|a|=\sum_\ell a_\ell$, $a!=\prod_\ell a_\ell!$,
$v^a=\prod_\ell v_\ell^{a_\ell}$, and $D^a$ for the corresponding
mixed derivative. The Taylor remainder satisfies
\begin{equation}
 \left|u_s(x)-
   \sum_{|a|\le j}\frac{D^a u_s(v)}{a!}(x-v)^a\right|
 \le L_m\|x-v\|_\infty^\gamma
 \qquad(x,v\in[0,1]^d).
 \label{eq:holder-mean-taylor}
\end{equation}
At an integer order this is the $C^{\gamma-1,1}$ convention,
consistent with the univariate definition in
Appendix~\ref{app:domain-details}.

We give the construction for one state and suppress its index.
Let $(X_i,Z_i)_{i=1}^n$ be independent pairs with a common covariate
distribution, such that $X_i\in[0,1]^d$, $Z_i\in[p_L,p_U]$, and
$\E[Z_i\mid X_i]=m(X_i)$. The conditional response distributions need not
be identical. Write $f=m-p_L$.
Put $j=\lceil\gamma\rceil-1$ and assume the density bounds in
Assumption~\ref{ass:calibration-density} and the degree-$j$ Taylor remainder
bound \eqref{eq:holder-mean-taylor}, with $u_s$ replaced by $f$.
No additional condition on the response distribution is imposed.

For an integer $J_b\ge1$, put $b=J_b^{-1}$ and partition the cube
into $M_b=J_b^d$ cubes of side length $b$.
Use half-open cells, except that the rightmost endpoint is included,
so that every $x\in[0,1]^d$ has a unique assigned cell $C(x)$.
Let $a_C$ be a cell's lower corner and define
\[
 \mathscr M_j=\{\alpha\in\mathbb N_0^d:|\alpha|\le j\},\qquad
 \Phi_j(v)=(v^\alpha)_{\alpha\in\mathscr M_j},\qquad
 D_{\rm H}=|\mathscr M_j|=\binom{d+j}{d}.
\]
In this subsection the zero multi-index is one coordinate of
$\Phi_j$, equal to one. Every coordinate has absolute value at most
one on the closed unit cube. Introduce the known moment matrix
\[
 V_j=\int_{[0,1]^d}\Phi_j(v)\Phi_j(v)^\top\,dv,\qquad
 (V_j)_{\alpha,\alpha'}=
 \prod_{\ell=1}^d\frac{1}{\alpha_\ell+\alpha'_\ell+1},
 \qquad v_{\rm H}=f_{\min}\lambda_{\min}(V_j)>0.
\]
The matrix is positive definite: the quadratic form of a nonzero
coefficient vector is the integral of the square of a nonzero
polynomial, which is positive on an open subset of the cube.
In particular, $v_{\rm H}\le f_{\min}\le f_{\max}$.

With $R_i=Z_i-p_L$, form
\begin{align}
 \widehat G_C
 &=\frac{1}{nb^d}\sum_{i=1}^n
  \one\{X_i\in C\}
  \Phi_j\!\left(\frac{X_i-a_C}{b}\right)
  \Phi_j\!\left(\frac{X_i-a_C}{b}\right)^\top,\notag\\
 \widehat t_C
 &=\frac{1}{nb^d}\sum_{i=1}^n
  \one\{X_i\in C\}
  \Phi_j\!\left(\frac{X_i-a_C}{b}\right)R_i .
 \label{eq:holder-cell-normal-equations}
\end{align}
The empirical design gate requires
$\lambda_{\min}(\widehat G_C)\ge v_{\rm H}/2$ for every cell.
If a cell fails, the routine reports a design failure.
Otherwise set $\widehat\vartheta_C=\widehat G_C^{-1}\widehat t_C$ and
\begin{equation}
 \widehat m(x)=\Pi_{[p_L,p_U]}
 \left\{p_L+
 \Phi_j\!\left(\frac{x-a_{C(x)}}b\right)^\top
                  \widehat\vartheta_{C(x)}\right\}.
 \label{eq:holder-cell-estimator}
\end{equation}
This is ordinary cellwise polynomial least squares: the normalization
by $nb^d$ cancels in each cell's normal equations.

\begin{lemma}[Uniform regression from a finite polynomial partition]
\label{lem:holder-cell-regression}
Under the preceding bounded-response, density and Taylor conditions,
fix $0<\varepsilon<1$ and an integer $J_b\ge1$.
Put
\[
 \ell_b=\log(4D_{\rm H}^2M_b/\varepsilon),\qquad
 c_G=\max\left\{1,\,
  \frac{32f_{\max}D_{\rm H}^2}{v_{\rm H}^2},\,
  \frac{8D_{\rm H}}{3v_{\rm H}}\right\}.
\]
If the deterministic gate $nb^d\ge c_G\ell_b$ passes, then
\begin{equation}
 \Pp\left(
  \text{the routine succeeds and }\ 
  \|\widehat m-m\|_\infty
       \le C_{\rm bias}b^\gamma+
          C_{\rm noise}\sqrt{\frac{\ell_b}{nb^d}}
       \right)\ge1-\varepsilon,
 \label{eq:holder-cell-uniform-bound}
\end{equation}
where the supremum includes the cell boundaries and
\[
 C_{\rm bias}=
 \left(1+\frac{4D_{\rm H}f_{\max}}{v_{\rm H}}\right)L_m,
 \qquad
 C_{\rm noise}=\frac{2D_{\rm H}B\sqrt{f_{\max}}}{v_{\rm H}}.
\]
For an explicit sample-size rule, let
\[
 L_n=\log(4D_{\rm H}^2n/\varepsilon),\qquad
 c_{\rm H}=\max\left\{2^{2\gamma+d},
                       c_G^{(2\gamma+d)/(2\gamma)}\right\}.
\]
If $n\ge c_{\rm H}L_n$, choose
\[
 J_b=\left\lfloor(n/L_n)^{1/(2\gamma+d)}\right\rfloor,\qquad
 b=J_b^{-1}.
\]
Then the deterministic gate above passes and
\eqref{eq:holder-cell-uniform-bound} holds with the radius in
\eqref{eq:holder-regression-radius}, for
$C_{\rm H}=2^\gamma C_{\rm bias}+C_{\rm noise}$.
These constants are independent of $n,\varepsilon$ and the unknown
conditional distribution of $Z_i$ given $X_i$.
\end{lemma}

\begin{proof}
First control the finitely many empirical Gram entries.
For any cell $C$ and coefficient vector $v$, a change of variables
and the density lower bound give
\[
 v^\top\E[\widehat G_C]v
 =\int_{[0,1]^d}
       \{v^\top\Phi_j(u)\}^2 f_X(a_C+bu)\,du
 \ge v_{\rm H}\|v\|_2^2.
\]
For one Gram entry, its unnormalized summand is in $[0,1]$ and its
variance is at most $f_{\max}b^d$. The bounded-scalar Bernstein
inequality therefore gives, with failure at most $2e^{-\ell_b}$,
\[
 |(\widehat G_C-\E\widehat G_C)_{\alpha,\alpha'}|
 \le \sqrt{\frac{2f_{\max}\ell_b}{nb^d}}
             +\frac{2\ell_b}{3nb^d}.
\]
This particular inequality follows from the elementary moment bound
$\log\E e^{tU}\le t^2\E[U^2]/\{2(1-|t|/3)\}$ for a centered
$|U|\le1$ and $|t|<3$, followed by the exponential Markov inequality.
The moment bound is obtained by expanding the exponential and using
$\E|U|^k\le\E U^2$ for $k\ge2$.
The deterministic gate makes the two displayed terms at most
$v_{\rm H}/(4D_{\rm H})$ each.
A union bound over $M_bD_{\rm H}^2$ entries yields an event
$\mathcal E_G$ with failure at most $\varepsilon/2$ on which
\[
 \|\widehat G_C-\E\widehat G_C\|_{\rm op}\le v_{\rm H}/2,\qquad
 \|\widehat G_C^{-1}\|_{\rm op}\le2/v_{\rm H}
 \quad\text{for every }C.
\]
Thus every design gate passes. The zero-index entry also gives
\[
 \frac{N_C}{nb^d}=(\widehat G_C)_{0,0}
 \le f_{\max}+\frac{v_{\rm H}}{2D_{\rm H}}
 \le2f_{\max},\qquad
 N_C=\sum_{i=1}^n\one\{X_i\in C\}.
\]

Next separate approximation from centered noise.
For each cell let $P_C$ be the degree-$j$ Taylor polynomial of $f$
at $a_C$, and let $\vartheta_C^0$ be its coefficients in the
rescaled basis. The Taylor condition gives
$\sup_{x\in\overline C}|f(x)-P_C(x)|\le L_mb^\gamma$.
Write $\zeta_i=R_i-f(X_i)$, and put
\[
 u_C^{\rm bias}=\frac1{nb^d}\sum_{i:X_i\in C}
       \Phi_j\!\left(\frac{X_i-a_C}{b}\right)
                        \{f(X_i)-P_C(X_i)\},
 \qquad
 u_C^{\rm noise}=\frac1{nb^d}\sum_{i:X_i\in C}
       \Phi_j\!\left(\frac{X_i-a_C}{b}\right)\zeta_i.
\]
The exact normal-equation identity is
\[
 \widehat\vartheta_C-\vartheta_C^0
       =\widehat G_C^{-1}
                     (u_C^{\rm bias}+u_C^{\rm noise}).
\]
On $\mathcal E_G$, the bias is bounded for every realized design:
\[
 \|u_C^{\rm bias}\|_2
 \le\sqrt{D_{\rm H}}\frac{N_C}{nb^d}L_mb^\gamma
 \le2\sqrt{D_{\rm H}}f_{\max}L_mb^\gamma.
\]

Conditional on all covariates, the variables $\zeta_i$ are independent
and have conditional mean zero. Since $0\le R_i\le B$, each summand
in one noise-score coordinate has conditional range at most $B$
and is zero outside its cell. Conditional Hoeffding concentration
and the preceding count bound show that, for a design in
$\mathcal E_G$,
\[
 \Pp\left(
 |(u_C^{\rm noise})_\alpha|>
 B\sqrt{\frac{f_{\max}\ell_b}{nb^d}}
 \,\middle|\,X_1,\ldots,X_n\right)\le2e^{-\ell_b}.
\]
Indeed, the sum of squared conditional ranges is at most $B^2N_C$.
Union over $M_bD_{\rm H}$ coordinates costs at most
$\varepsilon/(2D_{\rm H})\le\varepsilon/2$, conditional on any such
design. After integration, with total failure at most $\varepsilon$,
all Gram gates pass and all cells satisfy
\[
 \|u_C^{\rm noise}\|_2
 \le\sqrt{D_{\rm H}}B
                   \sqrt{\frac{f_{\max}\ell_b}{nb^d}}.
\]

On this joint event, use $\|\Phi_j(u)\|_2\le\sqrt{D_{\rm H}}$
uniformly for $u\in[0,1]^d$. For every point in a cell's closure,
\begin{align*}
 \left|\Phi_j\!\left(\frac{x-a_C}{b}\right)^\top
              \widehat\vartheta_C-f(x)\right|
 &\le L_mb^\gamma+
    \frac{2\sqrt{D_{\rm H}}}{v_{\rm H}}
             \{\|u_C^{\rm bias}\|_2+\|u_C^{\rm noise}\|_2\}\\
 &\le C_{\rm bias}b^\gamma+
              C_{\rm noise}\sqrt{\frac{\ell_b}{nb^d}}.
\end{align*}
This bound is uniform in $x$ because the finite coefficient bounds
hold in every cell and the basis is uniformly bounded; no union
over evaluation points is needed. Use the prescribed cell assignment
at boundaries. Clipping to $[p_L,p_U]$ cannot increase error relative
to $m(x)$ in that interval. This proves
\eqref{eq:holder-cell-uniform-bound}, including the routine-success
event.

Finally suppose $n\ge c_{\rm H}L_n$ and put
$x_n=(n/L_n)^{1/(2\gamma+d)}$. Then $x_n\ge2$ and
$x_n/2\le J_b\le x_n$, so
\[
 b^\gamma\le2^\gamma(L_n/n)^{\gamma/(2\gamma+d)},\qquad
 M_b\le n,\qquad \ell_b\le L_n.
\]
Moreover,
\[
 nb^d=\frac{n}{J_b^d}
 \ge L_n(n/L_n)^{2\gamma/(2\gamma+d)}
 \ge c_GL_n\ge c_G\ell_b,
 \qquad
 \sqrt{\frac{\ell_b}{nb^d}}
 \le (L_n/n)^{\gamma/(2\gamma+d)}.
\]
Substitution gives the stated $C_{\rm H}$.
The clipped error is also at most $B$, which permits the minimum
with $B$ in \eqref{eq:holder-regression-radius}.
Both the deterministic sample-size test and all empirical Gram
tests use only the observations and supplied constants. An empirical
failure is included in the failure probability, exactly as required
by Definition \ref{def:regression-interface}.
\end{proof}

\begin{proof}[Proof of Corollary~\ref{cor:holder-valuation}]
Lemma~\ref{lem:holder-cell-regression} and the state-quota argument
supply Definition \ref{def:regression-interface} with radius
\eqref{eq:holder-regression-radius}.
At fixed class constants $N_U$ is proportional to $n_u$ and the
per-state logarithm is $O(\log T)$.
Hence
\[
 \eta_m^2=\widetilde O(n_u^{-2\gamma/(2\gamma+d)}),
 \qquad
 \eta_\sigma^2=\widetilde O(\eta_m^2+n_u^{-1}).
\]
The choice $n_u\asymp T^{(2\gamma+d)/(4\gamma+d)}$ balances
$n_u$ and $T\eta_m^2$. It is sublinear and grows faster than
the deterministic regression threshold, while both calibration
radii tend to zero. The fixed precision gates therefore pass.
The prescribed warm initialization is polylogarithmic, and the
smallest active starting block is
$\widetilde O(T^{2\alpha-1})=o(T)$.
Thus the full elapsed-budget and remaining gates hold for large $T$.
Corollary~\ref{cor:regression-transfer} now gives
\eqref{eq:holder-pilot-consequence}, uniformly over the stated
class and initial state.
\end{proof}

\subsection{Uniform linear regression}

\begin{lemma}[A bounded-response least-squares interface]
\label{lem:bounded-ols}
Let $(X_i,Z_i)_{i=1}^n$ be independent pairs with $X_i\sim P_X$,
$Z_i\in[p_L,p_U]$ and
$\E(Z_i\mid X_i)=p_L+\phi(X_i)^\top\theta$.
The known map $\phi:\cX\to\R^D$ satisfies
$\sup_x\|\phi(x)\|_\infty\le L_\infty$ and
$\sup_x\|\phi(x)\|_2\le L_2$. Assume the supplied population Gram bound
\[
 G_\phi:=\E\{\phi(X)\phi(X)^\top\}\succeq\lambda_\phi I_D,
 \qquad \lambda_\phi>0.
\]
For $0<\varepsilon<1$, require the deterministic sample gate
\begin{equation}
 n\ge n_{G,\mathrm L}(\varepsilon):=
 \left\lceil\frac{8D^2L_\infty^4}{\lambda_\phi^2}
                      \log\frac{4D^2}{\varepsilon}\right\rceil.
 \label{eq:ols-sample-gate}
\end{equation}
Form $\widehat G=n^{-1}\sum_i\phi(X_i)\phi(X_i)^\top$ and report
an empirical design failure if
$\lambda_{\min}(\widehat G)<\lambda_\phi/2$. Otherwise set
\[
 \widehat\theta=\widehat G^{-1}\frac1n
                 \sum_{i=1}^n\phi(X_i)(Z_i-p_L),\qquad
 \widehat m(x)=\Pi_{[p_L,p_U]}\{p_L+\phi(x)^\top\widehat\theta\}.
\]
Then the probability that the routine succeeds and has error at most
\begin{equation}
 r_{\mathrm L}(n,\varepsilon)
 =B\wedge\left\{
 \frac{2BL_2L_\infty}{\lambda_\phi}
       \sqrt{\frac{2D\log(4D/\varepsilon)}n}\right\}
 \label{eq:ols-radius}
\end{equation}
uniformly over $x\in\cX$ is at least $1-\varepsilon$.
The reported design failure is included in this probability budget.
\end{lemma}

\begin{proof}
Write $\phi_i=\phi(X_i)$ and
$e_i=Z_i-p_L-\phi_i^\top\theta$. Boundedness of the response and its
conditional mean gives $|e_i|\le B$ and
$\E(e_i\mid X_i)=0$. For any Gram entry, the summands lie in
$[-L_\infty^2,L_\infty^2]$. Hoeffding's inequality and a union over
the $D^2$ entries give, with failure probability at most
$\varepsilon/2$,
\begin{equation}
 \|\widehat G-G_\phi\|_{\max}\le
 a_n:=L_\infty^2
       \sqrt{\frac{2\log(4D^2/\varepsilon)}n}.
 \label{eq:gram-entry-event}
\end{equation}
Here $\|A\|_{\max}=\max_{j,l}|A_{jl}|$. Since
$\|\widehat G-G_\phi\|_{\mathrm{op}}
 \le D\|\widehat G-G_\phi\|_{\max}$, the deterministic gate ensures
$\lambda_{\min}(\widehat G)\ge\lambda_\phi/2$ on this event.

Independence of the pairs and the conditional mean-zero identity
give a second Hoeffding bound. With failure probability at most
$\varepsilon/2$,
\begin{equation}
 \left\|\frac1n\sum_i\phi_i e_i\right\|_\infty
 \le t_n:=BL_\infty
                 \sqrt{\frac{2\log(4D/\varepsilon)}n}.
 \label{eq:bounded-score-event}
\end{equation}
Indeed, every coordinate summand is centered and belongs to
$[-BL_\infty,BL_\infty]$.
On the intersection, the ordinary least-squares identity yields
\[
 \|\widehat\theta-\theta\|_2
 \le\frac2{\lambda_\phi}
       \left\|\frac1n\sum_i\phi_i e_i\right\|_2
 \le\frac{2\sqrt D}{\lambda_\phi}t_n.
\]
Consequently the unprojected fitted mean has uniform error at most
$2L_2\sqrt D\,t_n/\lambda_\phi$. Projection cannot increase its
distance from a true mean in $[p_L,p_U]$ and also bounds that
distance by $B$. This proves \eqref{eq:ols-radius}; the
two failure allocations sum to $\varepsilon$.
\end{proof}

\subsection{Uniform sparse regression}

\begin{lemma}[A bounded-response Lasso interface]
\label{lem:bounded-lasso}
Let $(X_i,Z_i)_{i=1}^n$ be independent pairs with $X_i\sim P_X$,
$Z_i\in[p_L,p_U]$ and
$\E(Z_i\mid X_i)=p_L+\phi(X_i)^\top\theta$, where
$\sup_x\|\phi(x)\|_\infty\le L_\infty$. Suppose the true parameter
has at most $s_0$ nonzero coordinates and satisfies
\eqref{eq:sparse-compatibility}. Require only the
deterministic design gate
\begin{equation}
 n\ge n_{G,\mathrm S}(\varepsilon):=
 \left\lceil\frac{2048s_0^2L_\infty^4}{\kappa_{\mathrm c}^4}
                     \log\frac{4D^2}{\varepsilon}\right\rceil,
 \label{eq:lasso-sample-gate}
\end{equation}
and put
\[
 \lambda_{\rm pen}=2BL_\infty
                    \sqrt{\frac{2\log(4D/\varepsilon)}n}.
\]
Choose a fixed measurable minimizer
\begin{align}
 \widehat\theta&\in\arg\min_{u\in\R^D}
 \left\{\frac1{2n}\sum_{i=1}^n
      (Z_i-p_L-\phi(X_i)^\top u)^2+\lambda_{\rm pen}\|u\|_1\right\},
 \notag\\
 \widehat m(x)&=\Pi_{[p_L,p_U]}
                  \{p_L+\phi(x)^\top\widehat\theta\}.
 \label{eq:lasso-estimator}
\end{align}
No empirical compatibility check is performed. With probability at
least $1-\varepsilon$, the routine has uniform error at most
\begin{equation}
 r_{\mathrm S}(n,\varepsilon)
 =B\wedge\left\{
 \frac{48BL_\infty^2s_0}{\kappa_{\mathrm c}^2}
                  \sqrt{\frac{2\log(4D/\varepsilon)}n}\right\}.
 \label{eq:lasso-radius}
\end{equation}
All coordinates, including a constant feature if present, are
penalized and counted in $s_0$.
\end{lemma}

\begin{proof}
Use the Gram and score events
\eqref{eq:gram-entry-event}--
\eqref{eq:bounded-score-event}. Their proofs do not use a
population eigenvalue bound. On their intersection, the deterministic
sample-size gate
gives $a_n\le\kappa_{\mathrm c}^2/(32s_0)$. For
$J=\operatorname{supp}(\theta)\ne\varnothing$ and any $v$ in its
compatibility cone, $\|v\|_1\le4\|v_J\|_1$ and hence
\begin{align*}
 v^\top\widehat Gv
 &\ge v^\top G_\phi v-a_n\|v\|_1^2\\
 &\ge\left\{\frac{\kappa_{\mathrm c}^2}{|J|}-16a_n\right\}
                           \|v_J\|_1^2
 \ge\frac{\kappa_{\mathrm c}^2}{2|J|}\|v_J\|_1^2.
\end{align*}
This compatibility bound holds on the Gram concentration event
without a support test.

Let $v=\widehat\theta-\theta$. Optimality in
\eqref{eq:lasso-estimator} and
$\lambda_{\rm pen}=2t_n$ give the basic inequality
\[
 \frac12v^\top\widehat Gv+
    \frac{\lambda_{\rm pen}}2\|v_{J^c}\|_1
 \le\frac{3\lambda_{\rm pen}}2\|v_J\|_1.
\]
Thus $v$ lies in the compatibility cone, and
$v^\top\widehat Gv\le3\lambda_{\rm pen}\|v_J\|_1$.
Combining this upper bound with empirical compatibility gives
\[
 \|v_J\|_1\le\frac{6|J|\lambda_{\rm pen}}{\kappa_{\mathrm c}^2},
 \qquad
 \|v\|_1\le\frac{24s_0\lambda_{\rm pen}}{\kappa_{\mathrm c}^2}.
\]
If $J=\varnothing$, the basic inequality instead forces $v=0$ on
the score event. In both cases,
$\sup_x|\phi(x)^\top v|\le L_\infty\|v\|_1$ proves the claimed
radius before clipping, and projection supplies the minimum with
$B$. The two failure allocations sum to $\varepsilon$, proving the
uniform regression interface.
\end{proof}

\subsection{Calibration choices and the model-specific regrets}
\label{app:model-calibration-budgets}

For $\rho\in\{\mathrm L,\mathrm S\}$ write the respective mean
radius as $r_\rho(n,\varepsilon)\le
A_\rho\sqrt{\log(4D/\varepsilon)/n}$, where
\[
 A_{\rm L}=\frac{2\sqrt2 BL_2L_\infty\sqrt D}{\lambda_\phi},
 \qquad
 A_{\rm S}=\frac{48\sqrt2 BL_\infty^2s_0}{\kappa_{\rm c}^2}.
\]
Let $\Lambda_T=\ell_T+2\log(D+1)$ and put
\[
 A_{\sigma,\rho}=
 \frac B{\underline\sigma}A_\rho+
 \frac{3B^2}{\sqrt2\,\underline\sigma},
 \qquad
 \mathfrak A_\rho=A_\rho^2+A_{\sigma,\rho}^2.
\]
At the per-state, per-prefix confidence $\varepsilon_T$,
$\log(4D/\varepsilon_T)\le\Lambda_T$.
The scale radius in Appendix~\ref{app:operational-details} therefore gives
\begin{equation}
 \eta_m\le A_\rho\sqrt{\Lambda_T/N_U},\qquad
 \eta_\sigma\le A_{\sigma,\rho}\sqrt{\Lambda_T/N_U},\qquad
 \eta^2\le\mathfrak A_\rho\Lambda_T/N_U.
 \label{eq:parametric-pilot-radius}
\end{equation}

Here is a deterministic calibration choice that accounts for every
precision floor. The quantities on the right below are the settings
of Appendix~\ref{app:operational-details}. Define
\[
 \eta_{\rm cal}=\min\left\{\eta_0,\,
 \frac{\epsilon}{16C_{\rm app}},\,
 \sqrt{\frac{q_S(1-\lambda)r}{128C_pC_\varrho}},\,
 \frac{A_{\rm loc}}{8C_{\rm drift}},\,
 \frac{\rho_{\rm loc}}{32C_{\rm LS}}\right\}.
\]
The condition $\max\{\eta_m,\eta_\sigma\}\le\eta_{\rm cal}$
implies the mean, scale, pilot-forcing and localization precision
gates: in particular $\eta^2\le2\eta_{\rm cal}^2$ and
$\eta_m+\eta_\sigma\le2\eta_{\rm cal}$.
Let $n_{G,\rho}$ be the deterministic sample gate in
Lemma~\ref{lem:bounded-ols} or \ref{lem:bounded-lasso}, and set
\begin{align}
 g_\rho(T)&=\max\left\{1,16\ell_T,\,
 n_{G,\rho}(\varepsilon_T),\,
 \mathfrak A_\rho\Lambda_T/\eta_{\rm cal}^2\right\},\notag\\
 n_\rho^\star&=\left\lceil\max\left\{g_\rho(T),\,
            \sqrt{q_S\mathfrak A_\rho T\Lambda_T}\right\}\right\rceil,
 \qquad
 n_u=1+\left\lceil2n_\rho^\star/q_S\right\rceil.
 \label{eq:parametric-calibration-choice}
\end{align}
Then $N_U\ge n_\rho^\star$ and
$q_S(n_u-1)\ge32\ell_T$. Thus the regression sample gate and all
calibration precision gates pass. The remaining active-block and
elapsed-horizon gates are still checked by the policy.

\begin{proof}[Proof of Corollaries \ref{cor:linear-valuation}
and \ref{cor:sparse-valuation}]
The statewise sampling argument above verifies the regression
interface at failure level $\varepsilon_T$ per state and prefix.
Use the initialization quotas and smallest admissible $L_0$ from
Corollary~\ref{cor:regression-transfer}.
Define the non-calibration remainder
\[
 \mathcal R_T=1+\tau_{\rm w}+L_0+\ell_T^{17}T^\alpha
  +\ell_T^{34}T^{2\alpha-1}\log(eT)
  +\ell_T^2\log(eT)+T\delta .
\]
Substitution of \eqref{eq:parametric-pilot-radius} into the finite
master bound gives
\[
 \Reg_T\le C\left\{\mathcal R_T+
 \frac{n_\rho^\star}{q_S}
 +\frac{T\mathfrak A_\rho\Lambda_T}{n_\rho^\star}
 +\frac{T\mathfrak A_\rho^2\Lambda_T^2}{(n_\rho^\star)^2}\right\}.
\]
The constant here depends only on the common dynamic and pricing
bounds, not the displayed feature and sparsity factors.
The target quota is at least
$\sqrt{q_S\mathfrak A_\rho T\Lambda_T}$ and at most the sum of
that quantity, $g_\rho(T)$ and two. Consequently
\begin{equation}
 \Reg_T\le C\left\{\mathcal R_T+
 \frac{g_\rho(T)+1}{q_S}
 +\sqrt{\frac{\mathfrak A_\rho T\Lambda_T}{q_S}}
 +\frac{\mathfrak A_\rho\Lambda_T}{q_S}\right\}.
 \label{eq:parametric-balanced-regret}
\end{equation}
This bound retains the cost of a sample or precision gate that
exceeds the balancing quota.

For the fixed linear class,
$\mathfrak A_{\rm L}$ and $\eta_{\rm cal}$ are constant and
$n_{G,{\rm L}}(\varepsilon_T)=O(\log T)$.
Thus calibration uses $\widetilde O(\sqrt T)$ periods and its error
radii tend to zero. The initialization budget is polylogarithmic,
and $L_0=\widetilde O(T^{2\alpha-1})$.
All fixed precision and total-budget gates therefore pass for
large $T$. Equation \eqref{eq:parametric-balanced-regret}
proves Corollary~\ref{cor:linear-valuation} uniformly over the
stated class bounds.

For the sparse class, the compatibility constant satisfies
$\kappa_{\rm c}^2\le s_0L_\infty^2$ whenever the true support is
nonempty: apply compatibility to a coordinate vector on that support.
Under the valuation model a zero coefficient vector would give
mean $p_L$ and is incompatible with positive variance and
valuation support $[p_L,p_U]$. Thus the nonempty case applies.
It follows that
$\mathfrak A_{\rm S}\le C s_0^2/\kappa_{\rm c}^4$,
where $C$ depends on the fixed value, feature and scale bounds.
The same order controls the deterministic sample and precision
floors, up to logarithms. Using
$\mathcal R_T=\widetilde O(T^\alpha)$ gives
\eqref{eq:sparse-valuation-regret}, with all displayed dependencies
retained. If $D,s_0,\kappa_{\rm c}$ are fixed, the calibration budget
is again $\widetilde O(\sqrt T)$ and all gates eventually pass.
This proves Corollary~\ref{cor:sparse-valuation}.
\end{proof}

\subsection{The fixed-dimensional linear lower bound}

\begin{proof}[Proof of Theorem~\ref{thm:linear-map-lower}]
We embed the lower-bound-normalized pricing experiment of
\citet{orbit}, retaining its full physical price interval. We first
verify that one set of envelopes contains the entire embedded family.

Write $B_0=31/32$. In the source construction, an auxiliary random
variable $A_\omega\in[0,B_0]$ has survival function
\[
 H_\omega(a)=H_0(a)+\kappa w^\beta
       \sum_{j=1}^{M}\omega_j\varphi((a-z_j)/w),
 \quad w=c_w T^{-1/(4\beta-3)},\quad
 M=\lfloor(64w)^{-1}\rfloor,
\]
where $\varphi\in C_0^\infty([-1/8,1/8])$ is the source's fixed odd
bump with $\varphi'(0)=1$, $z_j=B_0/2-jw$, and
$\omega\in\{-1,1\}^M$. The baseline $H_0$ is smooth, agrees exactly
with $1-a/B_0$ on $[1/8,7/8]$, and is within a fixed arbitrarily small
$\varepsilon_0$ of the uniform survival function on the real line.
These properties, and the full-$[0,1]$ pricing lower bound for this
family, are verified in the lower-bound construction of \citet{orbit}.
The bumps have disjoint supports and zero integrals. Hence
\[
 \mu_0:=\E A_\omega=\int_0^{B_0} H_\omega(a)\,da
\]
does not depend on $\omega$. Let $\tau_\omega^2=\operatorname{Var}(A_\omega)$
and $\tau_*=B_0/\sqrt{12}$. The identities for the first two moments of
a bounded nonnegative variable give
\[
 |\mu_0-B_0/2|\le B_0\varepsilon_0,
 \qquad
 |\tau_\omega-\tau_*|\le C(\varepsilon_0+\kappa)
\]
after choosing the fixed constants small enough. In particular, one
fixed positive scale interval contains every $\tau_\omega$, uniformly
over $T$ and $\omega$.

For $X_1$ take the source's actual covariate distribution, uniform on
$\{\mu_0+2jw:1\le j\le M\}$. For $d>1$, append independent
Bernoulli$(1/2)$ coordinates, independent of the source experiment.
Use the same covariate distribution in every state and set
\[
 m_*(x,s)=x_1,\qquad \sigma_s=\tau_\omega,
 \qquad \xi_t=(A_{\omega,t}-\mu_0)/\tau_\omega,
 \qquad P_0(\cdot\mid s)=P_1(\cdot\mid s)=\nu,
\]
where $\nu$ is uniform on $[K]$. All innovations are independent.
The noise now has mean zero and variance one, while the valuation is
still $X_1+A_\omega-\mu_0$. At $X_1=\mu_0+2jw$ it equals
$2jw+A_\omega\in[0,1]$. Thus neither physical prices nor physical
revenues have been translated or changed by standardization.

We check the pricing and deployment conditions explicitly. At the
uniform baseline, the standardized noise survival function is
$g_*(r)=1/2-r/\sqrt{12}$ on the relevant residual interval. Choose
\[
 \begin{gathered}
 \mathcal Z=[1.65,1.90],\quad \mathcal Z^+=[1.60,2.00],\qquad
 \cQ=[1.50,2.00],\quad \cQ^+=[1.40,2.10],\\
 \mathcal I=[-0.70,0.60],\qquad
 \cW=[-0.12,0.07],\quad \cW^{\rm b}=[-0.16,0.11].
 \end{gathered}
\]
The true baseline indices lie in
$[\sqrt3,\sqrt3+\sqrt{12}/31]\subset\operatorname{int}(\mathcal Z)$.
For $(z,q)\in\mathcal Z^+\times\cQ^+$, $q-z\in[-0.60,0.50]$,
and
\[
 \psi_*(z)=(\sqrt3+z)/2\in[1.666,1.867]
       \subset\operatorname{int}(\cQ),\qquad
 -\partial_q^2\{qg_*(q-z)\}=2/\sqrt{12}.
\]
The auxiliary residual $\mu_0+\tau_\omega r$ for
$r\in\mathcal I$ remains strictly in $[1/8,7/8]$ once the fixed
moment perturbations are sufficiently small. On this interval the
baseline tail is exactly linear. The disjoint-bump construction has
uniform $\beta$-H\"older bounds and first- and second-derivative
perturbations at most $C\kappa$ for $\beta\ge2$.
It follows that, on the entire enlarged rectangle, the standardized
objective has second derivative between two fixed negative constants,
and its unique zero of the first derivative remains a fixed distance
inside $\cQ$. Taylor's theorem gives both quadratic-growth bounds.
The first-order equation is uniformly $(\beta-1)$-H\"older with its
action derivative bounded away from zero. For $\beta=2$, the mean-value
theorem gives a uniform Lipschitz root; for larger orders, implicit
differentiation and the Taylor remainder give the corresponding
$(\beta-1)$-H\"older bound, as in the oracle-map regularity argument
of \citet{orbit}. This verifies Assumptions~\ref{ass:tail-regularity} and \ref{ass:demand-geometry} with common
envelopes.

The full-interval optimum must also be retained. In the local source
coordinate $c\in[0,1/32]$, the uniform-tail revenue has optimizer
$(B_0+c)/2\in[31/64,1/2]$. Outside
$J=[13/32,9/16]$, its revenue gap is at least $1/(256B_0)$:
on the nonsaturated part this is the quadratic gap, and in the two
saturated parts the gap is larger. Uniform closeness of $H_\omega$
to the uniform tail preserves a positive outside gap. On $J$ the
objective is strictly concave by the preceding derivative bounds,
and its first-order equation places the optimum within $C\kappa$
of $(B_0+c)/2$. The full-$[0,1]$ optimizer is consequently unique.
Its standardized residual lies strictly in $\cW$, because the
baseline residuals lie in $[-\sqrt{12}/62,0]$.

Take the supplied coverage constant to be $a_0=1/2$; the constructed
rows satisfy this conservative certificate. The choice below one
does not force other members of the declared upper-bound class to
have purchase-independent transitions. At the baseline,
$g_*(r)\in[0.326,0.703]$ on $\mathcal I$, so, for example,
$\kappa_D=1/4$ is valid after reducing the fixed perturbations.
Choose $\kappa_0=1/100$ and scale bounds sufficiently close to
$\tau_*$ that $\kappa_0\le\underline\sigma^2$.
This verifies Assumption~\ref{ass:coverage} and the calibration-overlap
clause in Assumption~\ref{ass:implementation-domains}.

Because all states have identical pricing primitives, the reference
relative value is $h^*=0$ and the true shadow is zero. The full-price
optimum is inside the reference residual set with the verified margin,
giving Assumption~\ref{ass:utility-domain}. Choose a fixed small
shadow interval $\cB=[-b_0,b_0]$ and then a fixed small error radius
$\eta_0$. At the baseline, all residual exploration prices and all
prices $\widehat\sigma_s q-\widehat b_s$ lie strictly in $[0,1]$;
the fitted indices, true residuals, and transformed oracle actions
also have positive margins in their respective intervals. The same
is true uniformly after small mean, scale and shadow perturbations,
by continuity on these fixed compact intervals. Choosing
$\varepsilon_0,\kappa,b_0,\eta_0$ once, small enough for these finitely
many strict margins, proves Assumption~\ref{ass:implementation-domains}. All selected constants are
independent of $T$ and of the choice of $\omega$.

It remains to check the linear design rather than merely smooth
membership. For sufficiently large $T$, $M\ge2$ and $Mw\ge1/128$.
Therefore
\[
 \operatorname{Var}(X_1)=w^2(M^2-1)/3
      \ge (Mw)^2/4\ge2^{-16}.
\]
The additional coordinates give
$\operatorname{Cov}(X)\succeq v_0I_d$ with $v_0=2^{-16}$.
For the intercept-augmented feature map, put $\mu=\E X$. Then
\[
 \E(a+v^\top X)^2
 \ge v_0\|v\|_2^2+(a+v^\top\mu)^2,
 \qquad
 a^2+\|v\|_2^2
 \le2(a+v^\top\mu)^2+(2d+1)\|v\|_2^2.
\]
Thus a common choice $\lambda_\phi=v_0/(2d+1)$ suffices. The
coefficient vector selecting $x_1$ is fixed and one-sparse, so it
belongs to a fixed affine-linear coefficient envelope. The changing
finite covariate supports all belong to one horizon-independent
class of distributions satisfying this Gram floor.

Finally, a static simulator can run any dynamic pricing policy by
retaining the specified initial state and drawing each later state
independently from $\nu$. It also generates the additional coordinates.
None of this extra randomization depends on $\omega$. It supplies the
resulting state and covariates to the policy, posts its chosen physical
price, and returns the source purchase outcome. The observed transcript
is exactly that of the constructed dynamic instance. As state labels
do not affect valuations or transitions here, the dynamic finite-horizon
oracle is the static contextual oracle in every period. The two regrets
are therefore equal, for every adaptive policy and every $\omega$.
The source lower-bound theorem yields the asserted inequality.
\end{proof}

\end{appendices}

\begin{thebibliography}{99}
\bibitem[Agrawal and Tang(2024)]{agrawaltang2024}
S. Agrawal and W. Tang.
\newblock Dynamic Pricing and Learning with Long-term Reference Effects.
\newblock Full version, \href{https://arxiv.org/abs/2402.12562v2}{arXiv:2402.12562v2}, 2024.
Extended abstract in \emph{Proceedings of the 25th ACM Conference on Economics and Computation}, p.~72, 2024.
\newblock \href{https://doi.org/10.1145/3670865.3673599}{doi:10.1145/3670865.3673599}.

\bibitem[Agrawal et~al.(2021)Agrawal, Yin, and Zeevi]{agrawal2021bass}
S. Agrawal, S. Yin, and A. Zeevi.
\newblock Dynamic Pricing and Learning under the Bass Model.
\newblock Full version, \href{https://arxiv.org/abs/2103.05199v1}{arXiv:2103.05199v1}, 2021.
Extended abstract in \emph{Proceedings of the 22nd ACM Conference on Economics and Computation}, pp.~2--3, 2021.
\newblock \href{https://doi.org/10.1145/3465456.3467546}{doi:10.1145/3465456.3467546}.

\bibitem[Ban and Keskin(2021)]{bankeskin2021}
G.-Y. Ban and N. B. Keskin.
\newblock Personalized Dynamic Pricing with Machine Learning:
High-Dimensional Features and Heterogeneous Elasticity.
\newblock \emph{Management Science}, 67(9):5549--5568, 2021.
\newblock \href{https://doi.org/10.1287/mnsc.2020.3680}{doi:10.1287/mnsc.2020.3680}.

\bibitem[Broder and Rusmevichientong(2012)]{broderrusmevichientong2012}
J. Broder and P. Rusmevichientong.
\newblock Dynamic Pricing Under a General Parametric Choice Model.
\newblock \emph{Operations Research}, 60(4):965--980, 2012.
\newblock \href{https://doi.org/10.1287/opre.1120.1057}{doi:10.1287/opre.1120.1057}.

\bibitem[\c{C}elik and Maglaras(2008)]{celikmaglaras2008}
S. \c{C}elik and C. Maglaras.
\newblock Dynamic Pricing and Lead-Time Quotation for a Multiclass Make-to-Order Queue.
\newblock \emph{Management Science}, 54(6):1132--1146, 2008.
\newblock \href{https://doi.org/10.1287/mnsc.1070.0836}{doi:10.1287/mnsc.1070.0836}.

\bibitem[Chai et~al.(2026)Chai, Duan, Fan, and Wang]{chai2026featurediversity}
J. Chai, Y. Duan, J. Fan, and K. Wang.
\newblock Optimal Semiparametric Dynamic Pricing with Feature Diversity.
\newblock Working paper, \href{https://arxiv.org/abs/2605.04207v1}{arXiv:2605.04207v1}, 2026.

\bibitem[E.~Chen et~al.(2026)Chen, Chen, Gao, and Li]{chen2024doubly}
E. Chen, X. Chen, L. Gao, and J. Li.
\newblock Dynamic Contextual Pricing with Doubly Non-Parametric Random Utility Models.
\newblock \emph{Management Science}, forthcoming, 2026.
Full preprint, \href{https://arxiv.org/abs/2405.06866v3}{arXiv:2405.06866v3}, 2024.

\bibitem[Chen and Gallego(2021)]{chengallego2021}
N. Chen and G. Gallego.
\newblock Nonparametric Pricing Analytics with Customer Covariates.
\newblock \emph{Operations Research}, 69(3):974--984, 2021.
\newblock \href{https://doi.org/10.1287/opre.2020.2016}{doi:10.1287/opre.2020.2016}.

\bibitem[X.~Chen et~al.(2026)Chen, Hong, and Liu]{chen2026queues}
X. Chen, G. Hong, and Y. Liu.
\newblock Online Learning and Optimization for Queues with Unknown Arrival Rate
and Service Distribution.
\newblock \emph{Operations Research}, 74(4):2053--2074, 2026.
\newblock \href{https://doi.org/10.1287/opre.2023.0304}{doi:10.1287/opre.2023.0304}.

\bibitem[Cohen et~al.(2020)Cohen, Lobel, and Paes~Leme]{cohen2020features}
M. C. Cohen, I. Lobel, and R. Paes Leme.
\newblock Feature-Based Dynamic Pricing.
\newblock \emph{Management Science}, 66(11):4921--4943, 2020.
\newblock \href{https://doi.org/10.1287/mnsc.2019.3485}{doi:10.1287/mnsc.2019.3485}.

\bibitem[den~Boer and Keskin(2022)]{denboerkeskin2022}
A. V. den Boer and N. B. Keskin.
\newblock Dynamic Pricing with Demand Learning and Reference Effects.
\newblock \emph{Management Science}, 68(10):7112--7130, 2022.
\newblock \href{https://doi.org/10.1287/mnsc.2021.4234}{doi:10.1287/mnsc.2021.4234}.

\bibitem[Fan et~al.(2024)Fan, Guo, and Yu]{fan2024semiparametric}
J. Fan, Y. Guo, and M. Yu.
\newblock Policy Optimization Using Semiparametric Models for Dynamic Pricing.
\newblock \emph{Journal of the American Statistical Association},
119(545):552--564, 2024.
\newblock \href{https://doi.org/10.1080/01621459.2022.2128359}{doi:10.1080/01621459.2022.2128359}.

\bibitem[Fan et~al.(2026)Fan, Han, Lv, Xu, and Zhou]{orbit}
Y. Fan, Y. Han, J. Lv, X. Xu, and Z. Zhou.
\newblock Harnessing Unimodality in Semiparametric Contextual Pricing via
Oracle Price Map Learning.
\newblock Working paper, \href{https://arxiv.org/abs/2605.15411v1}{arXiv:2605.15411v1}, 2026.

\bibitem[Fokkema et~al.(2024)Fokkema, van~der Hoeven, Lattimore, and Mayo]{fokkema}
H. Fokkema, D. van der Hoeven, T. Lattimore, and J. J. Mayo.
\newblock Online Newton Method for Bandit Convex Optimisation.
\newblock Full version, \href{https://arxiv.org/abs/2406.06506v1}{arXiv:2406.06506v1}, 2024.
Extended abstract in \emph{Proceedings of the 37th Conference on Learning Theory},
PMLR 247:1713--1714, 2024.

\bibitem[Hu et~al.(2025)Hu, Li, and Wager]{huliwager}
Y. Hu, S. Li, and S. Wager.
\newblock Optimal Targeting in Dynamic Systems.
\newblock Working paper, \href{https://arxiv.org/abs/2507.00312v4}{arXiv:2507.00312v4},
2025 (revised 2026).

\bibitem[Jaksch et~al.(2010)Jaksch, Ortner, and Auer]{jaksch2010rl}
T. Jaksch, R. Ortner, and P. Auer.
\newblock Near-optimal Regret Bounds for Reinforcement Learning.
\newblock \emph{Journal of Machine Learning Research}, 11(51):1563--1600, 2010.
\newblock \href{https://jmlr.org/papers/v11/jaksch10a.html}{jmlr.org/papers/v11/jaksch10a.html}.

\bibitem[Jia et~al.(2024)Jia, Shi, and Shen]{jia2024reusable}
H. Jia, C. Shi, and S. Shen.
\newblock Online Learning and Pricing for Service Systems with Reusable Resources.
\newblock \emph{Operations Research}, 72(3):1203--1241, 2024.
\newblock \href{https://doi.org/10.1287/opre.2022.2381}{doi:10.1287/opre.2022.2381}.

\bibitem[Keskin and Li(2024)]{keskinli2024markov}
N. B. Keskin and M. Li.
\newblock Selling Quality-Dif\-fer\-en\-ti\-at\-ed Products in a Markovian Market
with Unknown Transition Probabilities.
\newblock \emph{Operations Research}, 72(3):885--902, 2024.
\newblock \href{https://doi.org/10.1287/opre.2022.0316}{doi:10.1287/\allowbreak opre.2022.0316}.

\bibitem[Kostami and Koenigsberg(2026)]{kostamikoenigsberg}
V. Kostami and O. Koenigsberg.
\newblock The Operational Challenges of Customer Growth.
\newblock HEC Paris Research Paper MOSI-2026-1664, 2026.
\newblock \href{https://doi.org/10.2139/ssrn.7566778}{doi:10.2139/ssrn.7566778}.

\bibitem[Lian(2026)]{lian2026competition}
Z. Lian.
\newblock Online Learning in Queue-Based Service Competition.
\newblock Working paper, SSRN:7569020, 2026.
\newblock \href{https://doi.org/10.2139/ssrn.7569020}{doi:10.2139/ssrn.7569020}.

\bibitem[Lu et~al.(2013)Lu, Musalem, Olivares, and Schilkrut]{lu2013queues}
Y. Lu, A. Musalem, M. Olivares, and A. Schilkrut.
\newblock Measuring the Effect of Queues on Customer Purchases.
\newblock \emph{Management Science}, 59(8):1743--1763, 2013.
\newblock \href{https://doi.org/10.1287/mnsc.1120.1686}{doi:10.1287/mnsc.1120.1686}.

\bibitem[Xu and Wang(2024)]{xuwang2024heteroscedastic}
J. Xu and Y.-X. Wang.
\newblock Pricing with Contextual Elasticity and Heteroscedastic Valuation.
\newblock In \emph{Proceedings of the 41st International Conference on Machine Learning},
PMLR 235:55286--55304, 2024.
\newblock \href{https://proceedings.mlr.press/v235/xu24x.html}{proceedings.mlr.press/v235/xu24x.html}.

\bibitem[Yuan et~al.(2026)Yuan, Du, and Hu]{yuan2026selfexciting}
Q. Yuan, L. Du, and M. Hu.
\newblock Dynamic Pricing Under Self-Exciting Arrival Processes.
\newblock \emph{Operations Research}, 74(3):1408--1424, 2026.
\newblock \href{https://doi.org/10.1287/opre.2024.1172}{doi:10.1287/opre.2024.1172}.

\bibitem[Zhang et~al.(2022)Zhang, Megahed, and Frazier]{zhang2022relationships}
X. Zhang, A. Megahed, and P. I. Frazier.
\newblock Dynamic Pricing with Long-Term Relationships.
\newblock Working paper, SSRN:4149982, 2022.
\newblock \href{https://doi.org/10.2139/ssrn.4149982}{doi:10.2139/ssrn.4149982}.

\end{thebibliography}
\end{document}